\documentclass[twoside]{article}

\usepackage[accepted]{aistats2022}
\usepackage{graphicx}				
\usepackage{amssymb}
\usepackage{amsmath}
\usepackage{fancyhdr}
\usepackage[flushleft]{threeparttable}
\usepackage{pgfplots}
\usepackage{enumerate}
\usepackage{mathtools}   
\usepackage[linesnumbered,ruled,vlined]{algorithm2e}
\SetKwInput{KwInput}{Input}                
\SetKwInput{KwOutput}{Output} 
\usepackage{graphicx}
\usepackage{graphics}
\usepackage{subfigure}
\usepackage{tabularx}
\usepackage{url}
\usepackage{xurl}
\usepackage{amsthm,amsfonts,amsbsy,latexsym,dsfont,tikz,enumitem}

\usepackage[
    colorlinks=true,%
    breaklinks,
    hyperindex,%
    plainpages=false,%
    bookmarksopen=false,%
    linkcolor=blue,%
    anchorcolor=blue,%
    citecolor=blue,%
    filecolor=black,%
    menucolor=black,%
    pagecolor=black,%
    urlcolor=blue,%
    pdfview=FitH,
    pdfstartview=FitH,
    hyperfigures,
    bookmarksnumbered%
  ]{hyperref} 

\usepackage{natbib}
\setcitestyle{authoryear,open={(},close={)}} 

\DeclareMathOperator{\tr}{tr}

\newcommand{\bP}{\mathbb{P}}\newcommand{\bR}{\mathbb{R}}

\newcommand{\vone}{\mathbf{1}}

\newcolumntype{L}[1]{>{\raggedright\arraybackslash}p{#1}}

\theoremstyle{remark}
\newtheorem{thm}{Theorem}[section]
\newtheorem{lem}[thm]{Lemma}
\newtheorem{cor}[thm]{Corollary}

\newtheorem{rem}[thm]{Remark}

\newtheorem{conj}[thm]{Conjecture}   

\begin{document}
\setlength{\belowcaptionskip}{-1.5em}

\twocolumn[

\aistatstitle{Sketch-and-Lift: Scalable Subsampled Semidefinite Program for $K$-means Clustering}

\aistatsauthor{ Yubo Zhuang \And Xiaohui Chen \And Yun Yang }

\aistatsaddress{ University of Illinois at Urbana-Champaign} ]

\begin{abstract}%
Semidefinite programming (SDP) is a powerful tool for tackling a wide range of computationally hard problems such as clustering. Despite the high accuracy, semidefinite programs are often too slow in practice with poor scalability on large (or even moderate) datasets. In this paper, we introduce a linear time complexity algorithm for approximating an SDP relaxed $K$-means clustering. The proposed \emph{sketch-and-lift} (SL) approach solves an SDP on a subsampled dataset and then propagates the solution to all data points by a nearest-centroid rounding procedure. It is shown that the SL approach enjoys a similar exact recovery threshold as the $K$-means SDP on the full dataset, which is known to be information-theoretically tight under the Gaussian mixture model. The SL method can be made adaptive with enhanced theoretic properties when the cluster sizes are unbalanced. Our simulation experiments demonstrate that the statistical accuracy of the proposed method outperforms state-of-the-art fast clustering algorithms without sacrificing too much computational efficiency, and is comparable to the original $K$-means SDP with substantially reduced runtime.
\end{abstract}

\section{INTRODUCTION}
Clustering is a widely explored unsupervised machine learning task to partition data into a fewer number of unknown groups. The $K$-means clustering is a classical clustering method with good empirical performance on recovering the cluster labels for Euclidean data~\citep{MacQueen1967_kmeans}. Under the Gaussian mixture model (GMM) with isotropic noise, $K$-means clustering is equivalent to the maximum likelihood estimator (MLE) for cluster labels, which is known to be worst-case NP-hard~\citep{aloise2009np}. Fast approximation algorithms to solve the $K$-means such as Lloyd's algorithm~\citep{Lloyd1982_TIT,LuZhou2016} and spectral clustering~\citep{Meila01learningsegmentation,NgJordanWeiss2001_NIPS,Vempala04aspectral,Achlioptas2005McSherry,vanLuxburg2007_spectralclustering,vonLuxburgBelkinBousquet2008_AoS} provably yield consistent recovery when different groups are well separated. Recently, semi-definite programming (SDP) relaxations~\citep{PengWei2007_SIAMJOPTIM,MixonVillarWard2016,LiLiLingStohmerWei2017,FeiChen2018,CHEN2021303,Royer2017_NIPS,GiraudVerzelen2018,BuneaGiraudRoyerVerzelen2016} have emerged as an important approach for clustering due to its superior empirical performance~\citep{PengWei2007_SIAMJOPTIM}, robustness against outliers and adversarial attack~\citep{FeiChen2018}, and attainment of the information-theoretic limit~\citep{chen2021cutoff}. Despite having polynomial time complexity, the SDP relaxed $K$-means has notoriously poor scalability to large (or even moderate) datasets for instance by interior point methods~\citep{Alizadeh1995,Jiang2020_FOCS}, as the typical runtime complexity of an interior point algorithm for solving the SDP is at least $O(n^{3.5})$, where $n$ is the sample size.
Hence the goal of this paper is to derive a computationally cheap approximation to the SDP relaxed $K$-means formulation for reducing the time complexity while maintaining statistical optimality.

Sketching, a popular numerical technique in randomized linear algebra to speed up matrix computations via compressing a matrix to a much smaller one by multiplying a random matrix~\citep{drineas2017lectures}, has been deployed in recent years as a valuable tool in many data science applications at scale. We refer papers from~\cite{BluhmFranca2019,yurtsever2017sketchy} for sketching semidefinite programs. In this paper, we consider \emph{subsampling sketches}, constructed by subsampling $m$ out of the total $n$ data points (seen as a random projection with independent Bernoulli weights). Subsampling sketches perform sample size reduction to substantially reduce the runtime complexity when $m \ll n$ by instead solving a much smaller $m$-dimensional SDP. 

In this paper, we propose a \emph{sketch-and-lift} (SL) approach for fast and scalable clustering. The main goal for the SL approach is to look for the smallest possible projected sample size $m$ in order to maximally reduce the computational cost without sacrificing too much statistical accuracy. Under the GMM, we show that the subsampled size $m$ can be made almost independent of $n$ to guarantee exact recovery on randomly subsampled data points when the signal is above a threshold depending on the down-sampling ratio $m/n$. To reconstruct a solution to the full dataset, we need to project back (or lift) from cluster labels estimated by an $m$-dimensional SDP to cluster labels of the entire $n$ data points. This back projection step takes $O(n)$ complexity. Thus the proposed SL approach has an overall \emph{linear} time complexity as long as $m=O(n^{c})$ for some constant $c \in (0,1)$, which substantially mitigates the high polynomial runtime complexity of solving the original SDP relaxed $K$-means. For instance, we can set $c = 2/7$ if the interior point method is used to solve the SDP~\citep{Jiang2020_FOCS}.

The baseline SL procedure begins with a uniform subsampling on the entire dataset. When the cluster sizes are unequal, the single down-sampling parameter $\gamma$ creates a non-trivial bias on restricting the data dimension growth rate. Motivated from this observation, we propose two SL variants: one based on bias-correction by equalizing the size of the estimated clusters from the subsampled SDP, and the other based on adaptively choosing the sampling weights on the input data points. By doing so, we show that the constraint on the data dimension is unnecessary after bias correction, and the bias-corrected SL and weighted SL boost numeric performance for unequal cluster size case.

The rest of paper is structured as follows. In Section~\ref{sec:background}, we describe some background on the $K$-means clustering and its SDP relaxation. In Section~\ref{sec:methods}, we present our SL approach and its variants. In Section~\ref{sec:theory}, we derive the guarantees for exact recovery of the SL methods under the standard Gaussian mixture model. In Section~\ref{sec:experiment}, we show some statistical and computational comparisons for the SL methods and the state-of-the-art  $K$-means algorithm in various settings.

\section{BACKGROUND}\label{sec:background}
We first provide some background on the $K$-means clustering. After that, we describe a matrix-lifting semidefinite relaxation scheme which turns the mixed integer program associated with the $K$-means into a convex one by throwing away the integer constraints, and review its theoretical properties.

\subsection{$K$-means clustering}
Let $X_1,\dots,X_n$ be a sequence of $p$-dimensional vectors and $X = (X_1, \dots, X_n) \in \bR^{p \times n}$ denote the data matrix with $n$ data points in $\bR^p$. Suppose that there is a clustering structure $G_1^*,\dots,G_K^*$ (i.e., a partition on $[n] := \{1,\dots,n\}$ such that $\bigsqcup_{k=1}^K G_k^* = [n]$, where $\bigsqcup$ denotes the disjoint union) on the $n$ data point indices. To recover the true clustering structure $G_1^*,\dots,G_K^*$ from data, we consider the $K$-means defined as the following constrained combinatorial optimization problem:
\begin{equation}\label{eqn:kmeans}
\begin{gathered}
\max_{G_{1},\dots,G_{K}} \sum_{k=1}^{K} {1 \over |G_{k}|} \sum_{i,j \in G_{k}} \langle X_{i}, X_{j} \rangle \\
\mbox{subject to } \bigsqcup_{k=1}^{K} G_{k} = [n],
\end{gathered}
\end{equation}
where $\langle X_i, X_j \rangle = X_i^\top X_j$ is the Euclidean inner product in $\bR^p$ that represents the similarity between two vectors $X_i$ and $X_j$. 
Note that the objective function in~\eqref{eqn:kmeans} is proportional to the log-likelihood function of cluster labels after profiling (maximizing) out the cluster centers as nuisance parameters in the Gaussian mixture model with constant and isotropic noise. Therefore, solving for~\eqref{eqn:kmeans} is equivalent to computing the maximum likelihood estimator.

The standard approach to finding an approximate solution of the $K$-means problem is the Lloyd's algorithm, also known as Voronoi iteration or $K$-means algorithm, which repeatedly finds the centroid of the points within each cluster $G_k$ and then re-assigns points to the $K$ clusters according to which of these centroids is closest. To overcome some shortcomings of the $K$-means algorithm that is unstable in both its running time and approximation accuracy, the $K$-means++ algorithm~\citep{K-means++} is proposed and becomes a state-of-the-art clustering algorithm hereafter. The $K$-means++ algorithm addresses the unstability issue by specifying a careful initialization procedure to seed the $K$-means algorithm. See the paper from~\cite{saxena2017review} for a recent review on different clustering techniques and their developments.

We end this subsection with a brief review on theoretical developments of the $K$-means clustering.
Consistency of the $K$-means estimation of the clustering centers is studied by~\cite{pollard1981}, without concerning the computational complexity. \cite{kumar2010clustering} and~\cite{awasthi2012improved}
show that if the true cluster centers are sufficiently well-separated relative to their spreads, then the Lloyd’s algorithm initialized by spectral clustering achieves exact recovery of the cluster labels. Partial recovery bounds of Lloyd's algorithm for local search solution to the $K$-means are derived by~\cite{LuZhou2016}. 


\subsection{$K$-means as mixed integer program}
Next, we describe an equivalent formulation of the $K$-means optimization~\eqref{eqn:kmeans} that will be useful in motivating its convex relaxation in the next subsection.
Note that we can express the cluster labels for each partition $G_{1},\dots,G_{K}$ of $[n]$ by their \emph{one-hot encoding}: we can associate each $(G_k)_{k=1}^K$ with a binary \emph{assignment matrix} $H = (h_{ik}) \in \{0, 1\}^{n \times K}$, where $h_{ij} = 1$ indicates $X_{i}$ belongs to cluster $k$, and $h_{ik} = 0$ otherwise. Because each row of $H$ contains exactly one non-zero entry, there is one-to-one mapping (up to assignment labeling) between the partition and the assignment matrix. Thus recovery of the true clustering structure is equivalent to recovery of the associated assignment matrix, and the $K$-mean clustering problem can be re-expressed as a (non-convex) mixed integer program:
\begin{equation}
\label{eqn:kmeans-mixed_integer}
\begin{gathered}
\max_{H} \langle A, H B H^\top \rangle \\
\mbox{subject to } H \in \{0,1\}^{n \times K}, \; H \vone_{K} = \vone_{n},
\end{gathered}
\end{equation}
where $A = X^\top X$ is the $n \times n$ similarity matrix and $\vone_n$ denotes the $n$-dimensional vector of all ones.

\subsection{SDP relaxed $K$-means}\label{Sec:SDP_rel}
Relaxing the above mixed integer program~\eqref{eqn:kmeans-mixed_integer} by changing variable $Z = H B H^\top$, we arrive at the SDP relaxed approximation of the $K$-means clustering problem:
\begin{equation}
\label{eqn:kmeans_sdp}
\begin{gathered}
\hat{Z} = \arg\max_{Z \in \bR^{n \times n}} \langle A, Z \rangle \\
\mbox{subject to } Z \succeq 0, \; \tr(Z) = K, \; Z \vone_{n} = \vone_{n}, \; Z \geq 0,
\end{gathered}
\end{equation}
where $Z \geq 0$ means each entry $Z_{ij} \geq 0$ and $Z \succeq 0$ means the matrix $Z$ is symmetric and positive semi-definite. This SDP approximation relaxes the integer constraint on $H$ into two linear constraints $\tr(Z) = K$ and $Z\geq 0$ that are satisfied by any $Z=HBH^T$ as $H$ ranges over feasible solutions of problem~\eqref{eqn:kmeans-mixed_integer}.
The SDP in~\eqref{eqn:kmeans_sdp} was first introduced by~\cite{PengWei2007_SIAMJOPTIM} and it was shown that this SDP relaxing the integer constraint is information-theoretically tight under the standard Gaussian mixture model~\citep{chen2021cutoff} (see our brief review below).

Membership matrix $Z^*$ corresponding to the true partition $G_1^*,\dots,G_K^*$ is a block diagonal matrix with $K$ blocks, each of which has size $n_k \times n_k$ with all entries equal to $n_k^{-1}$. Here $n_k = |G_k^*|$ is the size of cluster $k$. Note that the SDP solution $\hat{Z}$ of~\eqref{eqn:kmeans_sdp} is generally not integral in the sense that $\hat{Z}$ cannot be directly used to recover a partition estimate of the data points. In such case, we can apply some rounding technique to project $\hat{Z}$ back to yield a partition as the solution to the original discrete optimization problem~\eqref{eqn:kmeans}. For instance, we may cluster the top $K$ eigenvectors of $\hat{Z}$ to extract the estimated partition structure $\hat{G}_1,\dots,\hat{G}_K$. On the other hand, it is known that rounding is not necessary as the relaxed SDP solution can be directly used to recover the $K$-means in~\eqref{eqn:kmeans}, when the separation of cluster centers is large enough, a property often referred in literature as the \emph{hidden integrality}~\citep{FeiChen2018,chen2021cutoff,Ndaoud2018,Awasthi2015_ITCS}.

Formally, consider the standard GMM
where $n_k$ observations from the $k$-th cluster follow i.i.d.~$N(\mu_k,\sigma^2I_p)$ for $k\in[K]$. It is proved by~\cite{chen2021cutoff} that for any $\alpha>0$,
if the squared minimal separation $\Delta^2 = \min_{1 \leq k \neq l \leq K} \|\mu_k-\mu_l\|^2$ satisfies $\Delta^2\geq (1+\alpha) \bar\Delta_\ast^2$, where
\begin{equation}
\label{eqn:full_data_threshold}
\bar{\Delta}^2_\ast = 4 \sigma^2 \left( 1 + \sqrt{1+{p \over n_\ast \log n}} \right) \log{n},
\end{equation}
with $n_\ast =\min_{1\leq k\neq l\leq K} 2n_kn_l/(n_k+n_l)$ denoting the smallest pairwise harmonic average and $n=\sum_{k=1}^Kn_k$ the total sample size, then with probability at least $1-c_1K^2n^{-c_2\alpha}$ for some constants $c_1,c_2>0$, the SDP in~\eqref{eqn:kmeans_sdp} will produce the integral solution $Z^\ast$ that corresponds to exact recovery (cf.~Lemma~\ref{lem:SDP_bound_general} for a precise statement). Regarding the lower bound,~\cite{chen2021cutoff} shows that in the equal cluster size case where $n_k=n/K$ for each $k\in[K]$, if $\Delta^2\leq (1-\alpha) \bar\Delta_\ast^2$ holds for any $\alpha>0$, then with probability at least $1-cKn^{-1}$, no clustering algorithm can achieve simultaneous exact recovery of all cluster labels. In other words, $\bar{\Delta}^2_\ast$ is the cutoff value for exact recovery of GMM.

\section{PROPOSED LINEAR TIME APPROXIMATION ALGORITHMS}\label{sec:methods}
Now we introduce our proposed linear time complexity algorithm for approximating the SDP relaxed $K$-means problem. We also discuss some variants that significantly boost the numerical performance and are better suited to handle unequal cluster scenarios.

\subsection{Sketch-and-lift for the $K$-means SDP}
We first provide some intuition before formally describing our sketch-and-lift (SL) approach for the $K$-means SDP~\eqref{eqn:kmeans_sdp}. As in the Lloyd's algorithm, finding a best clustering scheme consists of two intermediate steps: 1.~estimate the center of each cluster; 2.~determine cluster labels based on which of these centers is closest. It turns out that the loss of statistical accuracy in estimating the centers in the first step due to using fewer but correctly labeled samples is much less severe than that due to using mislabeled samples. This motivates us to apply stable and reliable but computationally more expensive clustering algorithms such as the SDP~\eqref{eqn:kmeans_sdp} to a smaller subsample of size $m$ to extract correct cluster labels of the subsample. Based on the cluster labels, we obtain an estimator of the cluster centers (as within cluster averages) using the subsample, and then apply the estimated centers for clustering the entire data. As we will illustrate in the theoretical analysis, the sample size $m$ for estimating the centers via the $K$-means SDP~\eqref{eqn:kmeans_sdp} in the first step can be as small as $O\big((\log n)^2\big)$ (see the remark after Theorem~\ref{thm:SL_separation_upper_bound_bias_correction}) in order to guarantee the exact recovery of entire data cluster labels in the second step under suitable separation conditions. Due to this extremely low sample size requirement on $m$, the overall computational complexity will be dominated by the linear $O(n)$ complexity in the second step. Note that the subsampling in the first step corresponds to the ``sketch" operation; and the label recovery based on centers estimated from a subsample corresponds to the ``lift" operation. In principle, this SL idea can be incorporated with any accurate clustering method. We choose the SDP relaxed $K$-means in this paper mainly due to its theoretical optimality in cluster label recovery (cf.~Section~\ref{Sec:SDP_rel}).

Now we formally describe our SL approach.
Let $\gamma \in (0,1)$ be a pre-specified subsampling factor which may depend on the sample size $n$. We first randomly sample an index subset $T \subset [n]$ with i.i.d.~$\text{Ber}(\gamma)$. Denote the subsampled data matrix as $V = (X_i)_{i \in T}$, which is of size $p$-by-$m$ where $m = |T|$ follows a Binomial Bin$(n,\gamma)$ distribution. Here assuming the i.i.d.~sampling is mainly for technical convenience, and in practice one can also uniformly sample a subset of $[n]$ with fixed size $\lfloor n\gamma\rfloor$, where $\lfloor x\rfloor$ denotes the largest integer not exceeding $x$.
After the subsampling, we apply the SDP relaxed $K$-means~\eqref{eqn:kmeans_sdp} to $V$:
\begin{equation}
\label{eqn:kmeans_sketched_sdp}
\begin{gathered}
\hat{W} = \arg\max_{Z \in \bR^{m \times m}} \langle V^\top V, W \rangle \\
\mbox{subject to } W \succeq 0, \; \tr(W) = K, \; W \vone_{m} = \vone_{m}, \; W \geq 0.
\end{gathered}
\end{equation}
Once we obtain a partition estimate $\hat{R}_1,\dots,\hat{R}_K \subset T$ from $\hat{W}$ on the subset $V$ (perhaps after a rounding procedure), we compute the centroids $\bar{X}_k = |\hat{R}_k|^{-1} \sum_{j \in \hat{R}_k} X_j$ based on the estimated partition $T = \bigsqcup_{k=1}^K \hat{R}_k$. Finally, we project back the cluster labels to all data points in $X \setminus V$ by mapping them to the nearest centroid among $\bar{X}_1, \dots, \bar{X}_K$, i.e., for each $X_i \in X \setminus V$, we assign $i \in \hat{G}_k$ when $\|X_i - \bar{X}_k\| < \|X_i - \bar{X}_l\|$ for all $l \neq k$ and $l \in [K]$, where $\|\cdot\|$ denotes the $\ell_2$-norm. The SL algorithm is summarized in Algorithm~\ref{alg:SL} (steps 2-3 in subroutine Algorithm~\ref{alg:SDP} correspond to rounding).

\RestyleAlgo{boxruled}
\LinesNumbered
 \begin{algorithm}[h]\label{alg:SL}
 	 \DontPrintSemicolon
 	\KwInput{sampling weights $(w_1,\dots,w_n)$  with $w_1 = \dots = w_n = \gamma \in (0,1)$ being the subsampling factor.}
 	(Sketch) Independent sample an index subset $T \subset [n]$ via $\text{Ber}(w_i)$ and store the subsampled data matrix $V = (X_i)_{i \in T}$. \\
    Run subroutine Algorithm~\ref{alg:SDP} with input $V$ to get a partition estimate $\hat{R}_1,\dots,\hat{R}_K$ for $T$. \\ 
    Compute the centroids $\bar{X}_k = |\hat{R}_k|^{-1} \sum_{j \in \hat{R}_k} X_j$ for $k \in [K]$. \\
    (Lift) For each $i \in [n] \setminus T$, assign $i \in \hat{G}_k$ if\newline
        $\|X_i - \bar{X}_k\| < \|X_i - \bar{X}_l\|, \quad \forall l \neq k,\, l \in [K]$.\\
    \KwOutput{A partition estimate $\hat{G}_1,\dots,\hat{G}_K$ for $[n]$.}
 	\caption{Sketch-and-lift algorithm for $K$-means SDP with sampling weights $(w_1,\dots,w_n)$.} 
 \end{algorithm}
 
 \RestyleAlgo{boxruled}
\LinesNumbered
 \begin{algorithm}[h]\label{alg:SDP}
 	 \DontPrintSemicolon
 	\KwInput{Data matrix $V\in\mathbb R^{p\times m}$ containing $m$ points.}
    Solve the SDP in~\eqref{eqn:kmeans_sketched_sdp} using $V$ to get solution $\hat{W}$. \\
    Perform the spectral decomposition of $\hat{W}$ and take the top $K$ eigenvectors $(\hat{u}_1,\dots,\hat{u}_k)$. \\
    Run $K$-means clustering on $(\hat{u}_1,\dots,\hat{u}_k)$ and extract the cluster labels $\hat{R}_1,\dots,\hat{R}_K$ as a partition estimate for $[m]$. \\
    \KwOutput{A partition estimate $\hat{R}_1,\dots,\hat{R}_K$ for $[m]$.}
 	\caption{Subroutine for solving $K$-means SDP.} 
 \end{algorithm}
 
 We highlight that the SL approach has a \emph{linear} time complexity in the sample size $n$ if we choose $m = O(n^{c})$ for some small constant $c \in (0,1)$. Theoretically, it is shown in Section~\ref{sec:theory} that the subsampling factor $\gamma$ is allowed to vanish to zero while retaining statistical validity of SL. Obviously, any clustering algorithm has at least a linear time complexity since it should visit at least one time for each data point. On the other hand, it is shown that the SL enjoys a similar exact recovery threshold as the original SDP on all data points, which is known to achieve the information-theoretic limit~\citep{chen2021cutoff}. Empirically, we demonstrate in Section~\ref{sec:experiment} that around the sharp threshold of exact recovery, the SL approach statistically outperforms the widely used $K$-means++ algorithm~\citep{K-means++} by a large margin in terms of the error rates.

\begin{rem}[Multi-epoch with averaging]
We can repeat the above SL procedure for multiple epochs to enhance the empirical performance. A simple way to achieve this is to randomly partition the data points into $\lfloor n/m \rfloor$ blocks, each of which is a sequence of independent Bernoulli trials of size $n$ with success probability $\gamma$. Then we run $\lfloor n/m \rfloor$ SL procedure in Algorithm~\ref{alg:SL} on the independent data blocks and average the centroids estimated from the multiple epochs before lifting. Such procedures can be easily paralleled in a distributed system and therefore the computational burden for running multiple epochs is essentially the same as one sketch-and-lift pass. In Section~\ref{sec:experiment}, we present some numerical result for the multi-epoch SL with averaging. This multi-epoch SL approach is summarized in Algorithm~\ref{alg:ME_SL} below.\qed
\end{rem}

\RestyleAlgo{boxruled}
\LinesNumbered
 \begin{algorithm}[h]\label{alg:ME_SL}
 	 \DontPrintSemicolon
 	\KwInput{Subsample size $m$.}
 	Randomly partition data indices $[n]$ into $S=\lfloor n/m\rfloor$ blocks $T_1,\ldots,T_S$ with size $m$.\\
 	\For{$s=1,\ldots,S$} {(Sketch) 
    Run subroutine Algorithm~\ref{alg:SDP} with input $V_s = (X_i)_{i \in T_s}$ to get a partition estimate $\hat{R}_{s,1},\dots,\hat{R}_{s,K}$ for $T$. \\
    Compute the centroids $\bar{X}_{s,k} = |\hat{R}_{s,k}|^{-1} \sum_{j\in \hat{R}_{s,k}} X_j$ for $k \in [K]$.}
    Compute the aggregated centroids $\bar{X}_{k} = S^{-1} \sum_{s=1}^S \bar{X}_{s,k}$ for $k \in [K]$.\\
    (Lift) For each $i \in [n] \setminus T$, assign $i \in \hat{G}_k$ if\newline
        $\|X_i - \bar{X}_k\| < \|X_i - \bar{X}_l\|, \quad \forall l \neq k,\, l \in [K]$.\\
    \KwOutput{A partition estimate $\hat{G}_1,\dots,\hat{G}_K$ for $[n]$.}
 	\caption{Multi-epoch sketch-and-lift algorithm for $K$-means SDP.} 
 \end{algorithm}

\begin{rem}[Related work on stochastic block models]
\cite{mixon2020sketching} proposed a subsampled SDP for the two-component stochastic block model (SBM) with equal community size. The approach presented by~\cite{mixon2020sketching} first randomly subsamples a small vertex set according a Bernoulli process with rate $\gamma \in (0,1)$ and then solves the community detection problem on the induced subgraph. The solution on the subgraph is finally projected by a majority voting procedure to all nodes in the whole graph. It is shown by~\cite{mixon2020sketching,abdalla2021community} that the subsampling factor $\gamma > c$, where $c > 0$ is a constant depending on the edge connecting probabilities within-community and between-communities in the graph, is needed for exact community recovery with high probability. In our clustering problem, we allow the subsampling ratio $\gamma = o(1)$ (cf.~Theorem~\ref{thm:SL_separation_upper_bound} below), so the computational cost can be much further reduced than the subsampled SDP for the SBM. In particular, we can choose very small $\gamma$ such that the reduced SDP problem size $m=\lfloor n \gamma \rfloor$ is nearly independent of $n$ (up to some polylogarithmic factor $\log^c{n}$). Moreover, the subsampled SDP for SBM proposed by~\cite{mixon2020sketching} works only for two-component equal cluster size case, while our SL approach works for unbalanced $K$-component clusters and it can be further enhanced with bias-correction (Section~\ref{subsec:BCSL}) and non-uniform sampling weights (Section~\ref{subsec:WSL}) to better handle the general unequal cluster size scenario. \qed
\end{rem}

The SL approach performs the uniform subsampling on $n$ data points, which is natural for equal cluster size case. If the cluster sizes are not equal, the estimated centroids based on the partition given by the sketched SDP in~\eqref{eqn:kmeans_sketched_sdp} have different variances. Thus by comparing the distances between data point in $X \setminus V$ with $\bar{X}_k$ and $\bar{X}_l$ will create a larger bias than that in the equal cluster case. Theoretically, such an extra bias term will imposes the unnecessary constraint of $p=O\big((\gamma n/K)^2\big)$ (cf.~Theorem~\ref{thm:SL_separation_upper_bound}).
To mitigate this issue, we propose two procedures in the following subsections.

\subsection{Bias-corrected sketch-and-lift (BCSL)}
\label{subsec:BCSL}
Suppose we have obtained a partition $\hat{R}_1,\dots,\hat{R}_K$ for $V$ by the sketched SDP and an estimate of the cluster centers $\bar{X}_k$ and $\bar{X}_l$. Let $\underline{m} := \min_{k \in [K]} |\hat{R}_k|$ be the smallest cluster size. To fairly compare the distances $\|X_i-\bar{X}_k\|$ and $\|X_i-\bar{X}_l\|$ by matching the variance of all cluster center estimates $\{\bar{X}_k:\,k\in[K]\}$, we further down-sample $\hat{R}_1,\dots,\hat{R}_K$ to have the same size $\underline{m}$. In particular, we can randomly sample a subset $\tilde{R}_k$ with equal size $\underline{m}$ from each $\hat{R}_k$. Then we apply the lift step to propagate $\tilde{R}_1,\dots,\tilde{R}_K$ to the original data to obtain a partition $\hat{G}_1,\dots,\hat{G}_K$. As we will show in Theorem~\ref{thm:SL_separation_upper_bound_bias_correction}, this bias correction scheme removes the undesirable constraint on $p$ as needed in the SL approach.
The bias-corrected SL (BCSL) algorithm is summarized in Algorithm~\ref{alg:BCSL}.

\RestyleAlgo{boxruled}
\LinesNumbered
 \begin{algorithm}[h]\label{alg:BCSL}
  \DontPrintSemicolon
 	\KwInput{subsampling factor $\gamma \in (0,1)$.}
    Independently sample an index subset $T \subset [n]$ via $\text{Ber}(\gamma)$ and make the subsampled data matrix $V = (X_i)_{i \in T}$. \\
    Run subroutine Algorithm~\ref{alg:SDP} with input $V$ to get a partition estimate $\hat{R}_1,\dots,\hat{R}_K$ for $T$.\\
    For each $\hat{R}_k$, randomly sample a subset $\tilde{R}_k$ with equal size $\underline{m}$. \\
    Compute the centroids $\bar{X}_k = |\tilde{R}_k|^{-1} \sum_{j \in \tilde{R}_k} X_j$ for $k \in [K]$. \\
    For each $i \in [n] \setminus T$, assign $i \in \hat{G}_k$ if\newline
        $\|X_i - \bar{X}_k\| < \|X_i - \bar{X}_l\|, \quad \forall l \neq k,\, l \in [K]$.\\
    \KwOutput{A partition estimate $\hat{G}_1,\dots,\hat{G}_K$ for $[n]$.}
 	\caption{Bias-corrected sketch-and-lift algorithm for $K$-means SDP.} 
 \end{algorithm}
 
\subsection{Weighted sketch-and-lift (WSL)}
\label{subsec:WSL} 
Another bias correcting method is to subsample $X_1,\dots,X_n$ with non-uniform weights that convey the cluster size information, so that in the sketched data matrix $V$, all clusters have roughly the same number of points. For example, we increase (decrease) the sampling weights for those points from small (large) clusters. Compared to the BCSL approach, this weighted SL (WSL) approach has no waste of information when estimating the cluster centers based on partition centroids, given we know the \emph{ideal} sampling weights.
However, the WSL appears to incur a chicken and egg problem as the 
ideal sampling weights requires knowledge on the cluster membership of each data point. Fortunately, as we will discuss in Remark~\ref{rem:mr_WSL}, a multi-round extension of the WSL which iteratively applies the WSL to refine the sampling weights based on the clustering labels in the previous round has surprisingly good numerical performance in that the recovery error decays as the round increases (cf.~Figure~\ref{fig:multi-round_WSL16} in the supplement). 
Now let us formally describe the WSL approach.
For each data point $X_i$ for $i \in G_k^*$, we denote $w_i^* =\gamma n/(Kn_k)$ as the ideal sketch weight for $X_i$. 
Suppose in practice we have a set of approximating weights $w_i \in [0,1]$ such that there exists a subset $D \subset [n]$ which satisfies for some $(\epsilon, \delta) \in [0,\infty) \times [0,1]$
\begin{equation}
\label{eqn:sampling_weights_condition}
|D| \geq (1-\delta)\, n\ \ \mbox{and} \ \ \left| {w_i \over w_i^*} - 1 \right| \leq \epsilon.
\end{equation}
Condition~\eqref{eqn:sampling_weights_condition} requires the at least $(1-\delta)$ proportion of constructed sampling weights should be close to the true sampling weights with at most $\epsilon$ distortion. We call such weights \emph{a set of $(\epsilon, \delta)$-weights}. Ideal weights are $(0,0)$-weights. In practice, a priori estimate for the weights $w_i$ can be set through Lloyd's algorithm for the $K$-means. And the parameter $\gamma$ can be chosen as small as $O(\log (n)/n),$ which implicitly shows that the weights $w_i$'s would be as small as $o(1).$  The rest of the WSL is the same as Algorithm~\ref{alg:SL} with a general non-uniform sampling weighs $(w_1,\dots,w_n)$. In addition, we can also combine the BCSL with the WSL to enforce the equal cluster sizes when computing the centroids $\bar X_k$'s.
The WSL algorithm is summarized in Algorithm~\ref{alg:WSL}.

\RestyleAlgo{boxruled}
\LinesNumbered
 \begin{algorithm}[h]\label{alg:WSL}
  \DontPrintSemicolon
 	\KwInput{subsampling factor $\gamma \in (0,1)$.} 
    Run Lloyd's algorithm to obtain an initial partition estimate $\tilde G_1,\ldots,\tilde G_K$ for $[n]$, with sizes $\tilde n_1,\ldots,\tilde n_K$. \\
    For each $i\in[n]$, set $w_i = \gamma n/(K\tilde n_k)$ if $i\in \tilde G_k$.\\
    Run Algorithm~\ref{alg:SL} with weights $(w_1,\ldots,w_n)$.\\
    \KwOutput{A partition estimate $\hat{G}_1,\dots,\hat{G}_K$ for $[n]$.}
 	\caption{Weighted sketch-and-lift algorithm for $K$-means SDP.} 
 \end{algorithm}

\begin{rem}[Multi-round WSL]\label{rem:mr_WSL}
The priori estimate for the weights (e.g., by Lloyd's or $K$-means++ algorithm) can be viewed as a warm start of WSL. To further boost the performance of the WSL, we can iteratively apply WSL to refine the recovered cluster labels, which is summarized in Algorithm~\ref{alg:MR_WSL} below. In Section~\ref{sec:experiment}, we present some superior numerical result for the multi-round WSL.
\qed
\end{rem}

\RestyleAlgo{boxruled}
\LinesNumbered
 \begin{algorithm}[h]\label{alg:MR_WSL}
 \DontPrintSemicolon
 	\KwInput{subsampling factor $\gamma \in (0,1)$ and number of rounds $R$.}
 	Run Algorithm~\ref{alg:WSL} to get partition estimate $\tilde{G}_1,\dots,\tilde{G}_K$ for $[n]$.\\
    \For{$r=2,\ldots,R$}{
     For each $i\in[n]$, update sampling weight $w_i = \gamma n/(K\tilde n_k)$ if $i\in \tilde G_k$, where $\tilde n_k =|\tilde G_k|$.\\
      Run Algorithm~\ref{alg:SL} with weights $(w_1,\ldots,w_n)$ to update $\tilde{G}_1,\dots,\tilde{G}_K$.
    }
    \KwOutput{A partition estimate $\tilde{G}_1,\dots,\tilde{G}_K$ for $[n]$.}
 	\caption{Multi-round weighted sketch-and-lift algorithm for $K$-means SDP.} 
 \end{algorithm}

\section{EXACT RECOVERY GUARANTEES}\label{sec:theory}
To study the theoretic properties of SL, we follow the literature by using the standard GMM as our working model. Specifically, we assume $X_1,\dots,X_n$ to be from the following data generating model: if $i \in G_k^*$, then
\begin{equation}\label{eqn:GMM}
X_i = \mu_k + \epsilon_i,
\end{equation}
where $\mu_1,\dots,\mu_K \in \bR^p$ are the (unobserved) cluster centers and $\epsilon_i$ are i.i.d.~$N(0,\sigma^2 I_p)$ noise. Recall that $\Delta^2 = \min_{1 \leq k \neq l \leq K} \|\mu_k-\mu_l\|^2$ denotes the squared minimal separation between cluster centers.

\begin{thm}[Separation bound for exact recovery by SL]
\label{thm:SL_separation_upper_bound}
Let $\alpha > 0$ and $\gamma \in (0,1)$ be the subsampling ratio. Suppose that $n_1=\cdots=n_K=n/K$. If $\Delta^2 \geq (1+\alpha) \bar{\Delta}^2_\gamma$, where
\begin{equation}
\label{eqn:threshold}
\bar{\Delta}^2_\gamma = 4 \sigma^2 \left( 1 + \sqrt{1+{K p \over \gamma n \log n}} \right) \log{n},
\end{equation}
then the output $\hat{G}_1,\dots,\hat{G}_K$ from the SL Algorithm~\ref{alg:SDP} for the $K$-means SDP achieves exact recovery, i.e., $\hat{G}_k = G_k^*$ for all $k \in [K]$ with probability at least $1- C_1 (\log(\gamma n))^{-C_2}$, provided that $K \leq C_3 {\log(\gamma n) \over \log\log(\gamma n)} $ and $p \leq C_4(\gamma n/K)^{2}$, where $C_i, i=1,2,3,4$ are constants depending only on $\alpha$.
\end{thm}

Theorem~\ref{thm:SL_separation_upper_bound} considers the equal cluster case, and includes the exact recovery property for the original SDP $K$-means~\eqref{eqn:kmeans_sdp} as a special case when $\gamma=1$ (and no lift step is needed). Compared to the separation cutoff value~\eqref{eqn:full_data_threshold} for exact recovery of the entire data, 
the separation requirement in~\eqref{eqn:threshold} has an extra factor of $\gamma^{-1}$ inside the square root---the larger $Kp/(\gamma n\log n)$ term corresponds to the statistical fluctuation of using only $\gamma n/K$ samples to estimate the $p$-dimensional cluster centers instead of $n/K$ samples in the entire data, which appears to be inevitable for any single-epoch SL method. Interestingly, as we empirically observed in Section~\ref{sec:experiment}, the multi-epoch SL method summarized in Algorithm~\ref{alg:ME_SL} has a noticeable improvement over the SL and appears to attain the optimal cutoff value~\eqref{eqn:full_data_threshold} due to the usage of almost all data in estimating the cluster centers (by averaging across subsamples). Based on the numeric evidence, we pose the following conjecture as a future study.

\begin{conj}\label{conj:MESL}
Multi-epoch SL method with averaging (Algorithm~\ref{alg:ME_SL}) attains the information-theoretic threshold $\bar{\Delta}^2_\ast$ in~\eqref{eqn:full_data_threshold} as the SDP~\eqref{eqn:kmeans_sdp} on the entire data.
\end{conj}

For general subsampling ratio $\gamma \in (0,1)$, we note that there is an additional constraint $p\lesssim (\gamma n / K)^2$ to ensure the exact recovery. This constraint can be shown even stricter $p\lesssim (\gamma n / K)$ for unequal cluster size case. This condition comes from the fact that when lift is needed to obtain the full cluster labels on all data points, we need to ensure that the estimated cluster sizes from the subsampled SDP~\eqref{eqn:kmeans_sketched_sdp} are approximately equal to match the variances. In contrast, we shall show that in Theorems~\ref{thm:SL_separation_upper_bound_bias_correction} and~\ref{thm:WSL_separation_upper_bound} below that the BCSL does not require this condition, and the WSL still requires this condition $p\lesssim (\gamma n / K)^2$, but both will work for the unequal cluster size case as well.
\smallskip

\begin{thm}[Separation bound for exact recovery by BCSL]
\label{thm:SL_separation_upper_bound_bias_correction}
Let $\alpha > 0$, $\gamma \in (0,1)$ be the subsampling ratio and $\underline n=\min_{k\in[K]}n_k$. If $\Delta^2 \geq (1+\alpha) \bar{\Delta}^{'2}_\gamma$, where
\begin{equation}
\label{eqn:threshold_2}
\bar{\Delta}^{'2}_\gamma = 4 \sigma^2 \left( 1 + \sqrt{1+\frac{p}{\gamma \underline n\log n}} \right) \log{n},
\end{equation}
then the output $\hat{G}_1,\dots,\hat{G}_K$ from the BCSL Algorithm~\ref{alg:BCSL} achieves exact recovery with probability at least $1- C_1(\log\gamma n)^{-C_2}$,
provided that  $K \leq C_3 {\log(\gamma n) \over \log\log(\gamma n)}$, $\underline n\ge C_4 n/\log(\gamma n),$ $\log{n}/ \underline n \le C_5\gamma$, where  $  C_i, \ i=1,2,3,4,5$ are only depend on $\alpha$.
\end{thm}

According to Theorem~\ref{thm:SL_separation_upper_bound_bias_correction}, the subsampling factor $\gamma$ can be as small as $O\big((\log n)^2/n\big)$, corresponding to a minimal subsample size $m=O((\log n)^2)$. Consequently, the overall time complexity is dominated by the $O(n)$ complexity of the lift step.
\medskip

\begin{thm}[Separation bound for exact recovery by WSL]
\label{thm:WSL_separation_upper_bound}
Suppose we have a set of $(\epsilon, \delta)$-weights $w_i \in [0,1]$ satisfying ~\eqref{eqn:sampling_weights_condition}. Let $\alpha > 0$. If $\Delta^2 \geq (1+\alpha)\bar{\Delta}_\gamma^2$, where $\bar{\Delta}_\gamma^2$ is defined in~\eqref{eqn:threshold},
then the WSL Algorithm~\ref{alg:WSL} achieves exact recovery with probability at least 
$1- C_1(\log\gamma n)^{-C_2}$, provided that 
\begin{align*}
&K\leq C_3(\log\gamma n)/(\log\log \gamma n),\ \ p\leq C_4(\gamma n/K)^{2},\\
&\delta\leq C_5(\underline n/n)\min\big\{1,\sqrt{\gamma n/p}\,\big\}, \\
&\epsilon \leq C_6\min\big\{1, \gamma n \log n/(Kp)\big\},
\end{align*}
where constants $C_i$, $i=1,\ldots,6$, only depend on $\alpha$.
\end{thm}

We remark that the WSL by adjusting the bias with adaptive (non-uniform) weights essentially reduces the unequal sizes case to the equal size case. The cost of choosing the adaptive weights is that we need impose size conditions on $(\epsilon, \delta)$ (e.g., $\epsilon, \delta =o(p^{-c})$ for some $c>0$) such that they can absorb the effect coming from the growth of $p$. In Section~\ref{sec:more_simulation_results} in the supplement, we evaluate the effect of initial weights by $K$-means++ algorithm. In addition, from the numerical results, we conjecture that the multi-round WSL summarized in Algorithm~\ref{alg:MR_WSL}
can further relax the conditions to achieve exact recovery, for example, by throwing away the $p\leq C_4(\gamma n/K)^{2}$ constraint. We leave its formal theoretical study to a future direction.

\section{NUMERICAL EXPERIMENTS}
\label{sec:experiment}

In this section, we test the numerical performance for the SL method and its variants, and compare them with the $K$-means++ algorithm on synthetic data. MATLAB code implementing the SL approach and its variants are available at: \url{https://github.com/Yubo02/Sketch-and-Lift-Scalable-Subsampled-Semidefinite-Program-for-K-means-Clustering}

\subsection{Setup}
We generate data from a $4$-component Gaussian mixture model~\eqref{eqn:GMM} parametrized by $(p, n, \lambda^*)$, where parameter $\lambda^*>0$
characterizes the cluster center separation through $\Delta^2 = (\lambda^\ast \bar\Delta_\ast)^2$ and recall that $\bar\Delta_\ast^2$ is the theoretical cutoff in~\eqref{eqn:full_data_threshold}. We compare the following clustering methods.
\begin{itemize}[noitemsep,nolistsep]
    \item $M_0$ is the Matlab build-in $K$-means clustering  implementation (default algorithm is $K$-means++).
    \item $M_1$ is the sketch-and-lift (SL) method described in Algorithm~\ref{alg:SL}.
    \item $M_2$ is the bias-corrected sketch-and-lift (BCSL) method described in Algorithm~\ref{alg:BCSL}.
    \item $M_3$ is the weighted sketch-and-lift (WSL) method described in Algorithm~\ref{alg:WSL}.
    \item $M_4$ is the multi-epoch sketch-and-lift (ME-SL) with averaging described in Algorithm~\ref{alg:ME_SL}.
    \item $M_5$ is the multi-round weighted sketch-and-lift (MR-WSL) method described in Algorithm~\ref{alg:MR_WSL} with output at the $4$-th round.
\end{itemize}
For the SL methods (M1-M5), we vary the subsampling factor $\gamma$. 
We choose round number as $4$ in M5 since according to the additional numerical results reported in the supplement, the MR-WSL typically reaches its best performance after $3$-$4$ rounds.
We also compare with the original SDP relaxed $K$-means method~\eqref{eqn:kmeans_sdp} (method O) on the entire data points whenever it is feasible to run (in our case when $n \leq 3000$). We report the error rate in recovering cluster labels and the running time for these algorithms averaged over 100 simulations.

\subsection{Results}\label{subsec:experiment_result}

Due to the space limit, we report simulation results for equal cluster size case in this subsection. For complete numerical experiment results including the unequal cluster size case, we refer to Section~\ref{sec:more_simulation_results} in the supplementary material.

\begin{figure}[h!] 
   \centering
      \subfigure{\includegraphics[trim={1.5cm 7cm 1.5cm 7cm},clip,scale=0.38]{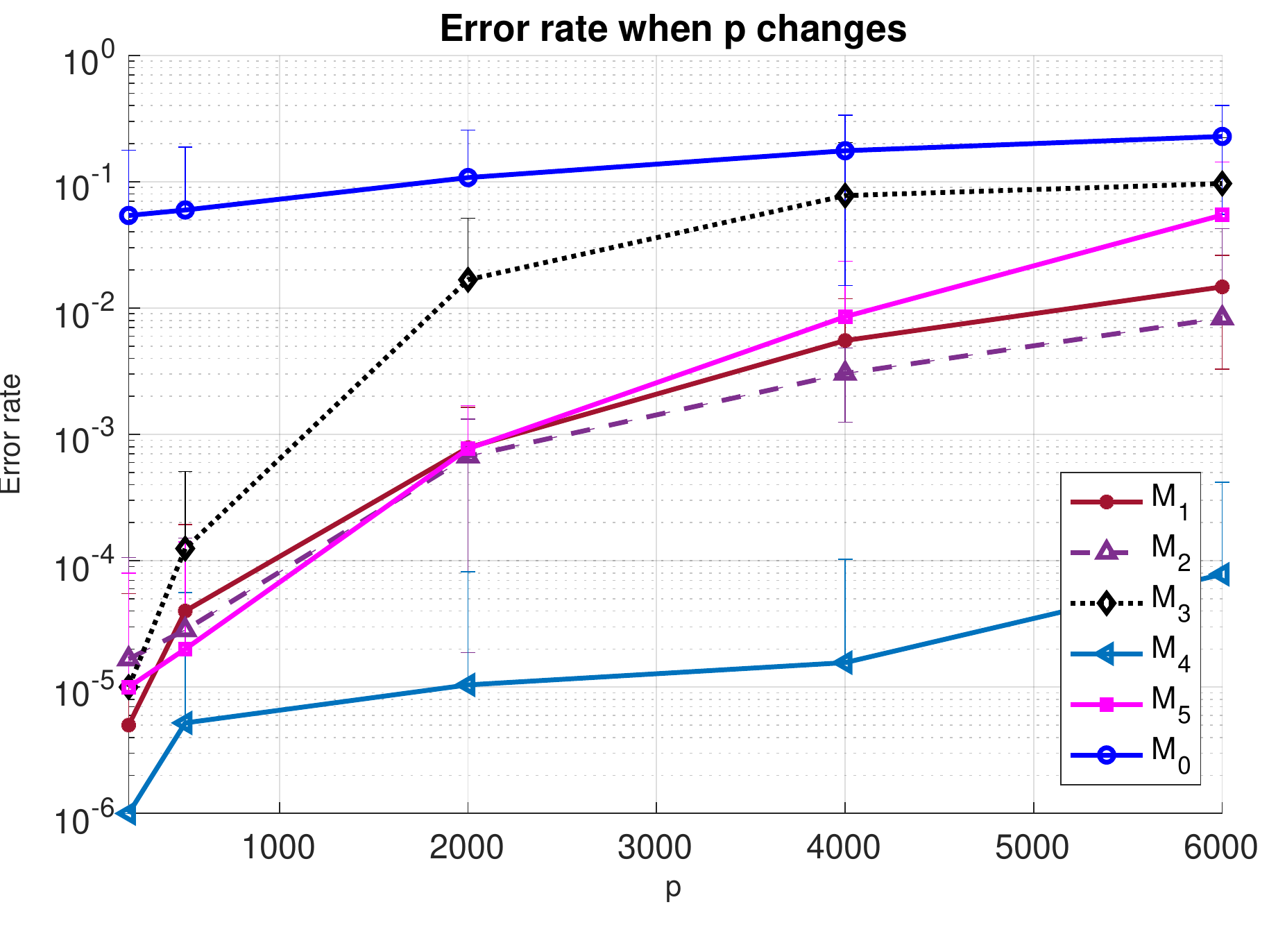}} \\[-1.5ex]
      \subfigure{\includegraphics[trim={1.5cm 7cm 1.5cm 7cm},clip,scale=0.38]{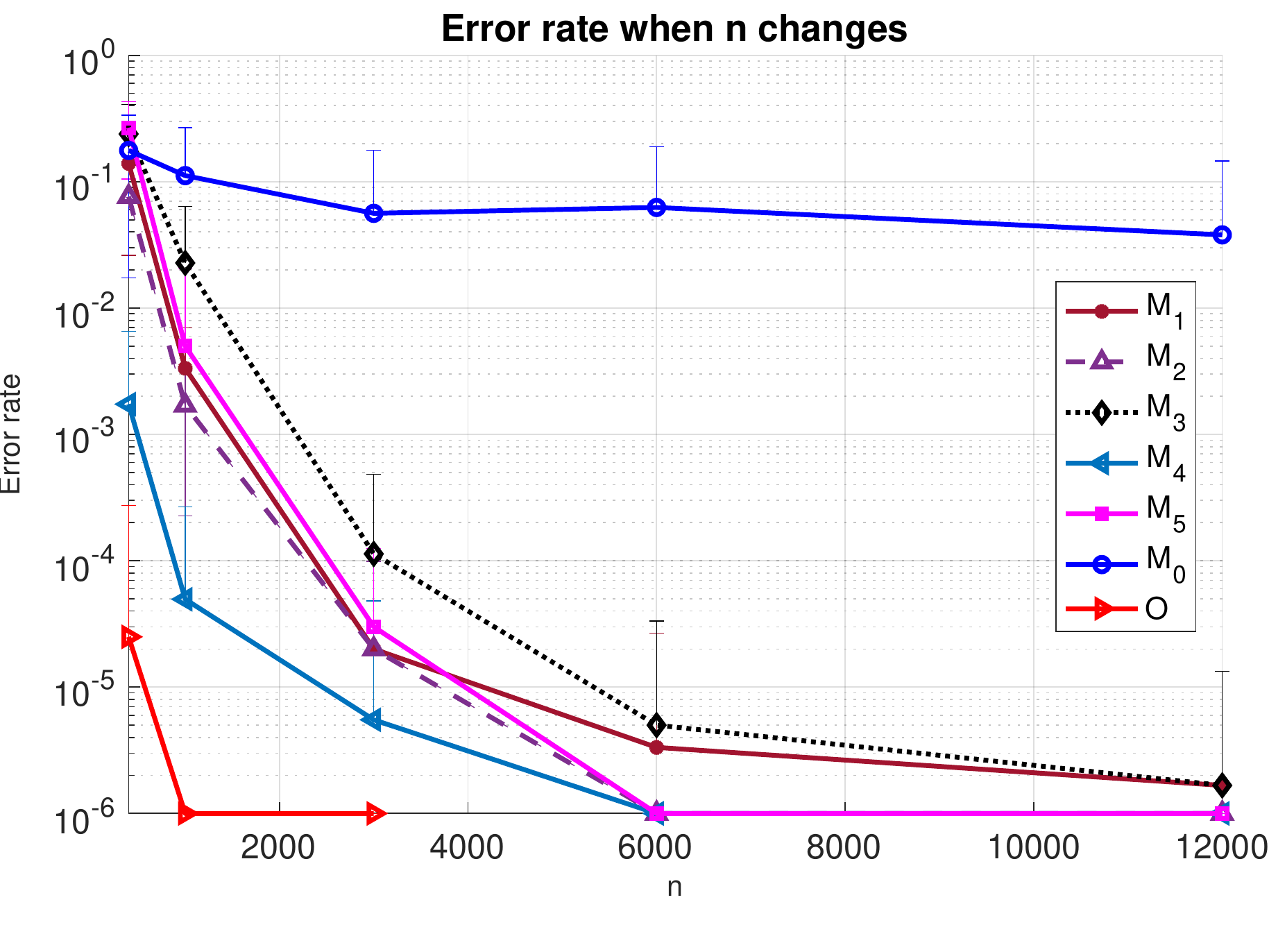}} \\[-1.5ex]
      \subfigure{\includegraphics[trim={1.5cm 7cm 1.5cm 7cm},clip,scale=0.38]{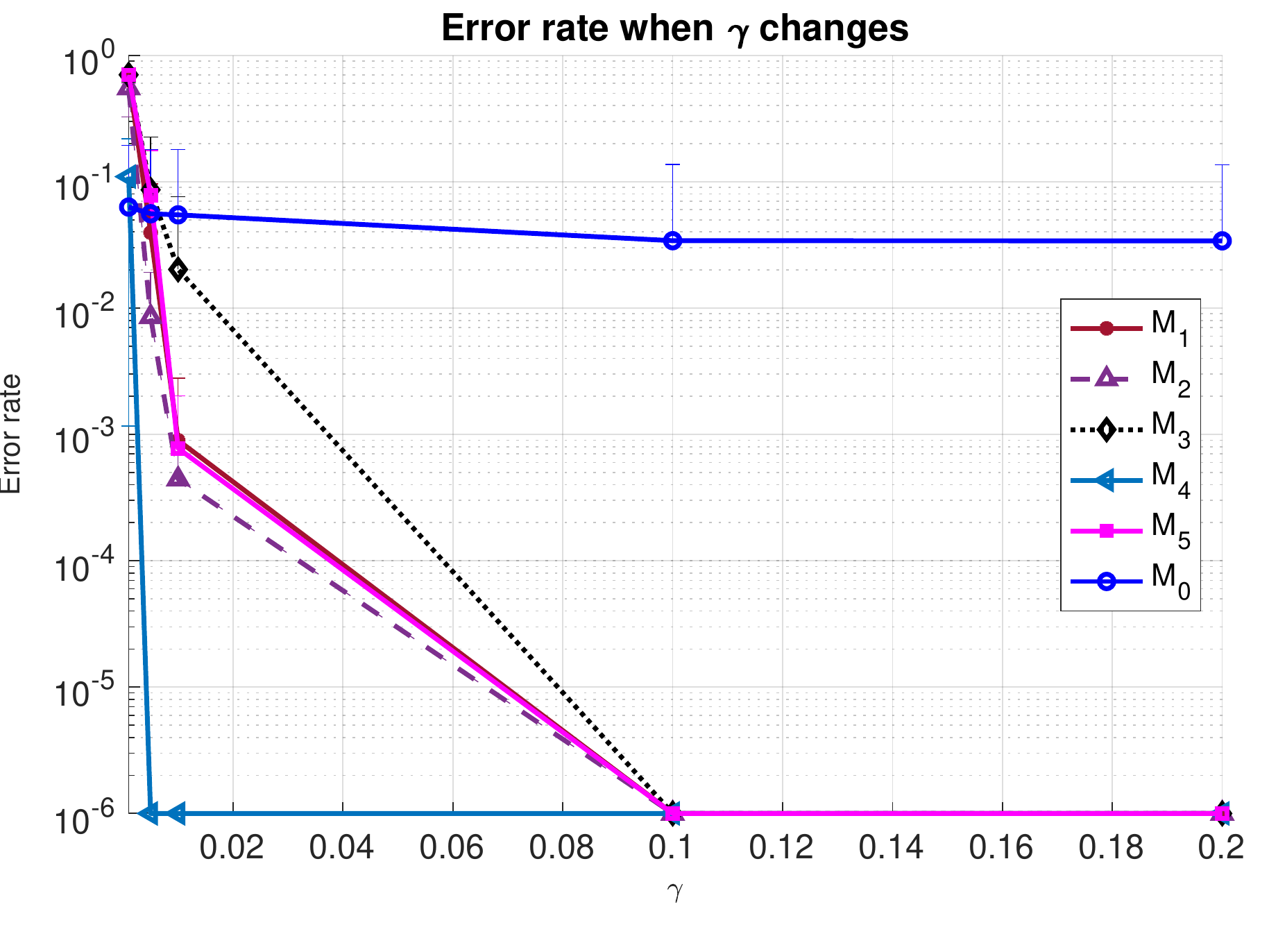}} \\[-1.5ex]
      \subfigure{\includegraphics[trim={1.5cm 7cm 1.5cm 7cm},clip,scale=0.38]{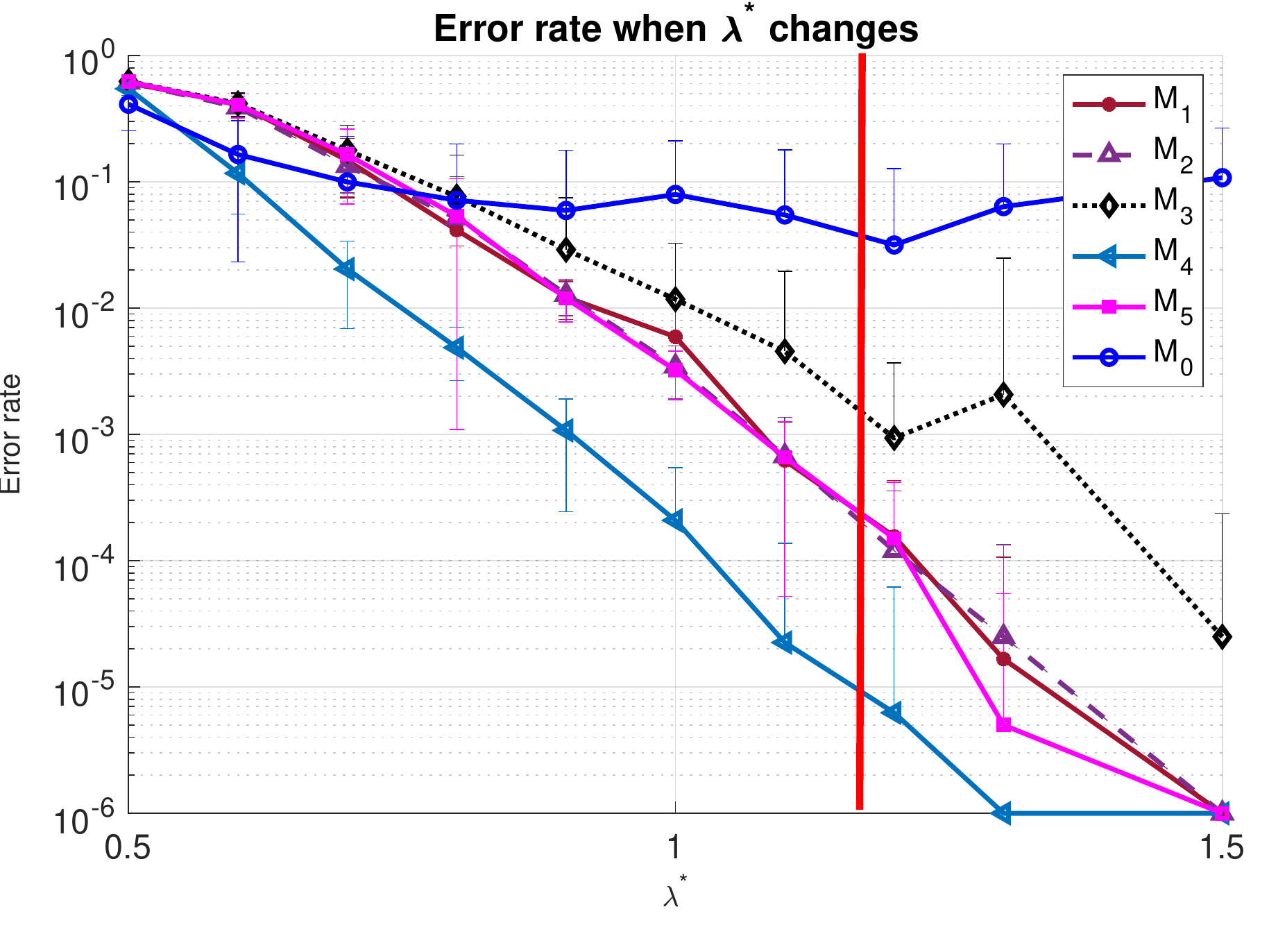}} \\[-2.5ex]
   \caption{Log-scale error rates (with error bars) when one parameter varies. Zero error is displayed as $10^{-6}$ in the log-scale plot. Red vertical line in the lowest plot indicates theoretical threshold $\bar\Delta_\gamma^2$ for SL methods.}
   \label{fig:error_rates}
\end{figure}

\begin{figure}[h!] 
   \centering
      \subfigure{\includegraphics[trim={1.5cm 7cm 1.5cm 7cm},clip,scale=0.38]{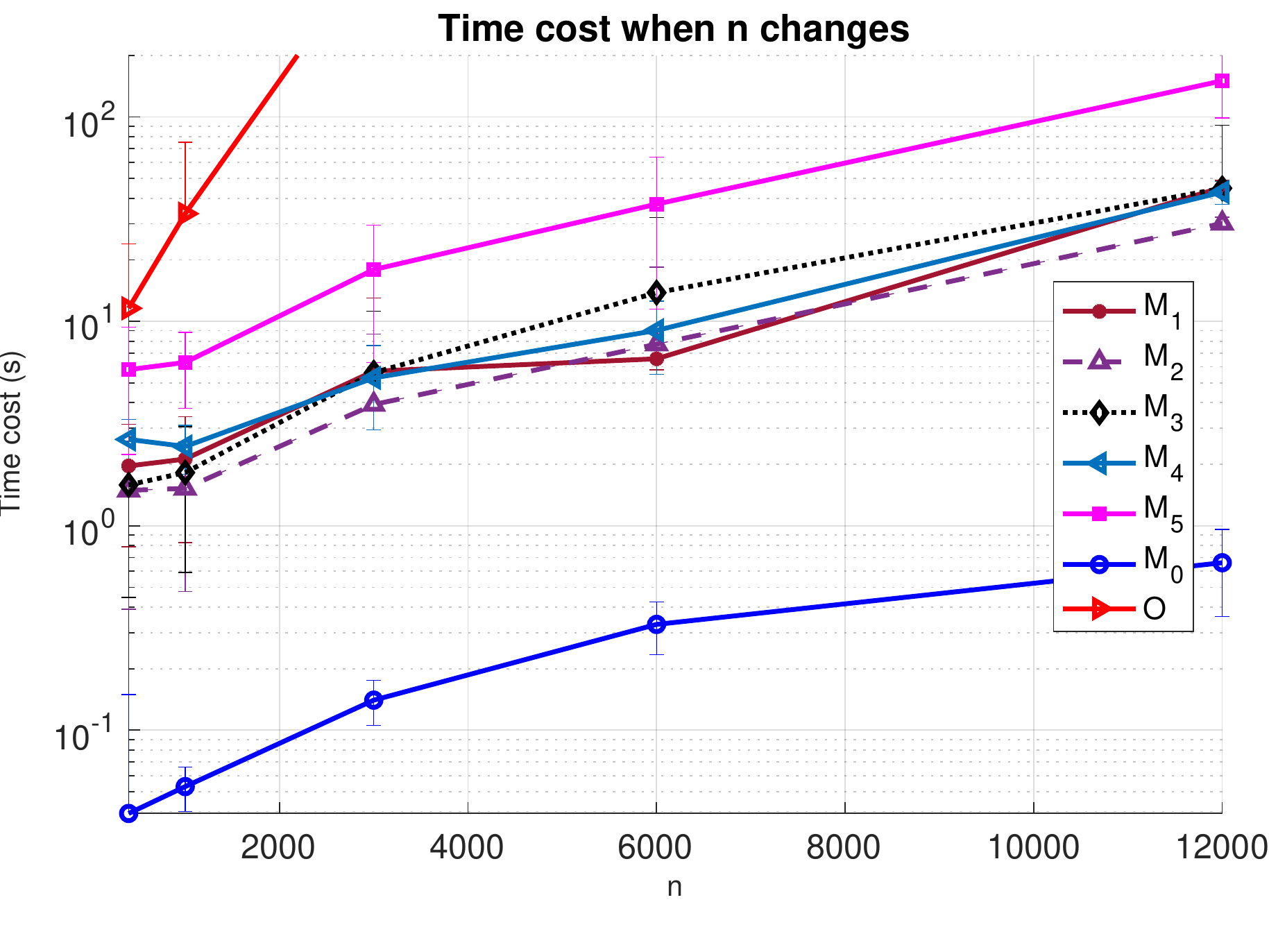}} \\[-3ex]
   \caption{Log-scale runtime (with error bars) v.s.~$n$.}
   \label{fig:run_time}
\end{figure}

The baseline setup is $p=1000$,  $n=2000$, $\gamma=0.1$ and $\lambda^{*}=1.2$, expect when $\gamma$ is changing, we use $n=10000$. In each simulation setting, we vary one parameter and report the error rate, which is summarized in Figure~\ref{fig:error_rates}. Figure~\ref{fig:run_time} compares the runtimes as $n$ changes.
We observe that all SL methods have \emph{significantly and uniformly smaller} error rate than the state-of-the-art $K$-means++ method (blue solid curve) in nearly all setups. We also note that runtime curves of the SL-methods on the log-scale are \emph{parallel} to the $K$-means++ algorithm, indicating that SL methods have the same linear $O(n)$ complexity as the fast $K$-means++ (with difference only occurring in the leading constant). In comparison, the original SDP (O) has super linear complexity as expected.

One interesting observation is that the multi-epoch SL method (M4) is almost always the best method across all four settings in Figure~\ref{fig:error_rates}, and has comparable performance as the original SDP method (O) in the range when it is feasible to run. Note that with sample size $n=2000$ as the baseline, only mis-specifying one cluster label in $1$ out of $100$ replicates will incur an error rate of $5\times 10^{-6}$, meaning that an average error rate (over $100$ replicates) of order below (or around) $10^{-5}$ can be viewed as perfect clustering (due to the log-scale, we display $10^{-6}$ when the actual error is zero).
This empirical observation provides the numerical evidence for supporting Conjecture~\ref{conj:MESL} about the information-theoretic optimality of the multi-epoch SL.

We also report the error decay for the MR-WSL with $K$-means++ algorithm as the warm start in Section~\ref{sec:more_simulation_results} in the supplement, where we observed that the MR-WSL has a surprisingly good recovery performance after 3-4 rounds. In particular, from Figure~\ref{fig:error_rates} we can see that 
the MR-WSL (M5) is the second best method in most settings (with the best being M4), and has significant improvements over its single round counterpart WSL (M3) due to progressive refinements on the estimated sampling weights (cf.~Section~\ref{sec:more_simulation_results}).

The plot at the very bottom in Figure~\ref{fig:error_rates} shows that the error rates as we change the separation parameter $\Delta^2=\min_{1\leq k\neq l\leq K}\|\mu_k-\mu_l\|^2$, where the red vertical line indicates the theoretical threshold $\bar\Delta_\gamma^2$ for SL methods given in Theorem~\ref{thm:SL_separation_upper_bound}. Since error of order $10^{-5}$ is very close to perfect clustering, the numerical results in the plot are consistent with our theory that $\bar\Delta_\gamma^2$ characterizes the cutoff value for SL methods.

We also assess the impact of initialization (i.e., warm start effect) of WSL by $K$-means by looking at the estimated $(\epsilon, \delta)$ parameters (Table \ref{tab:initial_weights} for the 1-st round in Section~\ref{sec:more_simulation_results}). In particular, we find that for fixed $\epsilon=0.2$, $\delta=0.25373$, $0.25598$, $0.37208$, $0.33282$, $0.38929$ for $p=200$, 500, 2000, 4000, 6000, respectively. Thus for increasing $p$ corresponding to more difficult clustering problems, the quality of initial weights deteriorates by the $K$-means. Still, our WLS (in particular, the MR-WLS refinement) maintains good quality of cluster label recovery. And $\delta=0.000005$, $0.00006$, $0.028845$, $0.088965$, $0.15886$, respectively when we perform the 2-nd round of WSL (Table \ref{tab:initial_weights2} in Section~\ref{sec:more_simulation_results}). Finally we can see that $\delta=0$ uniformly for the 4-th round WSL (Table \ref{tab:initial_weights4} in Section~\ref{sec:more_simulation_results}). This shows that the multi-round WSL refines the clustering errors and can eventually achieve the exact recovery as if we initialize with the ideal weights.

Finally, we applied our method to two benchmark datasets. The first one considers 32-dimensional mass cytometry (CyTOF) dataset, consisting of protein expression levels of healthy human bone marrow mononuclear cells (BMMCs) from two healthy individuals. Following~\cite{LEVINE2015184}, we run clustering analysis on individual H1, where $n = 72463$ cells were assigned to populations and $p = 32$. We report clustering results with $K = 14$ and $\gamma = 0.02$. The misclassification error for kmeans++ (our Algorithm~\ref{alg:MR_WSL} with 1-st round using kmeans++ as initialization) is $0.5709$ $(0.4719)$ with time cost $0.8757$ ($226.5155$).

The second one is for unbalanced synthetic $2$-D Gaussian clusters data presented by~\cite{UnbalanceSet}, where $n = 6500, p = 2, K = 8, \gamma = 0.01$. The misclassification error for kmeans++ (our Algorithm~\ref{alg:MR_WSL} with 1-st round using kmeans++ as initialization) is $0.4301$ ($0.2213$) and time cost is $0.0122$ ($0.5257$). Thus, the SL method is robust to the GMM assumption and can improve the accuracy on top of kmeans++ with similar scalability (both time costs increased by $O(10^2)$ times as $n$ becomes $O(10^2)$ times larger).

\subsubsection*{Acknowledgements}
Xiaohui Chen was partially supported by NSF CAREER Award DMS-1752614. Yun Yang was partially supported by NSF DMS-1907316.


\subsubsection*{References}
\begingroup
\renewcommand{\section}[2]{}%
\bibliography{Final-kmeans}
\endgroup

\clearpage
\appendix

\thispagestyle{empty}

\onecolumn \makesupplementtitle

\section{PROOF OF MAIN RESULTS}
In this section, we prove the main result of this paper.

\subsection{Auxiliary lemmas}

Recall that $n_k = |G_k^*|$ denotes the number of data points in $k$-th cluster and $\Delta^2 = \min_{1 \leq k \neq l \leq K} \|\mu_k-\mu_l\|^2$ is the minimal separation between cluster centers. Set $N = \min_{1 \leq k \neq l \leq K} \{{2 n_k n_l \over n_k + n_l}\}$ and $\underline{n} = \min_{k \in [K]} n_k$.

\begin{lem}[Separation bound for exact recovery by the full SDP: general case]\label{lem:SDP_bound_general}
If there exist constants $\tilde{\delta}>0$ and $\beta\in(0,1)$ such that 
\begin{align*}
\log n \geq \frac{(1-\beta)^2}{\beta^2} \frac{C_1n}{N}, \; \tilde{\delta}\leq \frac{\beta^2}{(1-\beta)^2}\frac{C_2}{K}, \; N \geq {4 (1+\tilde{\delta})^2 \over \tilde{\delta}^2},
\end{align*}
and
\begin{align*}
\Delta^2 \geq \frac{4\sigma^2(1+2\tilde{\delta})}{(1-\beta)^{2}} \left(1+\sqrt{1+\frac{(1-\beta)^{2}}{(1+\tilde{\delta})} \frac{p}{N\log n}+ C_3 r_{n} }\,\right) \log n
\end{align*}
with 
\[
r_{n} = {(1-\beta)^{2} \over (1+\tilde{\delta}) \log{n}} \left( {\sqrt{p\log{n}} \over \underline n} + {\log{n} \over \underline n} \right),
\]
then the SDP in~\eqref{eqn:kmeans_sdp} achieves exact recovery with probability at least $1-C_4\,K^2\,n^{-\tilde{\delta}}$, where $C_i$, $i=1,2,3,4$, are universal constants.
\end{lem}

Lemma~\ref{lem:SDP_bound_general} is proved by~\cite{chen2021cutoff} (Theorem II.1). Specializing Lemma~\ref{lem:SDP_bound_general} to equal cluster case $n_1 = \cdots = n_K = n/K$, we have the following corollary.

\begin{cor}[Separation bound for exact recovery by the full SDP: equal cluster case]\label{cor:SDP_bound_general}
Let $\alpha > 0$ and $\overline{\Delta}^{2}_1$ be defined in~\eqref{eqn:threshold}. Suppose that the cluster sizes are equal and $K \leq C_{1} \log(n) / \log\log(n)$ for some small constant $C_{1}>0$ depending only on $\alpha$. If $\Delta^{2} \geq (1+\alpha) \overline{\Delta}^{2}_1$, then the SDP in~\eqref{eqn:kmeans_sdp} achieves exact recovery with probability at least $1-C_{2} (\log{n})^{-c_3}$, where $C_{2}, c_{3}$ are constants depending only on $\alpha$.
\end{cor}

\subsection{Proof outline for Theorem~\ref{thm:SL_separation_upper_bound}}

Before presenting the rigorous proof, we first discuss the overall strategy for proving Theorem~\ref{thm:SL_separation_upper_bound} for equal cluster size case, which can be divided into three steps. Proofs for Theorems~\ref{thm:SL_separation_upper_bound_bias_correction} and~\ref{thm:WSL_separation_upper_bound} have the same architecture. We define the events
\begin{align*}
A := & \Big\{\hat{G}_1=G^*_1,\dots,\hat{G}_K=G^*_K   \Big\}, \\
B := & \Big\{\hat{R}_1=R_1,\dots,\hat{R}_K=R_K   \Big\}, \\
B_{\tau} := & \Big\{ (1-\tau) {n\gamma \over K} \leq |R_k| \leq (1+\tau) {n\gamma \over K}, \ \forall k \in [K]  \Big\},
\end{align*}
where $\tau \in (0,1)$ and $R_k=G^*_k \cap T$. Since probability of wrong recovery satisfies
\begin{equation}\label{eqn:SL_master_bound}
\bP(A^c) \leq 2 \bP(B_\tau^c) + \bP(B^c | B_\tau) + \bP(A^c \cap B | B_\tau),
\end{equation}
it suffices to bound the three terms on the right-hand side of~\eqref{eqn:SL_master_bound}, where the first term is due to the tolerance of approximate equality of the cluster sizes on subsampled data (step 1), the second term is due to wrong recovery using the subsampled SDP (step 2), and the third term is due to the lifting procedure (step 3) on the data points that are not sampled in step 1.

In step 1, we reduce the exact recovery problem on the entire $n$ data points to the subsampled $m \approx n \gamma$ data points with $K$ clusters with approximately equal size that resembles the problem structure of the original SDP clustering problem. The violation probability of approximate cluster size in step 1 can be controlled by the classical Chernoff bound
\[
\bP(B_\tau^c) \leq 2 n^{-1} \quad \text{for } \tau \asymp \sqrt{K \log(n) / n \gamma}.
\]
In step 2, since the subsampling procedure is independent of the original data points, we can treat the subsampled data points $V = (X_i)_{i \in T}$ as a new clustering problem. Based on this observation, we establish the exact recovery guarantee on subsampled data using the separation upper bound proved for the general unbalanced GMM by~\cite{chen2021cutoff}, from which we show that if the minimal separation $\Delta^2 \geq (1+\alpha) \bar{\Delta}_\gamma^2$ and $K \lesssim \log(\gamma n) / \log\log(\gamma n)$, then
\[
\bP(B^c | B_\tau) \lesssim 1/\log^c(\gamma n).
\]
In step 3, we use the nearest centroids $\bar{X}_1,\dots,\bar{X}_K$ estimated from step 2 for propagating the solution from the subsampled SDP to all data points that are not sampled in step 1. The key structure for the lift step to be successful is the independence between $V$ and $X \setminus V$. In particular, the independence among $\bar{X}_k, \bar{X}_l,X_i,i \in [n]/T$ entails that under the minimal separation $\Delta^2 \geq (1+\alpha) \bar{\Delta}_\gamma^2$, the error probability for the nearest-centroid procedure assigning $i \in \hat{G}_k$ via $\|X_i-\bar{X}_l\|^2 > \|X_i-\bar{X}_k\|^2$ for all $i \in G_k^* \setminus T$ vanishes
\[
\bP(A^c \cap B | B_\tau) \leq K^2 / n^c.
\]
Combining the above three steps with the master bound~\eqref{eqn:SL_master_bound}, we conclude that the probability of exact recovery $\bP(A) \geq 1 - C \log^{-c}(\gamma n)$ for large enough $n$, ensuring correctness of the SL approach.

\subsection{Proof of Theorem~\ref{thm:SL_separation_upper_bound}}

\textbf{Step 1: reduction to subsampled data points.} Let $T \subset [n]$ be the subsampled data point indices so that $V = (X_i)_{i \in T} \subset X$ and $m = |T|$. Let $R_k=G^*_k \cap T$.
Define the events
\begin{align*}
A := & \Big\{\hat{G}_1=G^*_1,\dots,\hat{G}_K=G^*_K   \Big\}, \\
B := & \Big\{\hat{R}_1=R_1,\dots,\hat{R}_K=R_K   \Big\}, \\
B_{\tau} := & \Big\{ (1-\tau) {n\gamma \over K} \leq |R_k| \leq (1+\tau) {n\gamma \over K}, \ \forall k \in [K]  \Big\},
\end{align*}
where $ \tau \in (0,1)$. Observe that
\begin{align*}
\bP(A^c) =& \bP(A^c \cap (B \cap B_\tau)) + \bP(A^c \cap (B^c \cup B_\tau^c) ) \\
\leq & \bP(A^c \cap B | B_\tau) \bP(B_\tau) + \bP(A^c \cap B^c) + \bP(A^c \cap B_\tau^c) \\
\leq & \bP(A^c \cap B | B_\tau) + \bP(A^c \cap B^c \cap B_\tau) + 2 \bP(B_\tau^c) \\
\leq & \bP(A^c \cap B | B_\tau) + \bP(A^c \cap B^c | B_\tau) + 2 \bP(B_\tau^c) \\
\leq & \bP(A^c \cap B | B_\tau) + \bP(B^c | B_\tau) + 2 \bP(B_\tau^c).
\end{align*}
Thus to bound the error probability for exact recovery, it suffices to bound the three terms on the right-hand side of the last inequality. Since the subsampled data points $V$ from $X = (X_1,\dots,X_n)$ are drawn with i.i.d. $\text{Ber}(\gamma)$, we apply the Chernoff bound and the union bound to get
\[
\bP(B_\tau^c) \leq 2 K \exp\Big(-{\tau^2 n \gamma \over 3 K}\Big).
\]
Choosing $\tau = \sqrt{6K\log(n) / n\gamma}$, we have
\[
\bP(B_\tau^c) \leq 2 n^{-1}.
\]

\textbf{Step 2: exact recovery for subsampled data.}
Since the subsampling procedure is independent of the original $X_i$ points, we can treat the $X_i \in V$ as the new cluster problem to apply Lemma~\ref{lem:SDP_bound_general} with $T=\bigcup_{k=1}^K R_k$, $n=m$ and $n_k=m_k$, where $m_k = |R_k|$. In particular, if there exist constants $\tilde{\delta}>0$ and $\beta \in (0,1)$ such that 
\begin{align*}
& \log m \geq \frac{(1-\beta)^2}{\beta^2}\frac{C_1 m}{M}, \\
& \tilde{\delta} \leq \frac{\beta^2}{(1-\beta)^2}\frac{C_2}{K}, \ \ N\geq \frac{4(1+\tilde{\delta})^2}{\tilde{\delta}^2},
\end{align*}
and 
\begin{align*}
\Delta^2 \geq & \frac{4\sigma^2(1+2\tilde{\delta})}{(1-\beta)^{2}} \left(1+\sqrt{1+\frac{(1-\beta)^2}{(1+\tilde{\delta})}\frac{p}{M\log m}+C_3 r_m}\,\right) \log m
\end{align*}
with 
$$r_m=\frac{(1-\beta)^2}{(1+\tilde{\delta})\log m} \Big(  \frac{\sqrt{p\log m }}{\underline m}+ \frac{\log m }{\underline m} \Big),$$ where $\underline m=\min_{k \in [K]} m_k$ and $M=\min_{1 \leq k\neq l \leq K} \frac{2 m_lm_k}{m_l+m_k}$, 
then the SDP achieves exact recovery, i.e., $\hat{R}_k=R_k, \ \forall k \in [K]$, with probability at least $1-C_4K^2m^{-\tilde{\delta}},$ where $C_i, \ i=1,2,3,4$ are universal constants. Note that on event $B_\tau$, we have
\begin{align*}
& (1-\tau) {n \gamma} \leq m \leq (1+\tau) {n \gamma}, \\
& {2 m_l m_k \over m_l + m_k} = {2 \over m_l^{-1} + m_k^{-1}} \geq (1-\tau) {n \gamma \over K}.
\end{align*}
Thus on the event $B_\tau$, we can choose an upper bound $\Delta'^2:$
\begin{align*}
\Delta'^2:=  \frac{4\sigma^2(1+2\tilde{\delta})}{(1-\beta)^2} \left(1+\sqrt{1+\frac{(1-\beta)^2}{(1+\tilde{\delta})}\frac{pK/((1-\tau)\gamma n)}{\log  ((1+\tau)\gamma n)}+C_3 r'_m}\,\right) \log  ((1+\tau)\gamma n)
\end{align*}
with 
\begin{align*}
r'_m=  \frac{(1-\beta)^2}{(1+\tilde{\delta})\log  ((1+\tau)\gamma n)}\left(  \frac{K\sqrt{p\log  ((1+\tau)\gamma n) }}{  (1-\tau)\gamma n}+ \frac{K\log  ((1+\tau)\gamma n) }{ (1-\tau)\gamma n}\,\right).
\end{align*}
Note that $\tau = \sqrt{6K\log(n) / n\gamma} = o(1)$ under the assumption ${K \log{n} \over n} = o(\gamma)$. Fix an $\alpha > 0$. By choosing small enough $\beta$ and $\tilde{\delta}$ that may also also depend on $\alpha$, we have for large enough $n$, if $K \leq C_1 {\log(\gamma n) \over \log\log(\gamma n)}$ for some constant $C_1$ depending on $\alpha$ and $\Delta^2 \geq (1+\alpha)\bar{\Delta}_\gamma^2$, where
$$\bar{\Delta}_\gamma^2=4\sigma^2 \left(1+\sqrt{1+\frac{Kp}{\gamma n\log n}} \right)\log n,$$ 
then SDP achieves exact recovery with probability at least $1-C_2(\log (\gamma n))^{-C_3}$, where  $C_2, C_3$ depend only on $\alpha$. Thus we conclude that
\[
\bP(B^c | B_\tau) \leq C_2(\log (\gamma n))^{-C_3}.
\]
\begin{rem}[Lower bound for $\gamma$]\label{rem:SDP_bound}
It can be seen that the lower bound condition ${K \log{n} \over n} = o(\gamma)$ for $\gamma$ can be relaxed to $K=o(n\gamma/ \log\log(\gamma n))$ given $K \leq C_1 {\log(\gamma n) \over \log\log(\gamma n)}$. i.e., we can think of $K \leq C_1 {\log(\gamma n) \over \log\log(\gamma n)}$ as another way to interpret the lower bound for $\gamma.$

\end{rem}
\textbf{Step 3: lift solution from sketched SDP to all the data points.} Recall that the lift solution to all $n$ data points is defined as
$$\hat{G}_k=\Big\{i \in [n] \setminus T : \| X_i-\bar{X}_k \|<\| X_i-\bar{X}_l \|, \forall l\neq k  \Big\}\cup \hat{R}_k, $$ 
where $\bar{X}_k=\sum_{j \in \hat{R}_k} X_j/m_k$ is the centroid of the $k$-th cluster output from the subsampled SDP. Since our goal in this step is to bound $\bP(A^c \cap B | B_\tau)$, the subsequent analysis will be on the event $B$, that is $\hat{R}_k = R_k$ for all $k \in [K]$. Then we have $\bar{X}_k=\sum_{j \in R_k} X_j/m_k$ and
$$\hat{G}_k=\Big\{i \in [n] \setminus T : \| X_i-\bar{X}_k \|<\| X_i-\bar{X}_l \|, \forall l\neq k  \Big\}\cup R_k.$$
Let $\mathcal{A}_{kl}^{(i)}=\Big\{\| X_i-\bar{X}_l \|^2- \| X_i-\bar{X}_k\|^2>\xi \Big\}, $ where $i \in G^*_k\backslash T,$ where $\xi$ is some number to be determined. Recall that $X_i=\mu_k+\epsilon_i, \forall i\in G^*_k$, where $\epsilon_i$ are i.i.d. $N(0, \sigma^2 I_p).$ Denote similarly $\bar{X}_k=\mu_k+\bar{\epsilon}_k$, where $\bar{\epsilon}_k = \sum_{j \in R_k} \epsilon_i / m_k$. For $i \in G^*_k\backslash T$, we note that $\epsilon_i,\bar{\epsilon}_k, \bar{\epsilon}_l$ are independent. 
We can write 
\begin{align*}
&  \| X_i-\bar{X}_l \|^2- \| X_i-\bar{X}_k\|^2\\
= & \|\theta+\epsilon_i-\bar{\epsilon}_l  \|^2-\|\epsilon_i - \bar{\epsilon}_k  \|^2 \\
= & \|\theta  \|^2+ \|  \bar{\epsilon}_l \|^2-  \|  \bar{\epsilon}_k  \|^2 -2 \langle \theta,\bar{\epsilon}_l  \rangle + 2 \langle \theta-\bar{\epsilon}_l +\bar{\epsilon}_k,   \epsilon_i \rangle,
\end{align*}
where $\theta = \mu_k - \mu_l$. Set $\zeta_n=2\log(Kn)$ and define 
\begin{align*}
\mathcal{B}_{kl,1}^{(i)}:=\Big\{& \| \bar{\epsilon}_l  \|^2 \geq m_l^{-1}(p-2\sqrt{p\zeta_n}), \\
                                &  \| \bar{\epsilon}_k  \|^2 \leq m_k^{-1} (p+2\sqrt{p\zeta_n}    +2\zeta_n),              \\
                                &\langle \theta,\bar{\epsilon}_l  \rangle \leq \sqrt{2 m_l^{-1}\zeta_n }\|\theta\| \Big\}
\end{align*}
and
\begin{align*}
\mathcal{B}_{kl,2}^{(i)}:=\Big\{ &\| \bar{\epsilon}_l- \bar{\epsilon}_k   \|^2 \leq  (m_l^{-1}+m_k^{-1})(p+2\sqrt{p\zeta_n}    +2\zeta_n),\\
&\langle \theta, \bar{\epsilon}_k -\bar{\epsilon}_l \rangle \leq \sqrt{2 (m_l^{-1}+m_k^{-1})\zeta_n }\|\theta\| \Big\}.
\end{align*}
Let $\mathcal{B}_{kl}^{(i)} = \mathcal{B}_{kl,1}^{(i)} \cup \mathcal{B}_{kl,2}^{(i)}$. Using the standard tail probability bound for $\chi^2$ distribution~\citep{LaurentMassart2000}, we have $\mathbb{P} (\mathcal{B}_{kl}^{(i)^c}  ) \leq 5 / (n^2 K^2)$. Since
$$\langle \theta-\bar{\epsilon}_l +\bar{\epsilon}_k,   \epsilon_i \rangle | {\{\bar{\epsilon}_l, \bar{\epsilon}_k\}}\sim N(0,\|\theta-\bar{\epsilon}_l +\bar{\epsilon}_k\|^2),$$
we have on the event $\mathcal{B}_{kl}^{(i)}$ that
\begin{align*}
\mathcal{C}^*:&= \mathbb{P}\Big(2 \langle \theta-\bar{\epsilon}_l +\bar{\epsilon}_k,   \epsilon_i \rangle \leq-(1-\beta)\|\theta\|^2\Big| \bar{\epsilon}_k , \bar{\epsilon}_l  \Big)         \\
                    &= 1-\Phi\Bigg( \frac{(1-\beta)\|\theta\|^2}{2\sqrt{\|\theta-\bar{\epsilon}_l +\bar{\epsilon}_k\|^2}}  \Bigg)         \\
                    &\leq 1-\Phi\Bigg( \frac{(1-\beta)\|\theta\|^2}{2\sqrt{r''_n}}  \Bigg),         
\end{align*} 
where $\beta\in (0,1),$
\begin{align*}
r''_n=\|\theta\|^2 + 2 \sqrt{2 (m_l^{-1}+m_k^{-1})\zeta_n }\|\theta\|+ (m_l^{-1}+m_k^{-1})(p+2\sqrt{p\zeta_n}    +2\zeta_n).
\end{align*} 
 Note that $\|\theta  \|^2 \geq \Delta^2\geq 8\log n, $ which implies $\sqrt{2 (m_l^{-1}+m_k^{-1})\zeta_n }\leq \|\theta\| \sqrt{2/M}.$
Now we choose $\eta>0$ such that $\frac{1+2\eta}{1+\eta}\geq 1+2\sqrt{2/M}$ (i.e., $M\geq 8(1+\eta^{-1})^2)$. In order to have $\mathcal{C}^*$ be bounded by $n^{-(1+\eta)},$ it is sufficient to require that 
\begin{align*}
 &1-\Phi\Bigg( \frac{(1-\beta)\|\theta\|^2}{2\sqrt{r''_n}}  \Bigg)\leq  1-\Phi( \sqrt{2(1+\eta)\log n} ).    
\end{align*}
or further
\begin{align*}
\frac{(1-\beta)^2}{8(1+\eta)\log n}\|\theta\|^4-(1+2\sqrt{2/M}) \|\theta  \|^2 -(p+2\sqrt{p\zeta_n}    +2\zeta_n)(m_l^{-1}  +m_k^{-1}  ) \geq 0.
\end{align*}
A sufficient condition for the last display is
\begin{align*}
\Delta^2 \geq &\frac{4\sigma^2(1+2\eta)}{(1-\beta)^2} \left(1+\sqrt{1+\frac{(1-\beta)^2}{(1+2\eta)}\frac{p}{M\log n}+2r'''_n}\,\right) \log n,
\end{align*}
where 
$$r'''_n=\frac{(1-\beta)^2}{(1+2\eta)\log n} \Big(  \frac{\sqrt{p\log (nK) }}{\underline m}+ \frac{\log (nK) }{\underline m} \Big).$$
Now if we put
\begin{align*}
\xi= \frac{m_k-m_l}{m_k m_l}p+\beta \|\theta\|^2-4\sqrt{ \log (nK) \over m_l}\|\theta\| -2 \frac{m_k+m_l}{m_k m_l}\sqrt{2p\log (nK)}-4 {\log (nK) \over m_k},  
\end{align*}
then we have
\begin{align*}
&\mathbb{P}\Big(\Big\{\| X_i-\bar{X}_l \|^2- \| X_i-\bar{X}_k\|^2>\xi, \\
& \qquad \qquad \forall i\in G^*_k\backslash T, \  \forall 1 \leq k\neq l \leq K \Big\}^c\Big)\\
&= \mathbb{P}\Big(\bigcup_{i=1}^n \bigcup_{1 \leq k \neq l \leq K} {A}_{kl}^{(i)^c}  \Big)         \\
                    &\leq \sum_{i=1}^n \sum_{1 \leq k \neq l \leq K} \mathbb{P}\Big(  {A}_{kl}^{(i)^c} \cap \mathcal{B}_{kl}^{(i)}  \Big)+ \mathbb{P} \Big(\mathcal{B}_{kl}^{(i)^c}  \Big)         \\
                    &\leq \sum_{i=1}^n \sum_{1 \leq k \neq l \leq K} \mathbb{E}[\mathcal{C}^*1_{\mathcal{B}_{kl,2}^{(i)}}] + {5 \over n} \\
                    &\leq {K^2 \over n^{\eta}} + {7 \over n}.      
\end{align*}
Next we claim that $\xi>0.$ Recall that on the event $B_\tau$, we have $m_k \in [(1- \tau)m_*,(1+ \tau)m_*], \ 1/M\in [\frac{1}{(1+ \tau)m_*},\frac{1}{(1- \tau)m_*}],$ where $m_*= n\gamma\slash K.$ Then,
$$\Big|\frac{m_k-m_l}{m_k m_l}p\Big|\leq \frac{2 \tau}{(1- \tau)^2}\frac{p}{m_*} \leq {6 p \sqrt{\log{n}} \over (1-\tau^2) m_*^{3/2}}.$$ Note that
\begin{align*}
\|\theta  \|^2 \geq \bar{\Delta}_\gamma^2 \geq 4\sigma^2 \left(1+\sqrt{1+\frac{p}{m_*\log n}}\right)\log n. 
\end{align*}
So if $p=O(\gamma n/K)^{2})$, then 
\[
\Big|\frac{m_k-m_l}{m_k m_l}p\Big|\leq \frac{\beta}{5} \|\theta  \|^2
\]
for large enough $n$. Similarly, we have 
\begin{align*}
&4\sqrt{ \log (nK)/m_l}\|\theta\|\leq\frac{\beta}{5} \|\theta  \|^2,\\
&2 \frac{m_k+m_l}{m_k m_l}\sqrt{2p\log (nK)}\leq \frac{\beta}{5} \|\theta  \|^2,\\
&4m_k^{-1} \log (nK) \leq \frac{\beta}{5} \|\theta  \|^2.
\end{align*}
For $\alpha>0,$ we can choose small enough $\beta := \beta(\alpha,\sigma) > 0$ and $\eta := \eta(\alpha).$ Then for $n$ large, we have if $\Delta^2 \geq (1+\alpha) \bar{\Delta}_\gamma^2$, then
\begin{align*}
& \bP(A^c \cap B | B_\tau) \\
= & \mathbb{P}\Big(\Big\{\| X_i-\bar{X}_l \|^2- \| X_i-\bar{X}_k\|^2>0, \forall i\in G^*_k\backslash T, \  \forall 1 \leq k\neq l \leq K \Big\}^c\Big) \\
\leq & { K^2 \over n^{\eta}}.
\end{align*}
Now, combining all pieces together, we conclude that, for all $n$ large enough,
\[
\bP(\hat{G}_1=G_1^*,\dots,\hat{G}_K=G_K^*) \geq 1 - C (\log(\gamma n))^{-c}.
\]

\subsection{Proof outline for Theorem~\ref{thm:SL_separation_upper_bound_bias_correction}}

The overall strategy for proving Theorem~\ref{thm:SL_separation_upper_bound_bias_correction} for unequal cluster size case is identical to the proof of Theorem~\ref{thm:SL_separation_upper_bound}, which can be divided into three steps. We will briefly talk about the difference here. And all the details are contained in the proof.

In step 1, we reduce the exact recovery problem on the entire $n$ data points to the subsampled $m \approx n \gamma$ data points with $K$ clusters with approximately $\gamma n_k$ many points for each cluster $\hat{G}_k$ that resembles the problem structure of the original SDP clustering problem. The parameter $\tau$ in the classical Chernoff bound now should be
\[
 \tau \asymp \sqrt{ \log(n) / \underline n \gamma}.
\]
In step 2, since the subsampling procedure is independent of the original data points, we can treat the subsampled data points $V = (X_i)_{i \in T}$ as a new clustering problem. Same as proof of Theorem~\ref{thm:SL_separation_upper_bound}, we establish the exact recovery guarantee on subsampled data using the separation upper bound proved for the general unbalanced GMM by~\cite{chen2021cutoff}, from which we get the similar conditions. i.e., $\Delta^2 \geq (1+\alpha) \bar{\Delta}_\gamma^{'2}$ and $K \lesssim \log(\gamma n) / \log\log(\gamma n)$.

In step 3, we first get the subsampled clusters from step 2 and for each cluster, we randomly down-sample same size (the minimum one) of points to make the new clusters have the same sample size. Then we use the nearest centroids $\bar{X}_1,\dots,\bar{X}_K$ of the new clusters for propagating the solution from the subsampled SDP to all data points that are not sampled in step 1. The independence among $\bar{X}_k, \bar{X}_l,X_i,i \in [n]/T$ entails that under the minimal separation $\Delta^2 \geq (1+\alpha) \bar{\Delta}_\gamma^2$, the error probability for the nearest-centroid procedure assigning $i \in \hat{G}_k$ via $\|X_i-\bar{X}_l\|^2 > \|X_i-\bar{X}_k\|^2$ for all $i \in G_k^* \setminus T$ vanishes.

\subsection{Proof of Theorem~\ref{thm:SL_separation_upper_bound_bias_correction}}

\textbf{Step 1: reduction to subsampled data points.}
Let $T \subset [n]$ be the subsampled data point indices so that $V = (X_i)_{i \in T} \subset X$ and $m = |T|$. Let $n_k=|G^*_k| $, $R_k=G^*_k \cap T$.
Define the events
\begin{align*}
A := & \Big\{\hat{G}_1=G^*_1,\dots,\hat{G}_K=G^*_K   \Big\}, \\
B := & \Big\{\hat{R}_1=R_1,\dots,\hat{R}_K=R_K   \Big\}, \\
B_{\tau} := & \Big\{ (1-\tau) {n_k\gamma } \leq |R_k| \leq (1+\tau) {n_k\gamma }, \ \forall k \in [K]  \Big\},
\end{align*}
where $ \tau \in (0,1)$. Observe that
\begin{align*}
\bP(A^c) =& \bP(A^c \cap (B \cap B_\tau)) + \bP(A^c \cap (B^c \cup B_\tau^c) ) \\
\leq & \bP(A^c \cap B | B_\tau) \bP(B_\tau) + \bP(A^c \cap B^c) + \bP(A^c \cap B_\tau^c) \\
\leq & \bP(A^c \cap B | B_\tau) + \bP(A^c \cap B^c \cap B_\tau) + 2 \bP(B_\tau^c) \\
\leq & \bP(A^c \cap B | B_\tau) + \bP(A^c \cap B^c | B_\tau) + 2 \bP(B_\tau^c) \\
\leq & \bP(A^c \cap B | B_\tau) + \bP(B^c | B_\tau) + 2 \bP(B_\tau^c).
\end{align*}

Thus to bound the error probability for exact recovery, it suffices to bound the three terms on the right-hand side of the last inequality. Since the subsampled data points $V$ from $X = (X_1,\dots,X_n)$ are drawn with i.i.d. $\text{Ber}(\gamma)$, we apply the Chernoff bound and the union bound to get
\[
\bP(B_\tau^c) \leq 2 K \exp\Big(-{\tau^2 \underline n \gamma \over 3 }\Big).
\]
Choosing $\tau = \sqrt{6\log(n) / \underline n\gamma}$, we have
\[
\bP(B_\tau^c) \leq 2 n^{-1}.
\]

\textbf{Step 2: exact recovery for subsampled data.}
Since the subsampling procedure is independent of the original $X_i$ points, we can treat the $X_i \in V$ as the new cluster problem to apply Lemma~\ref{lem:SDP_bound_general} with $T=\bigcup_{k=1}^K R_k$, $n=m$ and $n_k=m_k$, where $m_k = |R_k|$. In particular, if there exist constants $\tilde{\delta}>0$ and $\beta \in (0,1)$ such that 
$$\log m \geq \frac{(1-\beta)^2}{\beta^2}\frac{C_1 m}{M}, \ \ \ \tilde{\delta}\leq \frac{\beta^2}{(1-\beta)^2}\frac{C_2}{K}, \ \ \ N\geq \frac{4(1+\tilde{\delta})^2}{\tilde{\delta}^2}$$ and 
\begin{align*}
\Delta^2 \geq \frac{4\sigma^2(1+2\tilde{\delta})}{(1-\beta)^{2}} \left(1+\sqrt{1+\frac{(1-\beta)^2}{(1+\tilde{\delta})}\frac{p}{M\log m}+C_3 r_m}\,\right) \log m
\end{align*}
with 
$$r_m=\frac{(1-\beta)^2}{(1+\tilde{\delta})\log m} \Big(  \frac{\sqrt{p\log m }}{\underline m}+ \frac{\log m }{\underline m} \Big),$$ where $\underline m=\min_{k \in [K]} m_k$ and $M=\min_{1 \leq k\neq l \leq K} \frac{2 m_lm_k}{m_l+m_k}$, 
then the SDP achieves exact recovery, i.e., $\hat{R}_k=R_k, \ \forall k \in [K]$, with probability at least $1-C_4K^2m^{-\tilde{\delta}},$ where $C_i, \ i=1,2,3,4$ are universal constants. Note that on event $B_\tau$, we have

\begin{align*}
& (1-\tau) {n \gamma} \leq m \leq (1+\tau) {n \gamma}, \\
& {2 m_l m_k \over m_l + m_k} = {2 \over m_l^{-1} + m_k^{-1}} \geq (1-\tau) {\underline n \gamma }.
\end{align*}
Thus on the event $B_\tau$, we can choose an upper bound $\Delta'^2:$
\begin{align*}
\Delta'^2:=  \frac{4\sigma^2(1+2\tilde{\delta})}{(1-\beta)^2} \left(1+\sqrt{1+\frac{(1-\beta)^2}{(1+\tilde{\delta})}\frac{p/((1-\tau)\gamma \underline n)}{\log  ((1+\tau)\gamma n)}+C_3 r'_m}\,\right) \log  ((1+\tau)\gamma n)
\end{align*}
with 
\begin{align*}
r'_m=  \frac{(1-\beta)^2}{(1+\tilde{\delta})\log  ((1+\tau)\gamma n)} \left(  \frac{\sqrt{p\log  ((1+\tau)\gamma  n) }}{  (1-\tau)\gamma \underline n}+ \frac{\log  ((1+\tau)\gamma n) }{ (1-\tau)\gamma \underline n}\,\right).
\end{align*}
Note that $\tau = \sqrt{6\log(n) / \underline n\gamma} = o(1)$ under the assumption ${ \log{n} \over \underline n} = o(\gamma)$. Fix an $\alpha > 0$. By choosing small enough $\beta$ and $\tilde{\delta}$ that may also also depend on $\alpha$, we have for large enough $n$, if $K \leq C_1 {\log(\gamma n) \over \log\log(\gamma n)}, \underline n\ge C_2 n/\log(\gamma n)$ for some constant $C_1, C_2$ depending on $\alpha$ and $\Delta^2 \geq (1+\alpha)\bar{\Delta}_\gamma^2$, where
$$\bar{\Delta}_\gamma^2=4\sigma^2 \left(1+\sqrt{1+\frac{p}{\gamma \underline n\log n}} \right)\log n,$$ 
then SDP achieves exact recovery with probability at least $1-C_3(\log (\gamma n))^{-C_4}$, where  $C_3, C_4$ depend only on $\alpha$. Thus we conclude that
\[
\bP(B^c | B_\tau) \leq C_2(\log (\gamma n))^{-C_3}.
\]

\textbf{Step 3: lift solution from sketched SDP to all the data points.}

Recall that the lift solution to all $n$ data points is defined as
$$\hat{G}_k=\Big\{i \in [n] \setminus T : \| X_i-\bar{X}_k \|<\| X_i-\bar{X}_l \|, \forall l\neq k  \Big\}\cup \hat{R}_k, $$ 
 
where $\bar{X}_k = |\tilde{R}_k|^{-1} \sum_{j \in \tilde{R}_k}$ is the revised centroid of the $k$-th cluster output from the subsampled SDP. And $\tilde{R}_k$ is a randomly sampled subset of $\hat{R}_k$ with equal size $\underline{m}$. Since our goal in this step is to bound $\bP(A^c \cap B | B_\tau)$, the subsequent analysis will be on the event $B$, that is $\hat{R}_k = R_k$ for all $k \in [K]$.

Then we have $\bar{X}_k=\sum_{j\in \tilde{R}_k\subseteq R_k} X_j/\underline m$ and
$$\hat{G}_k=\Big\{i \in [n] \setminus T : \| X_i-\bar{X}_k \|<\| X_i-\bar{X}_l \|, \forall l\neq k  \Big\}\cup R_k.$$
Let $\mathcal{A}_{kl}^{(i)}=\Big\{\| X_i-\bar{X}_l \|^2- \| X_i-\bar{X}_k\|^2>\xi \Big\}, $ where $i \in G^*_k\backslash T,$ where $\xi$ is some number to be determined. If we further make the analysis on $T'\subseteq T,$ let $\mathcal{A}_{kl,t'}^{(i)}=\mathcal{A}_{kl}^{(i)}\cap \{T'=t'\},$ where $t'=\bigsqcup_{k=1}^K R'_k$ is any realization of $T'$. Recall that $X_i=\mu_k+\epsilon_i, \forall i\in G^*_k$, where $\epsilon_i$ are i.i.d. $N(0, \sigma^2 I_p).$ Denote similarly $\bar{X}_k=\mu_k+\bar{\epsilon}_k$, where $\bar{\epsilon}_k = \sum_{j \in R_k} \epsilon_i / \underline m$. For $i \in G^*_k\backslash T$, we note that $\epsilon_i,\bar{\epsilon}_k, \bar{\epsilon}_l$ are independent. 
We can write 
\begin{align*}
&  \| X_i-\bar{X}_l \|^2- \| X_i-\bar{X}_k\|^2\\
= & \|\theta+\epsilon_i-\bar{\epsilon}_l  \|^2-\|\epsilon_i - \bar{\epsilon}_k  \|^2 \\
= & \|\theta  \|^2+ \|  \bar{\epsilon}_l \|^2-  \|  \bar{\epsilon}_k  \|^2 -2 \langle \theta,\bar{\epsilon}_l  \rangle + 2 \langle \theta-\bar{\epsilon}_l +\bar{\epsilon}_k,   \epsilon_i \rangle,
\end{align*}
where $\theta = \mu_k - \mu_l$. Set $\zeta_n=2\log(Kn)$ and define 
\begin{align*}
\mathcal{B}_{kl,1}^{(i)}:=\Big\{& \| \bar{\epsilon}_l  \|^2 \geq \underline m^{-1}(p-2\sqrt{p\zeta_n}), \\
                                &  \| \bar{\epsilon}_k  \|^2 \leq \underline m^{-1} (p+2\sqrt{p\zeta_n}    +2\zeta_n),              \\
                                &\langle \theta,\bar{\epsilon}_l  \rangle \leq \sqrt{2 \underline m^{-1}\zeta_n }\|\theta\| \Big\}
\end{align*}
and
\begin{align*}
\mathcal{B}_{kl,2}^{(i)}:=\Big\{ &\| \bar{\epsilon}_l- \bar{\epsilon}_k   \|^2 \leq  2\underline m^{-1})(p+2\sqrt{p\zeta_n}    +2\zeta_n),\\
&\langle \theta, \bar{\epsilon}_k -\bar{\epsilon}_l \rangle \leq 2\sqrt{ (\underline m^{-1})\zeta_n }\|\theta\| \Big\}.
\end{align*}
Let $\mathcal{B}_{kl}^{(i)} = \mathcal{B}_{kl,1}^{(i)} \cup \mathcal{B}_{kl,2}^{(i)}$. Using the standard tail probability bound for $\chi^2$ distribution~\citep{LaurentMassart2000}, we have $\mathbb{P} (\mathcal{B}_{kl}^{(i)^c}  ) \leq 5 / (n^2 K^2)$. Since
$$\langle \theta-\bar{\epsilon}_l +\bar{\epsilon}_k,   \epsilon_i \rangle | {\{\bar{\epsilon}_l, \bar{\epsilon}_k\}}\sim N(0,\|\theta-\bar{\epsilon}_l +\bar{\epsilon}_k\|^2),$$
we have on the event $\mathcal{B}_{kl}^{(i)}$ that
\begin{align*}
\mathcal{C}^*:&= \mathbb{P}\Big(2 \langle \theta-\bar{\epsilon}_l +\bar{\epsilon}_k,   \epsilon_i \rangle \leq-(1-\beta)\|\theta\|^2\Big| \bar{\epsilon}_k , \bar{\epsilon}_l  \Big)         \\
                    &= 1-\Phi\Bigg( \frac{(1-\beta)\|\theta\|^2}{2\sqrt{\|\theta-\bar{\epsilon}_l +\bar{\epsilon}_k\|^2}}  \Bigg)         \\
                    &\leq 1-\Phi\Bigg( \frac{(1-\beta)\|\theta\|^2}{2\sqrt{r''_n}}  \Bigg),         
\end{align*} 
where $\beta\in (0,1),$
\begin{align*}
r''_n=\|\theta\|^2 + 4 \sqrt{ \underline m^{-1}\zeta_n }\|\theta\|+ 2\underline m^{-1}(p+2\sqrt{p\zeta_n}    +2\zeta_n).
\end{align*} 
 Note that $\|\theta  \|^2 \geq \Delta^2\geq 8\log n, $ which implies $2 \sqrt{ \underline m^{-1}\zeta_n }\leq \|\theta\| \sqrt{2/\underline m}.$
Now we choose $\eta>0$ such that $\frac{1+2\eta}{1+\eta}\geq 1+2\sqrt{2/\underline m}$ (i.e., $\underline m\geq 8(1+\eta^{-1})^2)$. In order to have $\mathcal{C}^*$ be bounded by $n^{-(1+\eta)},$ it is sufficient to require that 
\begin{align*}
 &1-\Phi\Bigg( \frac{(1-\beta)\|\theta\|^2}{2\sqrt{r''_n}}  \Bigg)\leq  1-\Phi( \sqrt{2(1+\eta)\log n} ).    
\end{align*}
or further
\begin{align*}
\frac{(1-\beta)^2}{8(1+\eta)\log n}\|\theta\|^4-(1+2\sqrt{2/M}) \|\theta  \|^2-2(p+2\sqrt{p\zeta_n}    +2\zeta_n)\underline m^{-1} \geq 0.
\end{align*}
A sufficient condition for the last display is
\begin{align*}
\Delta^2 \geq \frac{4\sigma^2(1+2\eta)}{(1-\beta)^2} \left(1+\sqrt{1+\frac{(1-\beta)^2}{(1+2\eta)}\frac{p}{\underline m\log n}+2r'''_n}\,\right) \log n,
\end{align*}
where 
$$r'''_n=\frac{(1-\beta)^2}{(1+2\eta)\log n} \Big(  \frac{\sqrt{p\log (nK) }}{\underline m}+ \frac{\log (nK) }{\underline m} \Big).$$
Now if we put
\begin{align*}
\xi= \beta \|\theta\|^2-4\sqrt{ \log (nK) \over \underline m}\|\theta\|-4 \underline m^{-1}\sqrt{2p\log (nK)}-4 {\log (nK) \over \underline m},  
\end{align*}
notice that 
\begin{align*}
 \mathbb{P}\Big( {A}_{kl}^{(i)^c}  \Big)&= \sum_{t'\in \mathcal{T}'} \mathbb{P}\Big( \mathcal{A}_{kl,t'}^{(i)}\Big)\cdot \mathbb{P}\Big( T'=t'  \Big) \le \max_{t'\in \mathcal{T}'} \mathbb{P}\Big( \mathcal{A}_{kl,t'}^{(i)}\Big),
\end{align*}
where $\mathcal{T}'$ is all the possible subset of $T$ s.t. $|\tilde{R_i}|=\underline m$. Then we have
\begin{align*}
&\mathbb{P}\Big(\Big\{\| X_i-\bar{X}_l \|^2- \| X_i-\bar{X}_k\|^2>\xi, \\
& \qquad \qquad \forall i\in G^*_k\backslash T, \  \forall 1 \leq k\neq l \leq K \Big\}^c\Big)\\
&= \mathbb{P}\Big(\bigcup_{i=1}^n \bigcup_{1 \leq k \neq l \leq K} {A}_{kl}^{(i)^c}  \Big)         \\
                     &\leq \sum_{i=1}^n \sum_{1 \leq k \neq l \leq K} \mathbb{P}\Big(  {A}_{kl}^{(i)^c}   \Big) \\
                    &\leq \sum_{i=1}^n \sum_{1 \leq k \neq l \leq K} \max_{t'\in T'} \mathbb{P}\Big( \mathcal{A}_{kl,t'}^{(i)} \cap \mathcal{B}_{kl}^{(i)}  \Big)+ \mathbb{P} \Big(\mathcal{B}_{kl}^{(i)^c}  \Big)         \\
                    &\leq \sum_{i=1}^n \sum_{1 \leq k \neq l \leq K} \mathbb{E}[\mathcal{C}^*1_{\mathcal{B}_{kl,2}^{(i)}}] + {5 \over n} \\
                    &\leq {K^2 \over n^{\eta}} + {7 \over n}.      
\end{align*}
Next we claim that $\xi>0.$ Recall that on the event $B_\tau$, we have $m_k \in [(1- \tau)\gamma n_k,(1+ \tau)\gamma n_k], \ 1/\underline m \in [\frac{1}{(1+ \tau)\gamma \underline n},\frac{1}{(1- \tau)\gamma\underline n}].$ Note that
\begin{align*}
\|\theta  \|^2 \geq \bar{\Delta}_\gamma^2 \geq 4\sigma^2 \left(1+\sqrt{1+\frac{p}{\gamma\underline n\log n}}\right)\log n. 
\end{align*}
So if $\gamma \underline n\to \infty$ as $n\to \infty, n$ large, then we have 
\begin{align*}
&4\sqrt{ \log (nK)/\underline m}\|\theta\|\leq\frac{\beta}{5} \|\theta  \|^2,\\
&4 \underline m^{-1} \sqrt{2p\log (nK)}\leq \frac{\beta}{5} \|\theta  \|^2,\\
&4\underline m ^{-1} \log (nK) \leq \frac{\beta}{5} \|\theta  \|^2.
\end{align*}
For $\alpha>0,$ we can choose small enough $\beta := \beta(\alpha,\sigma) > 0$ and $\eta := \eta(\alpha).$ Then for $n$ large, we have if $\Delta^2 \geq (1+\alpha) \bar{\Delta}_\gamma^2$, then
\begin{align*}
& \bP(A \cap B | B_\tau) \\
= & \mathbb{P}\Big(\Big\{\| X_i-\bar{X}_l \|^2- \| X_i-\bar{X}_k\|^2>0, \forall i\in G^*_k\backslash T, \  \forall 1 \leq k\neq l \leq K \Big\}^c\Big) \\
\leq & { K^2 \over n^{\eta}}.
\end{align*}
Now, combining all pieces together, we conclude that, for all $n$ large enough,
\[
\bP(\hat{G}_1=G_1^*,\dots,\hat{G}_K=G_K^*) \geq 1 - C (\log(\gamma n))^{-c}.
\]

\subsection{Proof outline for Theorem~\ref{thm:WSL_separation_upper_bound}}

The overall strategy for proving Theorem~\ref{thm:WSL_separation_upper_bound} for unequal cluster size case is identical to the proof of Theorem~\ref{thm:SL_separation_upper_bound}, which can be divided into three steps. Again we will briefly talk about the difference here. 

In step 1, we reduce the exact recovery problem on the entire $n$ data points to the subsampled $m \approx n \gamma$ data points with $K$ clusters with approximately equal size that resembles the problem structure of the original SDP clustering problem. Here we use the weighted sampling ratio $w_i$ for each point $i\in[n],$ where majority of them should be around the true weights $w_i^*, \ i\in[n].$ We will apply the classical Chernoff bound by setting appropriate $\tau$ through the definition of $(\epsilon, \delta)$ pairs. i.e.,
\[
\tau \asymp \sqrt{ \log(n) / \gamma \underline n }+\epsilon+\delta n/\underline n.
\]
In step 2, we assume $\tau=o(1)$ by setting each summand of $\tau$ to be $o(1)$. And we set the same conditions for minimal separation  $\Delta^2 \geq (1+\alpha) \bar{\Delta}_\gamma^2$ and $K \lesssim \log(\gamma n) / \log\log(\gamma n)$.

In step 3, we use the nearest centroids $\bar{X}_1,\dots,\bar{X}_K$ estimated from step 2 for propagating the solution from the subsampled SDP to all data points that are not sampled in step 1. The discussion here is similar to the proof of Theorem~\ref{thm:SL_separation_upper_bound}. The only difference here is that now we bound the size difference by $\tau$ through $(\epsilon,\delta)$ pairs. i.e.,
\[
|m_k-m_l|<2\tau\asymp \sqrt{ \log(n) / \gamma \underline n }+\epsilon+\delta n/\underline n,
\] where $\ k\ne l\in [n].$  

\subsection{Proof of Theorem~\ref{thm:WSL_separation_upper_bound}}

\textbf{Step 1: reduction to subsampled data points.} Let $T \subset [n]$ be the subsampled data point indices so that $V = (X_i)_{i \in T} \subset X$ and $m = |T|$. Let $R_k=G^*_k \cap T$.
Define the events
\begin{align*}
A := & \Big\{\hat{G}_1=G^*_1,\dots,\hat{G}_K=G^*_K   \Big\}, \\
B := & \Big\{\hat{R}_1=R_1,\dots,\hat{R}_K=R_K   \Big\}, \\
B_{\tau} := & \Big\{ (1-\tau) {n\gamma \over K} \leq |R_k| \leq (1+\tau) {n\gamma \over K}, \ \forall k \in [K]  \Big\}, \\
B_{\tau_1} := & \Big\{ (1-\tau_1) {m_k^*} \leq |R_k| \leq (1+\tau_1) {m_k^*}, \ \forall k \in [K]  \Big\},
\end{align*}
where $m_k^*=\sum_{i\in G^*_k }w_i, \tau,\tau_1 \in (0,1)$. Observe that
\begin{align*}
\bP(A^c) =& \bP(A^c \cap (B \cap B_\tau)) + \bP(A^c \cap (B^c \cup B_\tau^c) ) \\
\leq & \bP(A^c \cap B | B_\tau) \bP(B_\tau) + \bP(A^c \cap B^c) + \bP(A^c \cap B_\tau^c) \\
\leq & \bP(A^c \cap B | B_\tau) + \bP(A^c \cap B^c \cap B_\tau) + 2 \bP(B_\tau^c) \\
\leq & \bP(A^c \cap B | B_\tau) + \bP(A^c \cap B^c | B_\tau) + 2 \bP(B_\tau^c) \\
\leq & \bP(A^c \cap B | B_\tau) + \bP(B^c | B_\tau) + 2 \bP(B_\tau^c).
\end{align*}
Thus to bound the error probability for exact recovery, it suffices to bound the three terms on the right-hand side of the last inequality. Since the subsampled data points $V$ from $X = (X_1,\dots,X_n)$ are drawn with i.i.d. $\text{Ber}(\gamma)$, we apply the Chernoff bound and the union bound to get
\[
\bP(B_{\tau_1}^c) \leq 2 \sum_{k\in[K]} \exp\Big(-{\tau^2 m_k^* \over 3 }\Big).
\]
Choosing $\tau_1 = \sqrt{6\log(n) /\underline m^*}$, we have
\[
\bP(B_{\tau_1}^c) \leq 2 n^{-1}.
\]
If we choose $\tau=\tau_1+\epsilon +\delta (1+\epsilon) \bar{w}^*K/\gamma ,$
then 
\[
\bP(B_{\tau}^c) \leq\bP(B_{\tau_1}^c) \leq 2 n^{-1}.
\]

\textbf{Step 2: exact recovery for subsampled data.}
Since the subsampling procedure is independent of the original $X_i$ points, we can treat the $X_i \in V$ as the new cluster problem to apply Lemma~\ref{lem:SDP_bound_general} with $T=\bigcup_{k=1}^K R_k$, $n=m$ and $n_k=m_k$, where $m_k = |R_k|$. In particular, if there exist constants $\tilde{\delta}>0$ and $\beta \in (0,1)$ such that 
$$\log m \geq \frac{(1-\beta)^2}{\beta^2}\frac{C_1 m}{M}, \ \ \ \tilde{\delta}\leq \frac{\beta^2}{(1-\beta)^2}\frac{C_2}{K}, \ \ \ N\geq \frac{4(1+\tilde{\delta})^2}{\tilde{\delta}^2}$$ and 
\begin{align*}
\Delta^2 \geq &\frac{4\sigma^2(1+2\tilde{\delta})}{(1-\beta)^{2}} \left(1+\sqrt{1+\frac{(1-\beta)^2}{(1+\tilde{\delta})}\frac{p}{M\log m}+C_3 r_m}\,\right) \log m
\end{align*}
with 
$$r_m=\frac{(1-\beta)^2}{(1+\tilde{\delta})\log m} \Big(  \frac{\sqrt{p\log m }}{\underline m}+ \frac{\log m }{\underline m} \Big),$$ where $\underline m=\min_{k \in [K]} m_k$ and $M=\min_{1 \leq k\neq l \leq K} \frac{2 m_lm_k}{m_l+m_k}$, 
then the SDP achieves exact recovery, i.e., $\hat{R}_k=R_k, \ \forall k \in [K]$, with probability at least $1-C_4K^2m^{-\tilde{\delta}},$ where $C_i, \ i=1,2,3,4$ are universal constants. Note that on event $B_\tau$, we have
\begin{align*}
& (1-\tau) {n \gamma} \leq m \leq (1+\tau) {n \gamma}, \\
& {2 m_l m_k \over m_l + m_k} = {2 \over m_l^{-1} + m_k^{-1}} \geq (1-\tau) {n \gamma \over K}.
\end{align*}
Thus on the event $B_\tau$, we can choose an upper bound $\Delta'^2:$
\begin{align*}
\Delta'^2:=  \frac{4\sigma^2(1+2\tilde{\delta})}{(1-\beta)^2}\left(1+\sqrt{1+\frac{(1-\beta)^2}{(1+\tilde{\delta})}\frac{pK/((1-\tau)\gamma n)}{\log  ((1+\tau)\gamma n)}+C_3 r'_m}\,\right) \log  ((1+\tau)\gamma n)
\end{align*}
with 
\begin{align*}
r'_m=  \frac{(1-\beta)^2}{(1+\tilde{\delta})\log  ((1+\tau)\gamma n)} \left(  \frac{K\sqrt{p\log  ((1+\tau)\gamma n) }}{  (1-\tau)\gamma n}+ \frac{K\log  ((1+\tau)\gamma n) }{ (1-\tau)\gamma n}\,\right).
\end{align*}
Note that $\tau=\tau_1+\epsilon +\delta (1+\epsilon) \bar{w}^*K/\gamma = o(1)$ under the assumption $\tau_1 = o(1), \epsilon =o(1), \delta (1+\epsilon) \bar{w}^*K/\gamma = o(1)$. i.e. $\frac{K\log n}{n}=o(\gamma), \delta=o(\underline n/n), \epsilon =o(1)$.Fix an $\alpha > 0$. By choosing small enough $\beta$ and $\tilde{\delta}$ that may also also depend on $\alpha$, we have for large enough $n$, if $K \leq C_1 {\log(\gamma n) \over \log\log(\gamma n)}$ for some constant $C_1$ depending on $\alpha$ and $\Delta^2 \geq (1+\alpha)\bar{\Delta}_\gamma^2$, where
$$\bar{\Delta}_\gamma^2=4\sigma^2 \left(1+\sqrt{1+\frac{Kp}{\gamma n\log n}} \right)\log n,$$ 
then SDP achieves exact recovery with probability at least $1-C_2(\log (\gamma n))^{-C_3}$, where  $C_2, C_3$ depend only on $\alpha$. Thus we conclude that
\[
\bP(B^c | B_\tau) \leq C_2(\log (\gamma n))^{-C_3}.
\]
\textbf{Step 3: lift solution from sketched SDP to all the data points.} Recall that the lift solution to all $n$ data points is defined as
$$\hat{G}_k=\Big\{i \in [n] \setminus T : \| X_i-\bar{X}_k \|<\| X_i-\bar{X}_l \|, \forall l\neq k  \Big\}\cup \hat{R}_k, $$ 
where $\bar{X}_k=\sum_{j \in \hat{R}_k} X_j/m_k$ is the centroid of the $k$-th cluster output from the subsampled SDP. Since our goal in this step is to bound $\bP(A^c \cap B | B_\tau)$, the subsequent analysis will be on the event $B$, that is $\hat{R}_k = R_k$ for all $k \in [K]$. Then we have $\bar{X}_k=\sum_{j \in R_k} X_j/m_k$ and
$$\hat{G}_k=\Big\{i \in [n] \setminus T : \| X_i-\bar{X}_k \|<\| X_i-\bar{X}_l \|, \forall l\neq k  \Big\}\cup R_k.$$
Let $\mathcal{A}_{kl}^{(i)}=\Big\{\| X_i-\bar{X}_l \|^2- \| X_i-\bar{X}_k\|^2>\xi \Big\}, $ where $i \in G^*_k\backslash T,$ where $\xi$ is some number to be determined. Recall that $X_i=\mu_k+\epsilon_i, \forall i\in G^*_k$, where $\epsilon_i$ are i.i.d. $N(0, \sigma^2 I_p).$ Denote similarly $\bar{X}_k=\mu_k+\bar{\epsilon}_k$, where $\bar{\epsilon}_k = \sum_{j \in R_k} \epsilon_i / m_k$. For $i \in G^*_k\backslash T$, we note that $\epsilon_i,\bar{\epsilon}_k, \bar{\epsilon}_l$ are independent. 
We can write 
\begin{align*}
&  \| X_i-\bar{X}_l \|^2- \| X_i-\bar{X}_k\|^2\\
= & \|\theta+\epsilon_i-\bar{\epsilon}_l  \|^2-\|\epsilon_i - \bar{\epsilon}_k  \|^2 \\
= & \|\theta  \|^2+ \|  \bar{\epsilon}_l \|^2-  \|  \bar{\epsilon}_k  \|^2 -2 \langle \theta,\bar{\epsilon}_l  \rangle + 2 \langle \theta-\bar{\epsilon}_l +\bar{\epsilon}_k,   \epsilon_i \rangle,
\end{align*}
where $\theta = \mu_k - \mu_l$. Set $\zeta_n=2\log(Kn)$ and define 
\begin{align*}
\mathcal{B}_{kl,1}^{(i)}:=\Big\{& \| \bar{\epsilon}_l  \|^2 \geq m_l^{-1}(p-2\sqrt{p\zeta_n}), \\
                                &  \| \bar{\epsilon}_k  \|^2 \leq m_k^{-1} (p+2\sqrt{p\zeta_n}    +2\zeta_n),              \\
                                &\langle \theta,\bar{\epsilon}_l  \rangle \leq \sqrt{2 m_l^{-1}\zeta_n }\|\theta\| \Big\}
\end{align*}
and
\begin{align*}
\mathcal{B}_{kl,2}^{(i)}:=\Big\{ &\| \bar{\epsilon}_l- \bar{\epsilon}_k   \|^2 \leq  (m_l^{-1}+m_k^{-1})(p+2\sqrt{p\zeta_n}    +2\zeta_n),\\
&\langle \theta, \bar{\epsilon}_k -\bar{\epsilon}_l \rangle \leq \sqrt{2 (m_l^{-1}+m_k^{-1})\zeta_n }\|\theta\| \Big\}.
\end{align*}
Let $\mathcal{B}_{kl}^{(i)} = \mathcal{B}_{kl,1}^{(i)} \cup \mathcal{B}_{kl,2}^{(i)}$. Using the standard tail probability bound for $\chi^2$ distribution~\citep{LaurentMassart2000}, we have $\mathbb{P} (\mathcal{B}_{kl}^{(i)^c}  ) \leq 5 / (n^2 K^2)$. Since
$$\langle \theta-\bar{\epsilon}_l +\bar{\epsilon}_k,   \epsilon_i \rangle | {\{\bar{\epsilon}_l, \bar{\epsilon}_k\}}\sim N(0,\|\theta-\bar{\epsilon}_l +\bar{\epsilon}_k\|^2),$$
we have on the event $\mathcal{B}_{kl}^{(i)}$ that
\begin{align*}
\mathcal{C}^*:&= \mathbb{P}\Big(2 \langle \theta-\bar{\epsilon}_l +\bar{\epsilon}_k,   \epsilon_i \rangle \leq-(1-\beta)\|\theta\|^2\Big| \bar{\epsilon}_k , \bar{\epsilon}_l  \Big)         \\
                    &= 1-\Phi\Bigg( \frac{(1-\beta)\|\theta\|^2}{2\sqrt{\|\theta-\bar{\epsilon}_l +\bar{\epsilon}_k\|^2}}  \Bigg)         \\
                    &\leq 1-\Phi\Bigg( \frac{(1-\beta)\|\theta\|^2}{2\sqrt{r''_n}}  \Bigg),         
\end{align*} 
where $\beta\in (0,1),$
\begin{align*}
r''_n=\|\theta\|^2 + 2 \sqrt{2 (m_l^{-1}+m_k^{-1})\zeta_n }\|\theta\|+ (m_l^{-1}+m_k^{-1})(p+2\sqrt{p\zeta_n}    +2\zeta_n).
\end{align*} 
 Note that $\|\theta  \|^2 \geq \Delta^2\geq 8\log n, $ which implies $\sqrt{2 (m_l^{-1}+m_k^{-1})\zeta_n }\leq \|\theta\| \sqrt{2/M}.$
Now we choose $\eta>0$ such that $\frac{1+2\eta}{1+\eta}\geq 1+2\sqrt{2/M}$ (i.e., $M\geq 8(1+\eta^{-1})^2)$. In order to have $\mathcal{C}^*$ be bounded by $n^{-(1+\eta)},$ it is sufficient to require that 
\begin{align*}
 &1-\Phi\Bigg( \frac{(1-\beta)\|\theta\|^2}{2\sqrt{r''_n}}  \Bigg)\leq  1-\Phi( \sqrt{2(1+\eta)\log n} ).    
\end{align*}
or further
\begin{align*}
\frac{(1-\beta)^2}{8(1+\eta)\log n}\|\theta\|^4-(1+2\sqrt{2/M}) \|\theta  \|^2-(p+2\sqrt{p\zeta_n}    +2\zeta_n)(m_l^{-1}  +m_k^{-1}  ) \geq 0.
\end{align*}
A sufficient condition for the last display is
\begin{align*}
\Delta^2 \geq &\frac{4\sigma^2(1+2\eta)}{(1-\beta)^2} \left(1+\sqrt{1+\frac{(1-\beta)^2}{(1+2\eta)}\frac{p}{M\log n}+2r'''_n}\,\right) \log n,
\end{align*}
where 
$$r'''_n=\frac{(1-\beta)^2}{(1+2\eta)\log n} \Big(  \frac{\sqrt{p\log (nK) }}{\underline m}+ \frac{\log (nK) }{\underline m} \Big).$$
Now if we put
\begin{align*}
\xi= \frac{m_k-m_l}{m_k m_l}p+\beta \|\theta\|^2-4\sqrt{ \log (nK) \over m_l}\|\theta\|-2 \frac{m_k+m_l}{m_k m_l}\sqrt{2p\log (nK)}-4 {\log (nK) \over m_k},  
\end{align*}
then we have
\begin{align*}
&\mathbb{P}\Big(\Big\{\| X_i-\bar{X}_l \|^2- \| X_i-\bar{X}_k\|^2>\xi, \\
& \qquad \qquad \forall i\in G^*_k\backslash T, \  \forall 1 \leq k\neq l \leq K \Big\}^c\Big)\\
&= \mathbb{P}\Big(\bigcup_{i=1}^n \bigcup_{1 \leq k \neq l \leq K} {A}_{kl}^{(i)^c}  \Big)         \\
                    &\leq \sum_{i=1}^n \sum_{1 \leq k \neq l \leq K} \mathbb{P}\Big(  {A}_{kl}^{(i)^c} \cap \mathcal{B}_{kl}^{(i)}  \Big)+ \mathbb{P} \Big(\mathcal{B}_{kl}^{(i)^c}  \Big)         \\
                    &\leq \sum_{i=1}^n \sum_{1 \leq k \neq l \leq K} \mathbb{E}[\mathcal{C}^*1_{\mathcal{B}_{kl,2}^{(i)}}] + {5 \over n} \\
                    &\leq {K^2 \over n^{\eta}} + {7 \over n}.      
\end{align*}
Next we claim that $\xi>0.$ Recall that on the event $B_\tau$, we have $m_k \in [(1- \tau)m_*,(1+ \tau)m_*], \ 1/M\in [\frac{1}{(1+ \tau)m_*},\frac{1}{(1- \tau)m_*}],$ where $m_*= n\gamma\slash K.$ Then,
\begin{align*}
    g\Big|\frac{m_k-m_l}{m_k m_l}p\Big| \leq \frac{2 \tau}{(1- \tau)^2}\frac{p}{m_*} \leq {6 p \sqrt{\log{n}} \over (1-\tau)^2 m_*\sqrt{\underline m^*}}+\frac{2 \epsilon}{(1- \tau)^2}\frac{p}{m_*}+\frac{2 (\lambda (1+\epsilon) \bar{w}^*K/\gamma)}{(1- \tau)^2}\frac{p}{m_*}.
\end{align*}
Note that $\underline m^*\ge m_*(1-\epsilon)-\delta n(1+\epsilon)\bar{w}^*, $
\begin{align*}
\|\theta  \|^2 \geq \bar{\Delta}_\gamma^2 \geq 4\sigma^2 \left(1+\sqrt{1+\frac{p}{m_*\log n}}\right)\log n. 
\end{align*}
So if $p=O(\gamma n/K)^{2}), \epsilon =O( \frac{\gamma n}{Kp} \log n),\delta=O( \sqrt{\gamma n/p} \cdot \underline n/n)$ then 
\[
\Big|\frac{m_k-m_l}{m_k m_l}p\Big|\leq \frac{\beta}{5} \|\theta  \|^2
\]
for large enough $n$. Similarly, we have 
\begin{align*}
&4\sqrt{ \log (nK)/m_l}\|\theta\|\leq\frac{\beta}{5} \|\theta  \|^2,\\
&2 \frac{m_k+m_l}{m_k m_l}\sqrt{2p\log (nK)}\leq \frac{\beta}{5} \|\theta  \|^2,\\
&4m_k^{-1} \log (nK) \leq \frac{\beta}{5} \|\theta  \|^2.
\end{align*}
For $\alpha>0,$ we can choose small enough $\beta := \beta(\alpha,\sigma) > 0$ and $\eta := \eta(\alpha).$ Then for $n$ large, we have if $\Delta^2 \geq (1+\alpha) \bar{\Delta}_\gamma^2$, then
\begin{align*}
& \bP(A \cap B | B_\tau) \\
= & \mathbb{P}\Big(\Big\{\| X_i-\bar{X}_l \|^2- \| X_i-\bar{X}_k\|^2>0, \forall i\in G^*_k\backslash T, \  \forall 1 \leq k\neq l \leq K \Big\}^c\Big) \\
\leq & { K^2 \over n^{\eta}}.
\end{align*}
Now, combining all pieces together, we conclude that, for all $n$ large enough,
\[
\bP(\hat{G}_1=G_1^*,\dots,\hat{G}_K=G_K^*) \geq 1 - C (\log(\gamma n))^{-c}.
\]

\section{FULL NUMERICAL EXPERIMENT RESULTS}
\label{sec:more_simulation_results}
\begin{figure}[h!] 
   \centering
      \subfigure{\includegraphics[trim={1.5cm 7cm 1.5cm 7cm},clip,scale=0.45]{epo.pdf}} 
      \subfigure{\includegraphics[trim={1.5cm 7cm 1.5cm 7cm},clip,scale=0.45]{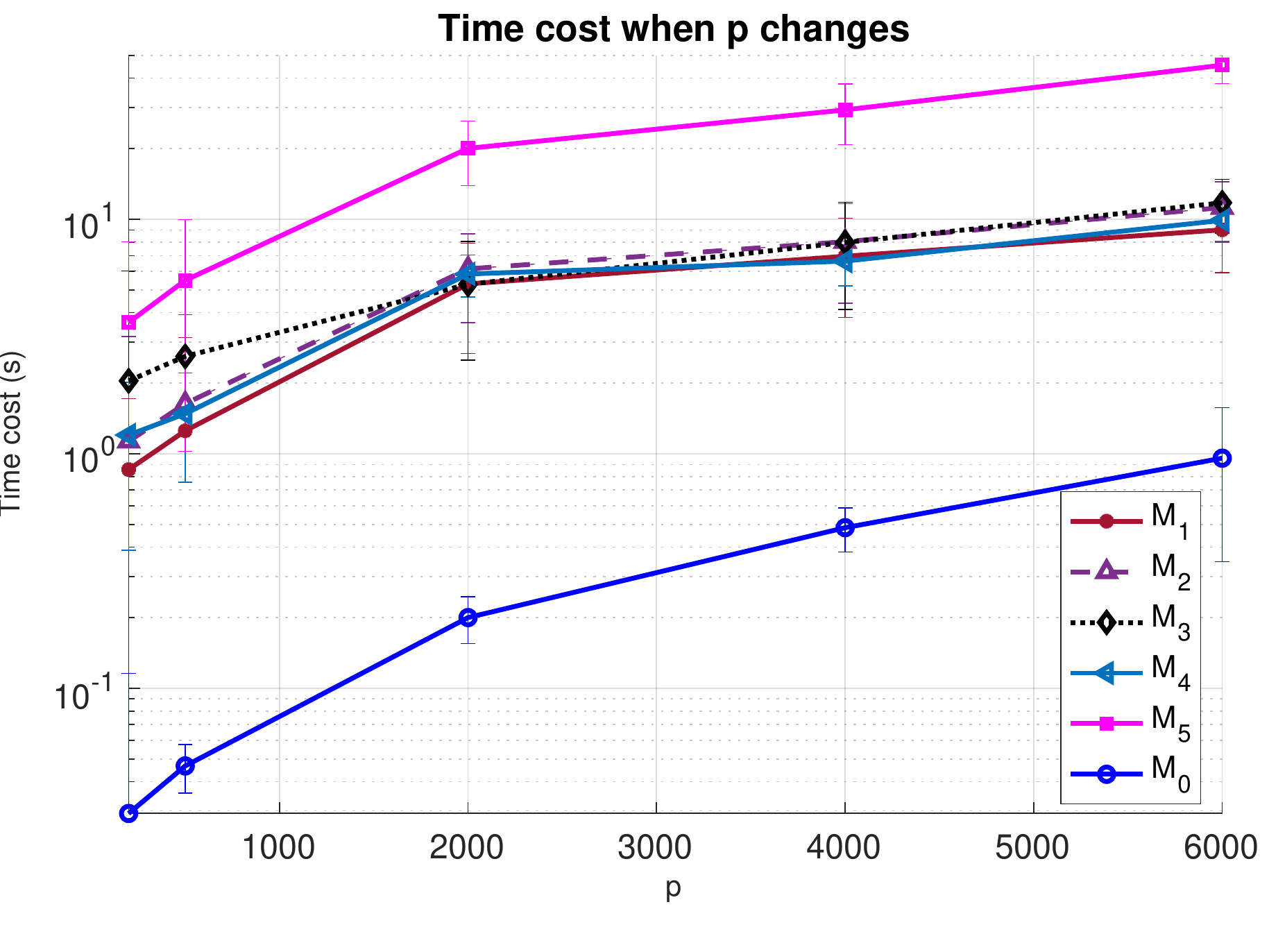}} \\[-3ex]
      
   \caption{Log-scale error rates and runtime (with error bars) v.s. $p$ under the setting of equal cluster size and $\Delta^2= (\lambda^{*}\bar\Delta_\ast)^{2}$. Zero error is displayed as $10^{-6}$ in the log-scale plot.}
   \label{fig:run_time1}
\end{figure}

\begin{figure}[h!] 
   \centering
      \subfigure{\includegraphics[trim={1.5cm 7cm 1.5cm 7cm},clip,scale=0.45]{ego.pdf}} 
      \subfigure{\includegraphics[trim={1.5cm 7cm 1.5cm 7cm},clip,scale=0.45]{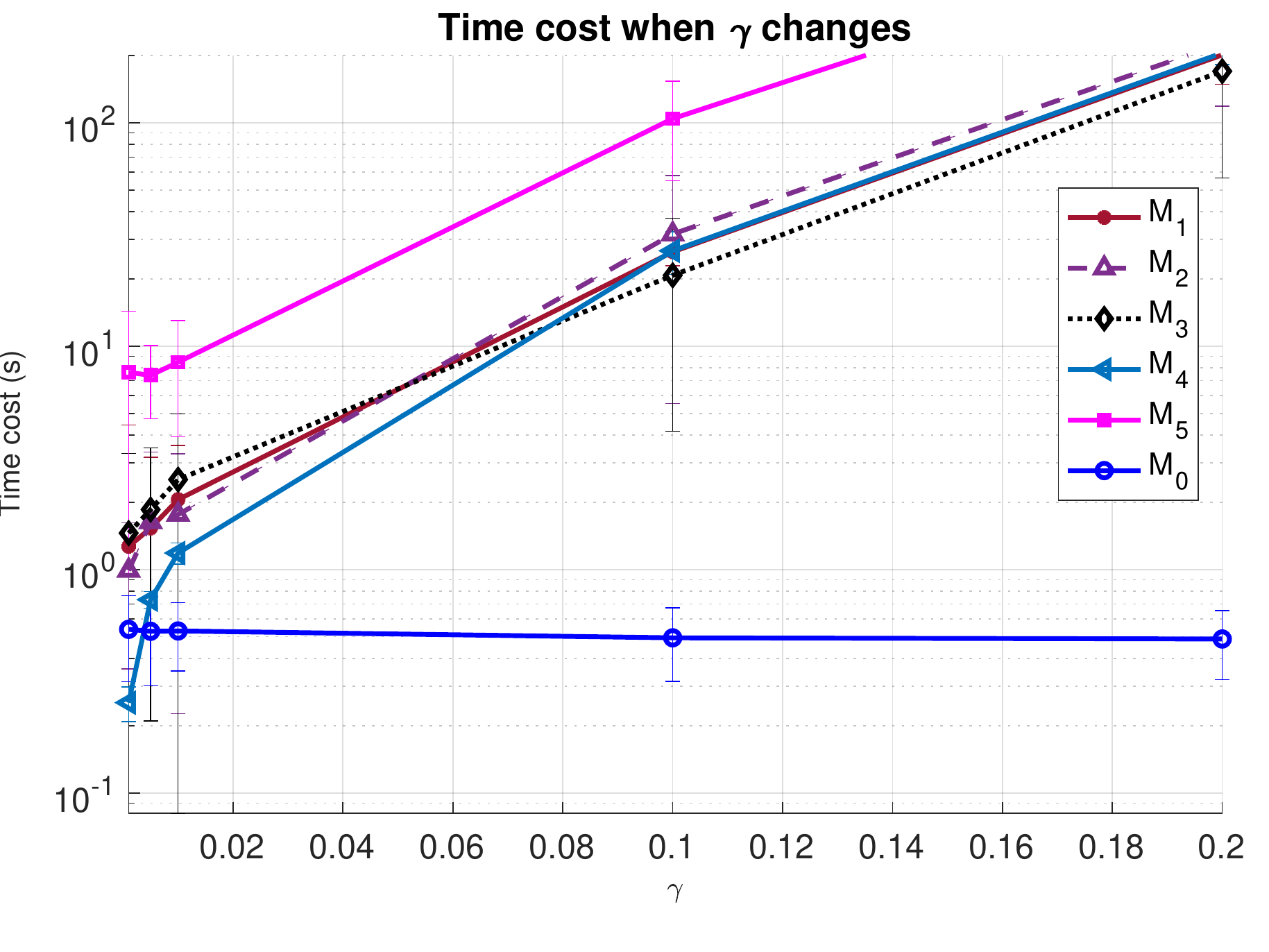}} \\[-3ex]
      
   \caption{Log-scale error rates and runtime (with error bars) v.s. $\gamma$ under the setting of equal cluster size and $\Delta^2= (\lambda^{*}\bar\Delta_\ast)^{2}$. Zero error is displayed as $10^{-6}$ in the log-scale plot.}
   \label{fig:run_time3}
\end{figure}

\begin{figure}[h!] 
   \centering
      \subfigure{\includegraphics[trim={1.5cm 7cm 1.5cm 7cm},clip,scale=0.45]{elo.pdf}} 
      \subfigure{\includegraphics[trim={1.5cm 7cm 1.5cm 7cm},clip,scale=0.45]{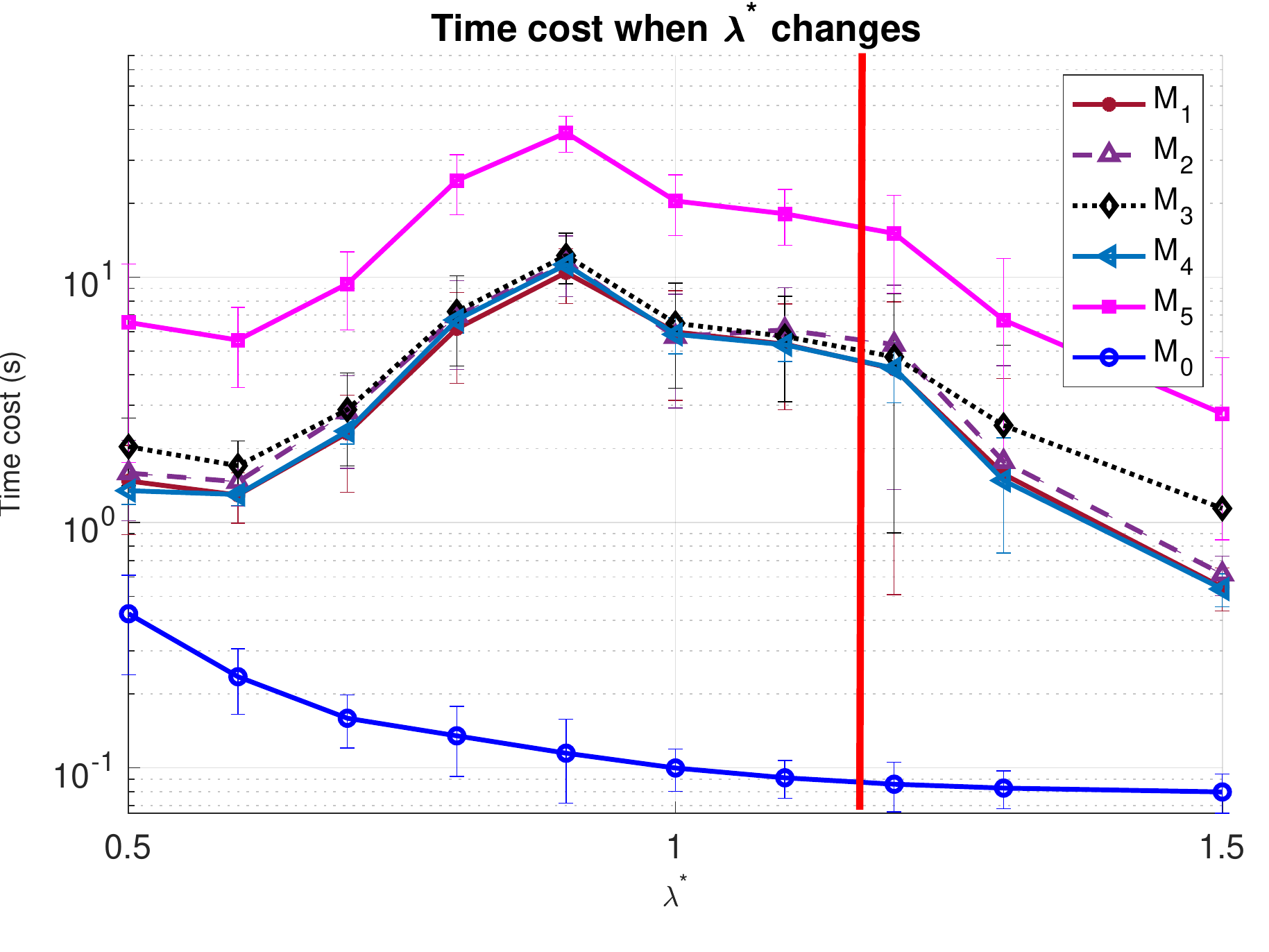}} \\[-3ex]
      
   \caption{Log-scale error rates and runtime (with error bars) v.s. $\lambda^*$ under the setting of equal cluster size and $\Delta^2= (\lambda^{*}\bar\Delta_\ast)^{2}$. Red vertical line indicates theoretical threshold $\bar\Delta_\gamma^2$ for SL methods. Zero error is displayed as $10^{-6}$ in the log-scale plot.}
   \label{fig:run_time4}
\end{figure}

\begin{figure}[h!] 
   \centering
      \subfigure{\includegraphics[trim={1.5cm 7cm 1.5cm 7cm},clip,scale=0.45]{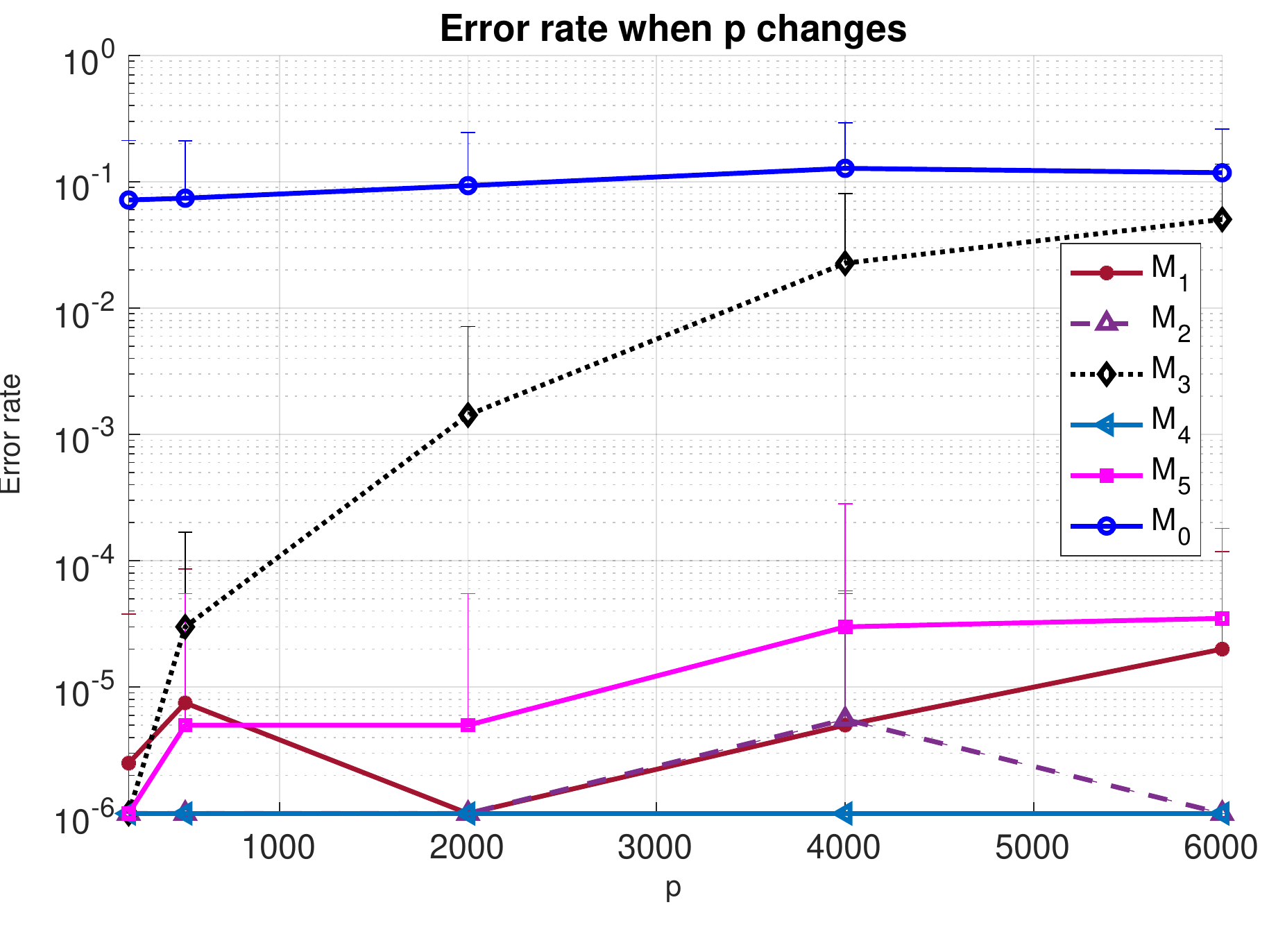}} 
      \subfigure{\includegraphics[trim={1.5cm 7cm 1.5cm 7cm},clip,scale=0.45]{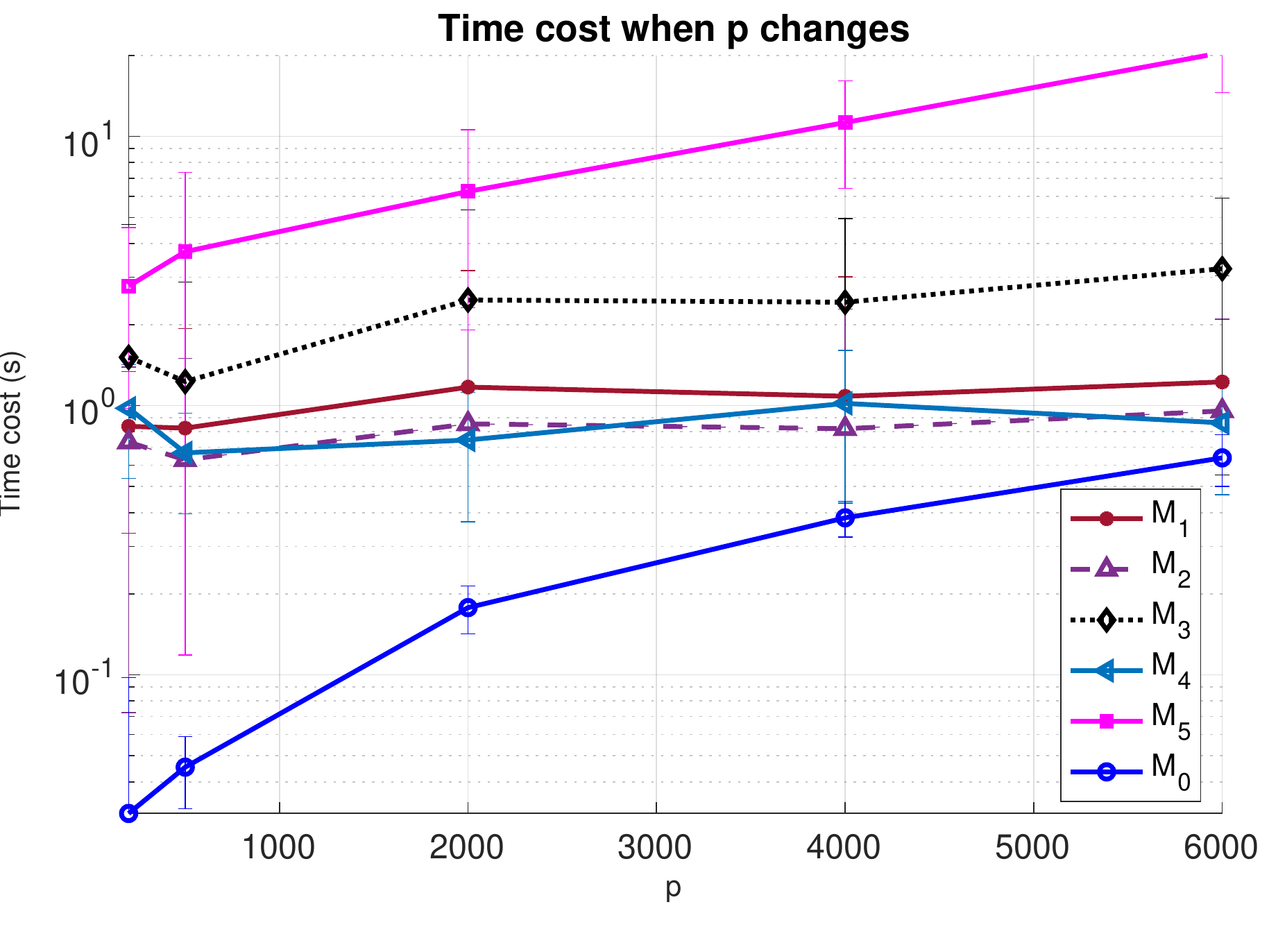}} \\[-3ex]
      
   \caption{Log-scale error rates and runtime (with error bars) v.s. $p$ under the setting of equal cluster size and $\Delta^2= (\lambda^{*}\bar\Delta_\gamma)^{2}$. Zero error is displayed as $10^{-6}$ in the log-scale plot.}
   \label{fig:run_time5}
\end{figure}

\begin{figure}[h!] 
   \centering
      \subfigure{\includegraphics[trim={1.5cm 7cm 1.5cm 7cm},clip,scale=0.45]{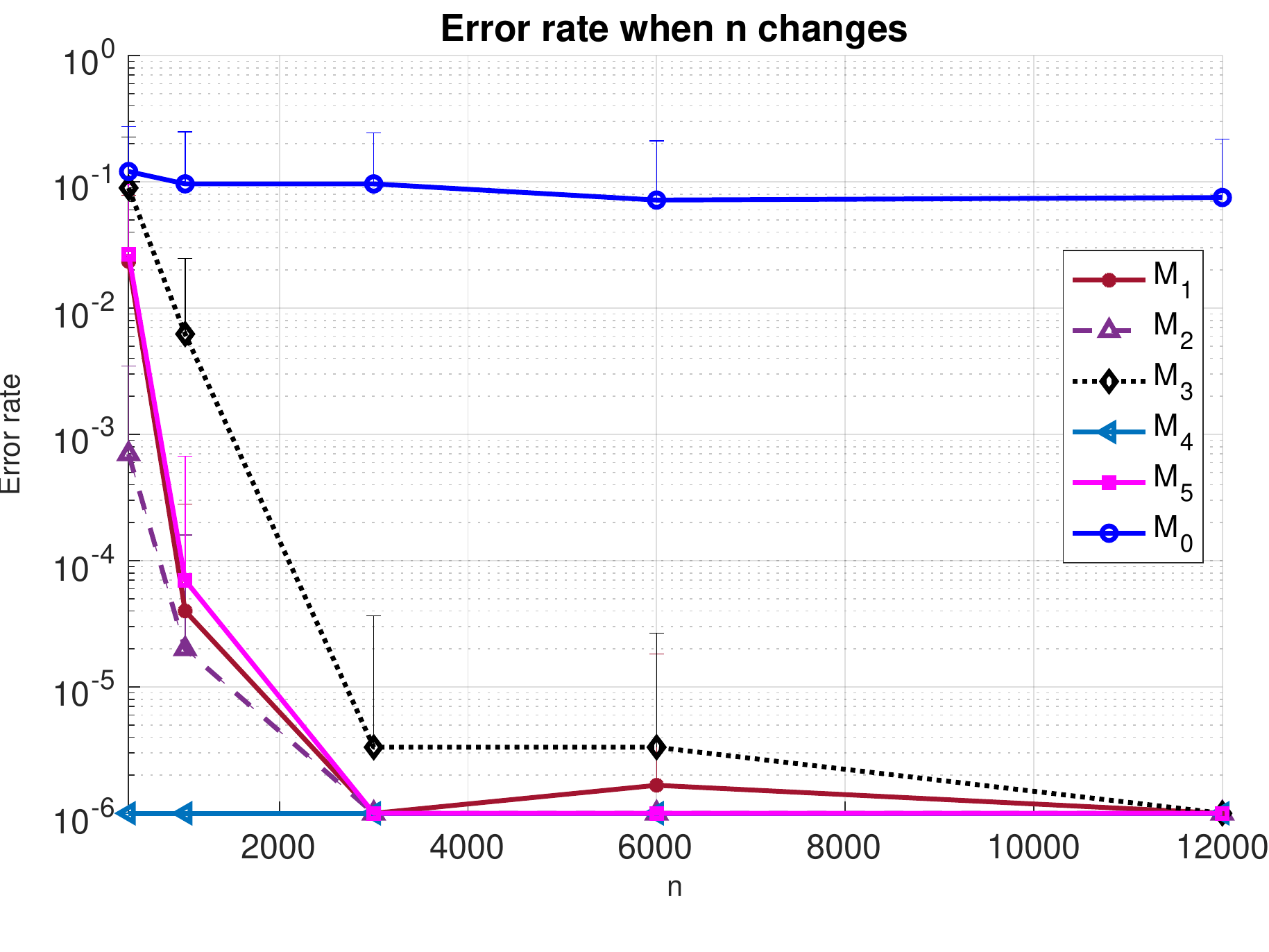}} 
      \subfigure{\includegraphics[trim={1.5cm 7cm 1.5cm 7cm},clip,scale=0.45]{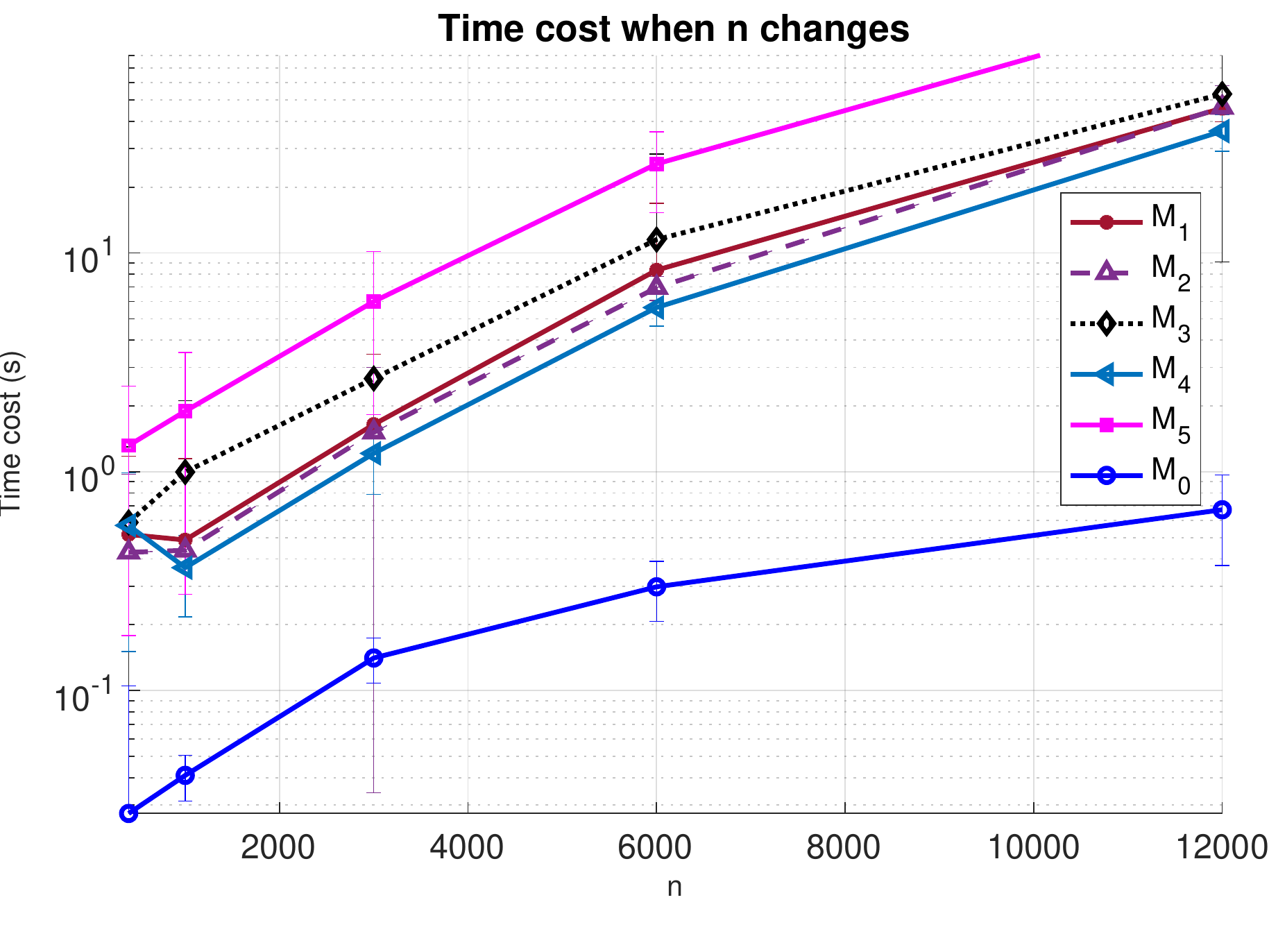}} \\[-3ex]
      
   \caption{Log-scale error rates and runtime (with error bars) v.s. $n$ under the setting of equal cluster size and $\Delta^2= (\lambda^{*}\bar\Delta_\gamma)^{2}$. Zero error is displayed as $10^{-6}$ in the log-scale plot.}
   \label{fig:run_time6}
\end{figure}

\begin{figure}[h!] 
   \centering
      \subfigure{\includegraphics[trim={1.5cm 7cm 1.5cm 7cm},clip,scale=0.45]{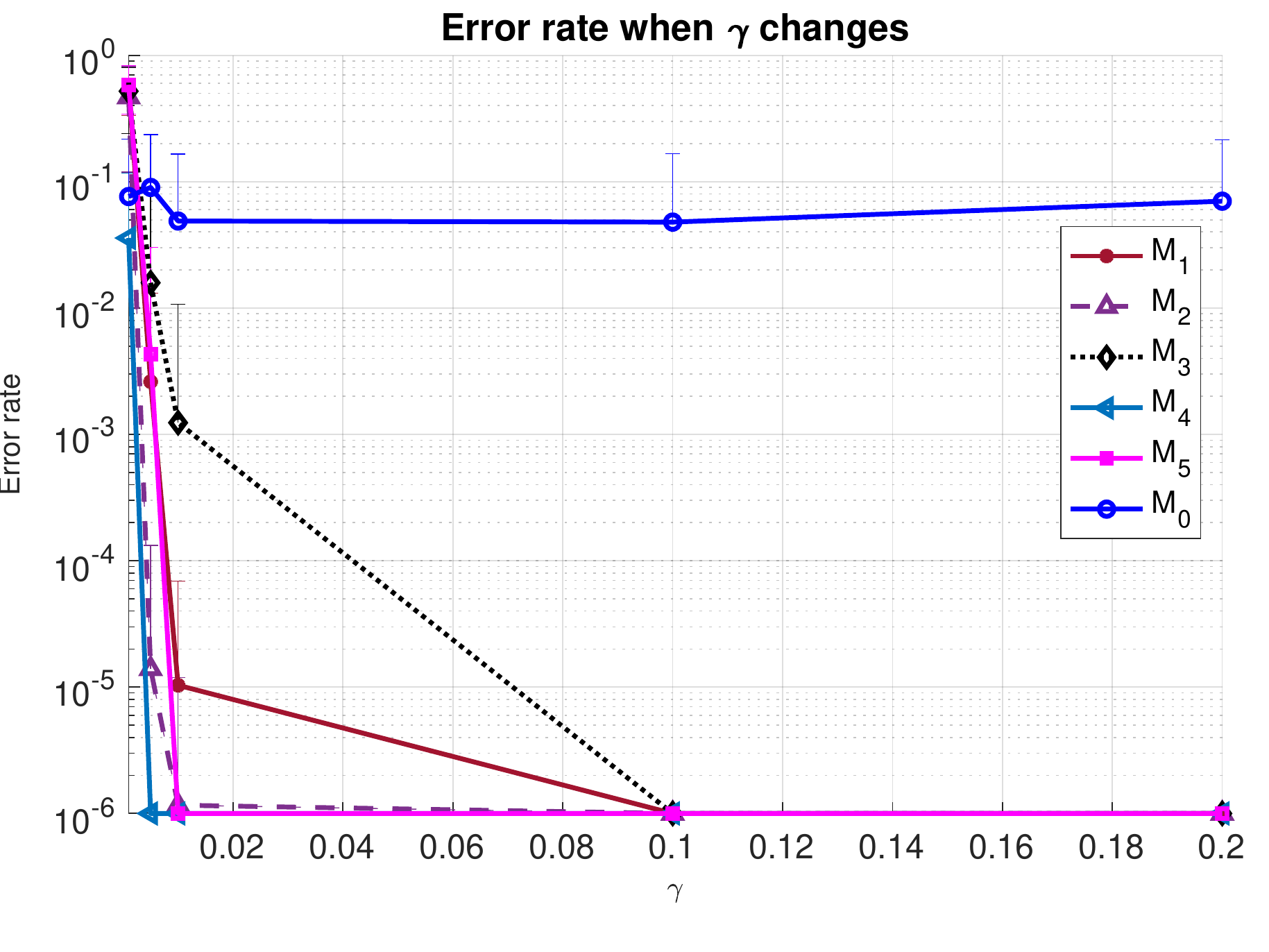}} 
      \subfigure{\includegraphics[trim={1.5cm 7cm 1.5cm 7cm},clip,scale=0.45]{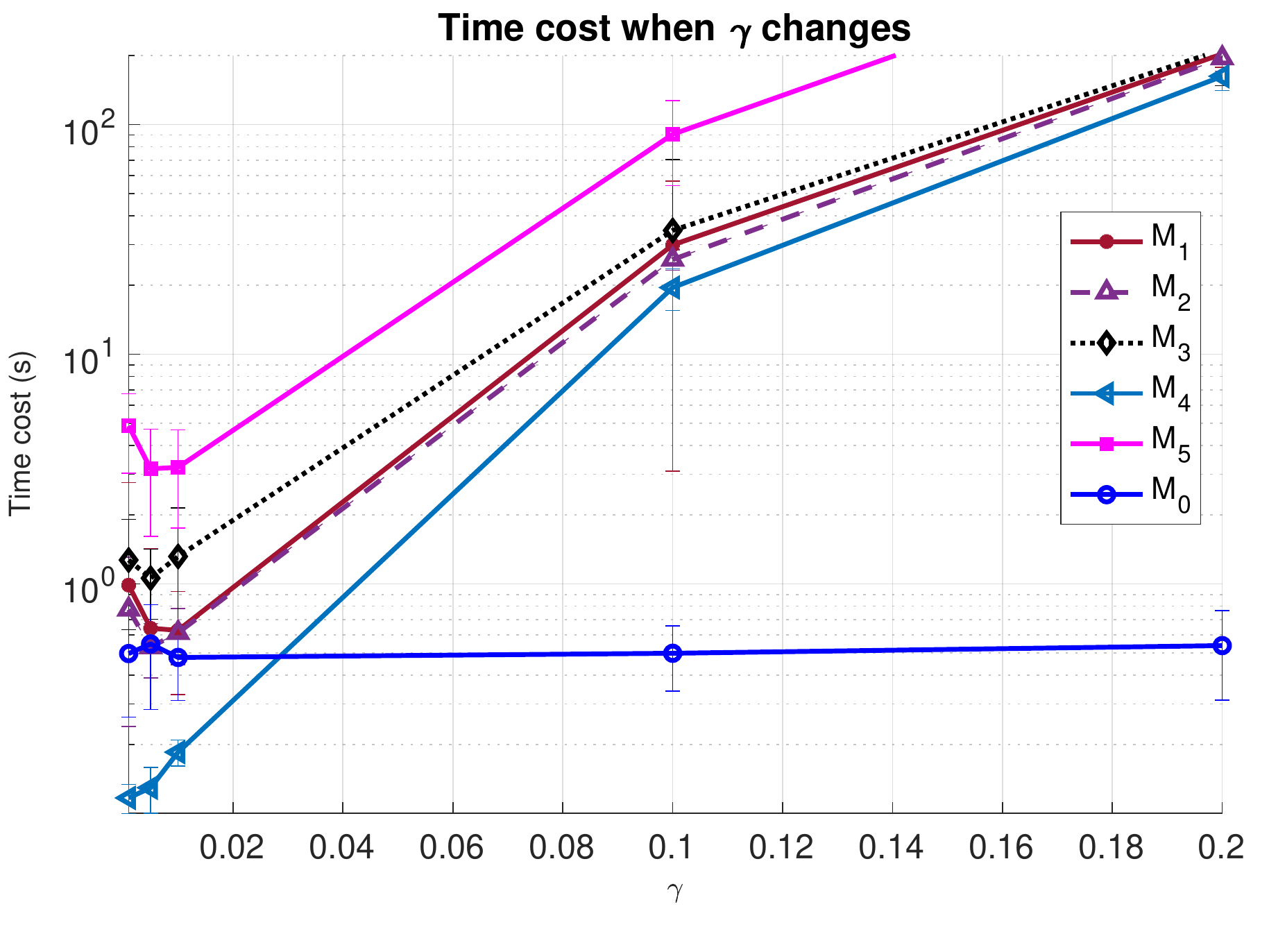}} \\[-3ex]
      
   \caption{Log-scale error rates and runtime (with error bars) v.s. $\gamma$ under the setting of equal cluster size and $\Delta^2= (\lambda^{*}\bar\Delta_\gamma)^{2}$. Zero error is displayed as $10^{-6}$ in the log-scale plot.}
   \label{fig:run_time7}
\end{figure}

\begin{figure}[h!] 
   \centering
      \subfigure{\includegraphics[trim={1.5cm 7cm 1.5cm 7cm},clip,scale=0.45]{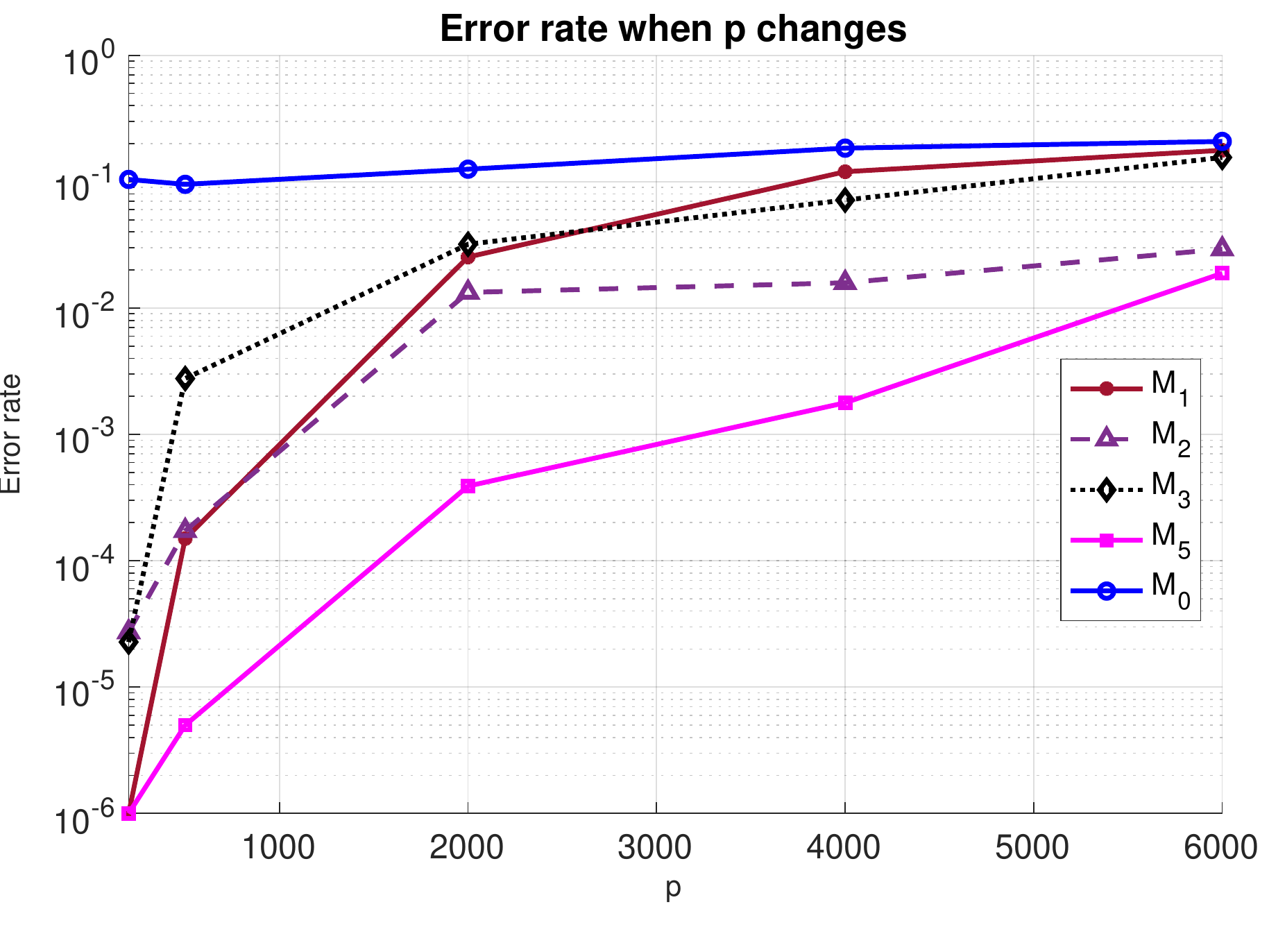}} 
      \subfigure{\includegraphics[trim={1.5cm 7cm 1.5cm 7cm},clip,scale=0.45]{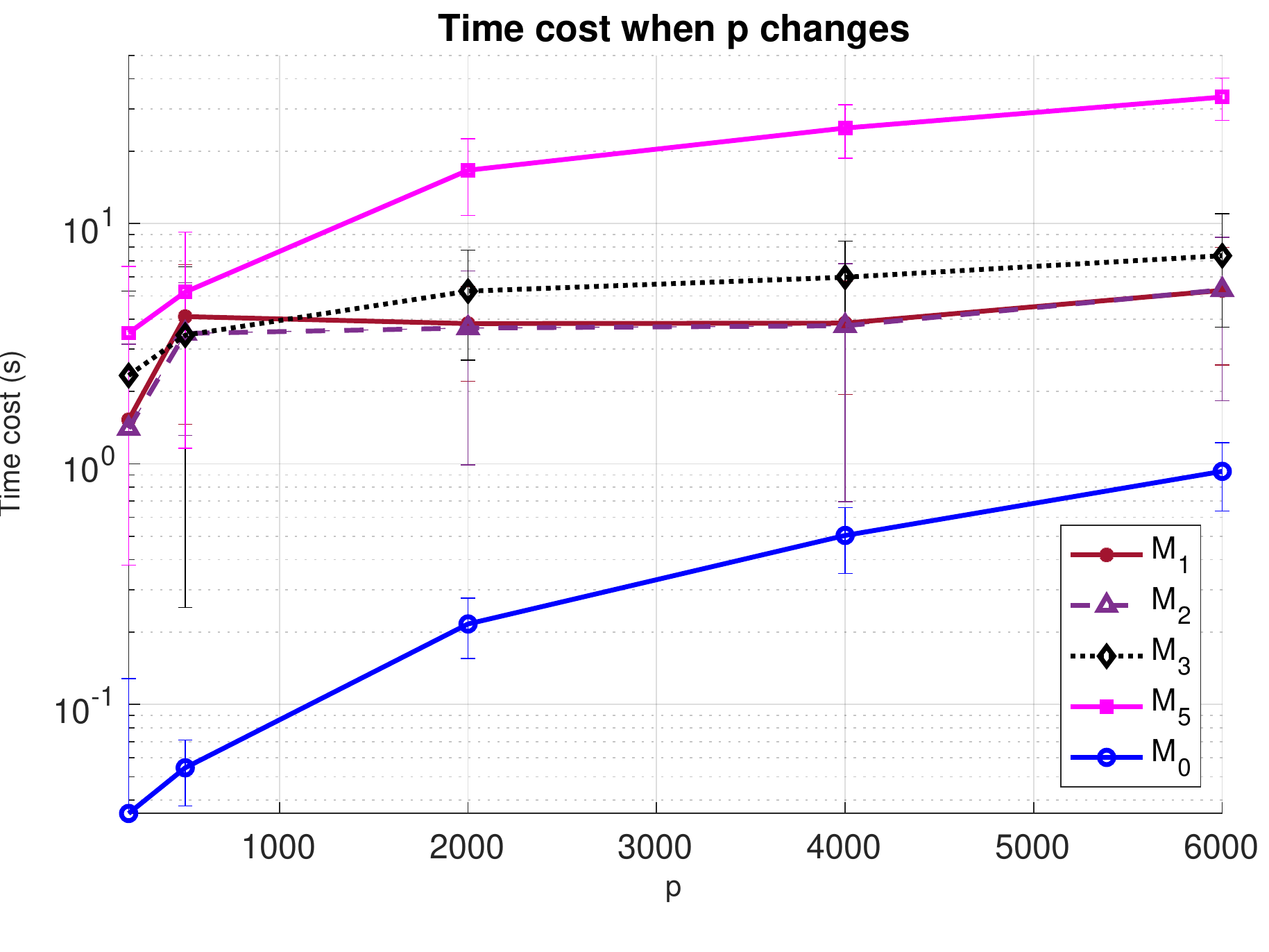}} \\[-3ex]
      
   \caption{Log-scale error rates and runtime (with error bars) v.s. $p$ under the setting of unequal cluster size ($n_1=n_2=n/8, n_3=n_4=3n/8$) and $\Delta^2= (\lambda^{*}\bar\Delta_\ast)^{2}$. Zero error is displayed as $10^{-6}$ in the log-scale plot. }
   \label{fig:run_time8}
\end{figure}

\begin{figure}[h!] 
   \centering
      \subfigure{\includegraphics[trim={1.5cm 7cm 1.5cm 7cm},clip,scale=0.45]{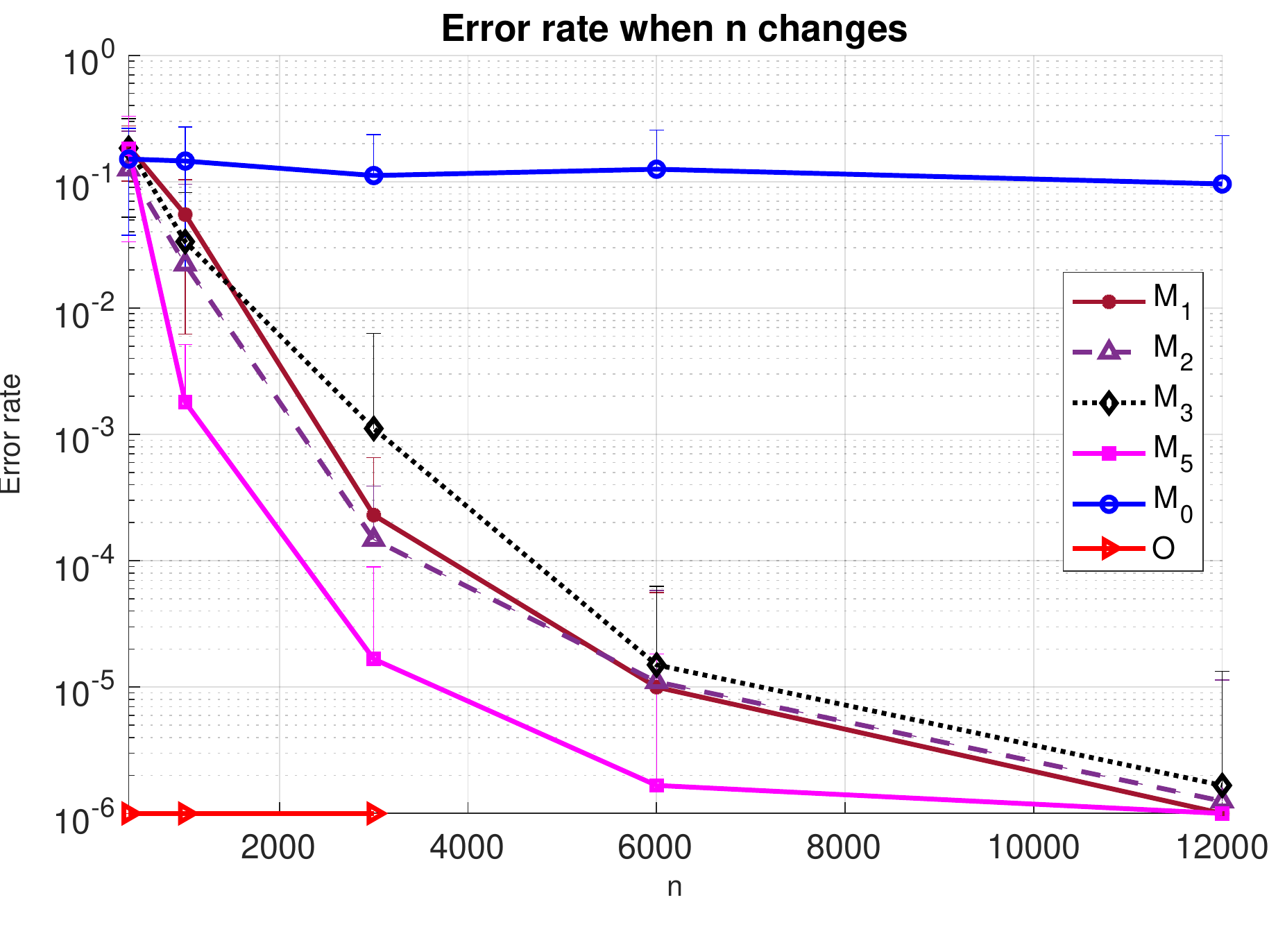}} 
      \subfigure{\includegraphics[trim={1.5cm 7cm 1.5cm 7cm},clip,scale=0.45]{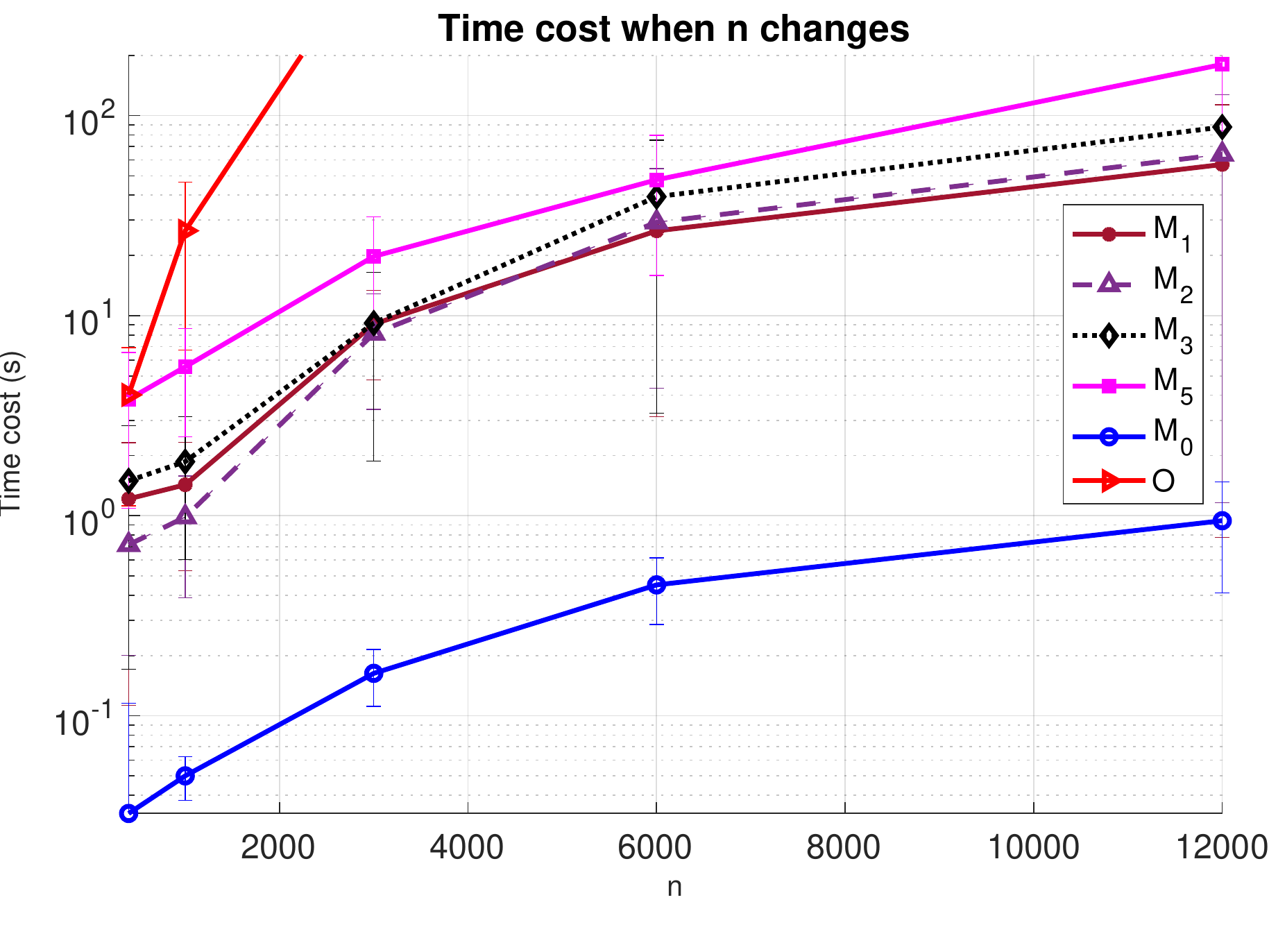}} \\[-3ex]
      
   \caption{Log-scale error rates and runtime (with error bars) v.s. $n$ under the setting of unequal cluster size ($n_1=n_2=n/8, n_3=n_4=3n/8$) and $\Delta^2= (\lambda^{*}\bar\Delta_\ast)^{2}$. Zero error is displayed as $10^{-6}$ in the log-scale plot.}
   \label{fig:run_time9}
\end{figure}

\begin{figure}[h!] 
   \centering
      \subfigure{\includegraphics[trim={1.5cm 7cm 1.5cm 7cm},clip,scale=0.45]{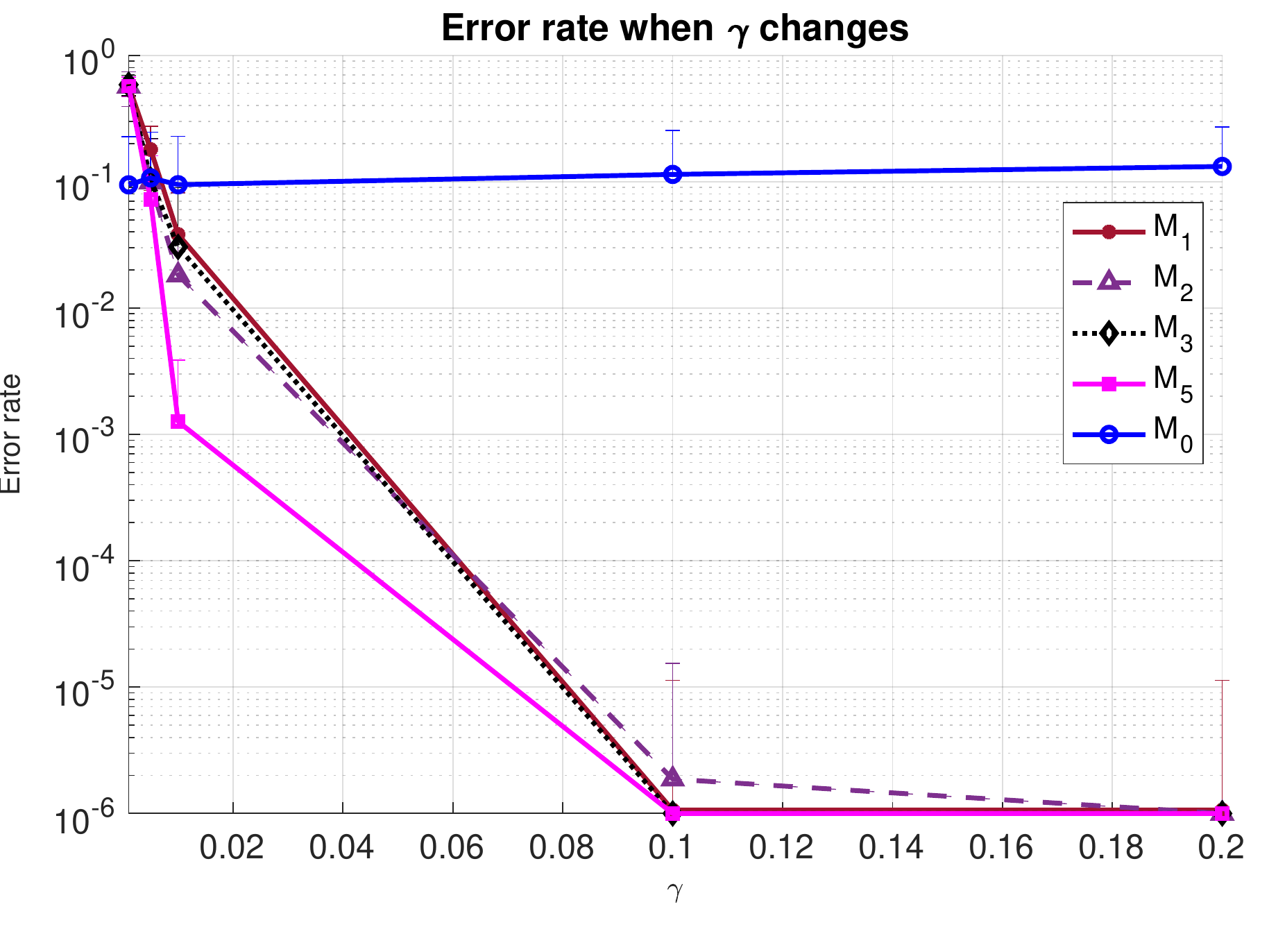}} 
      \subfigure{\includegraphics[trim={1.5cm 7cm 1.5cm 7cm},clip,scale=0.45]{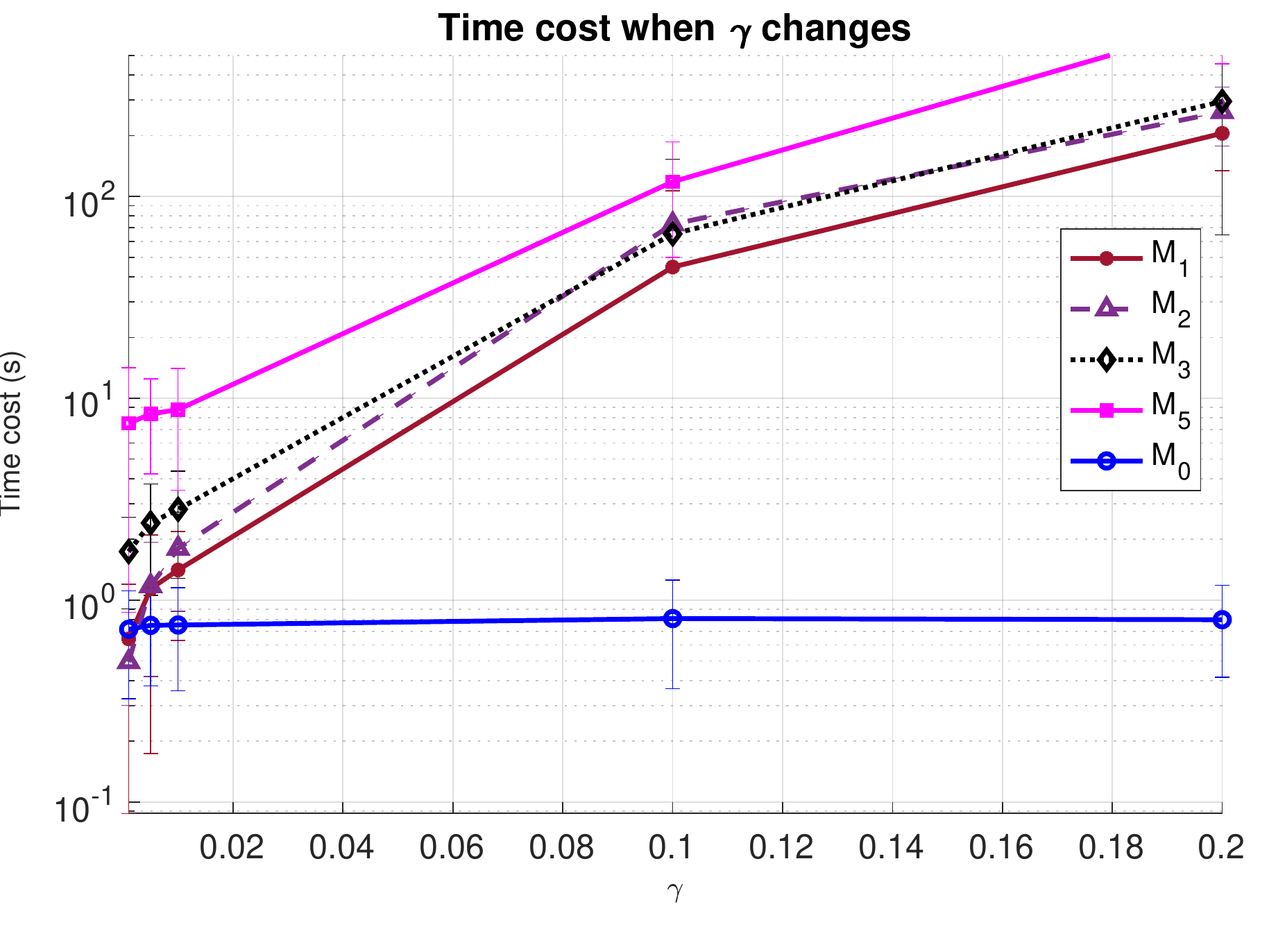}} \\[-3ex]
      
   \caption{Log-scale error rates and runtime (with error bars) v.s. $\gamma$ under the setting of unequal cluster size ($n_1=n_2=n/8, n_3=n_4=3n/8$) and $\Delta^2=(\lambda^{*}\bar\Delta_\ast)^{2}$. Zero error is displayed as $10^{-6}$ in the log-scale plot.}
   \label{fig:run_time10}
\end{figure}

\begin{figure}[h!] 
   \centering
      \subfigure{\includegraphics[trim={1.5cm 7cm 1.5cm 7cm},clip,scale=0.45]{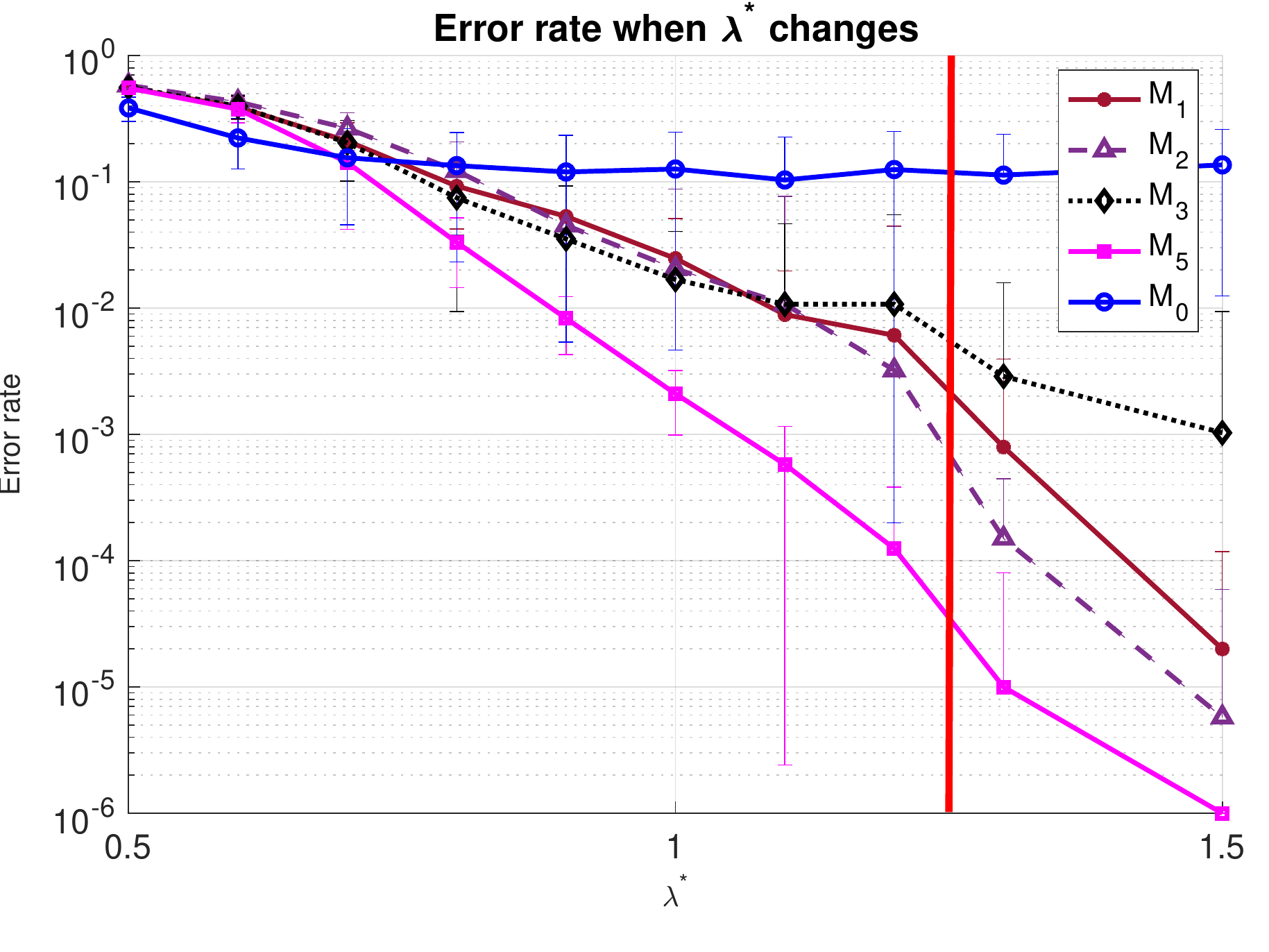}} 
      \subfigure{\includegraphics[trim={1.5cm 7cm 1.5cm 7cm},clip,scale=0.45]{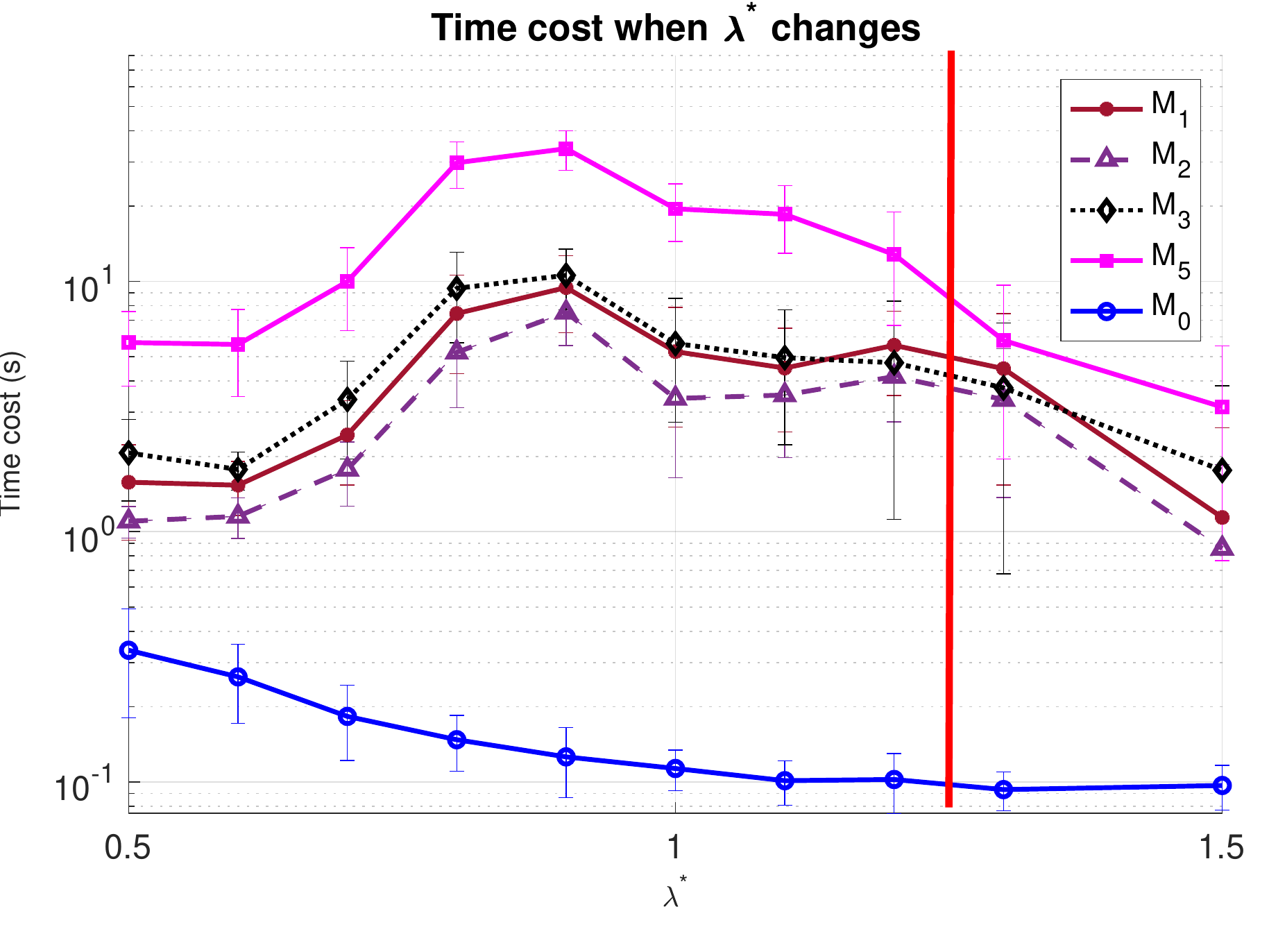}} \\[-3ex]
      
   \caption{Log-scale error rates and runtime (with error bars) v.s. $\lambda^*$ under the setting of unequal cluster size ($n_1=n_2=n/8, n_3=n_4=3n/8$) and $\Delta^2= (\lambda^{*}\bar\Delta_\ast)^{2}$. Red vertical line indicates theoretical threshold $\bar\Delta_\gamma^{'2}$ for SL methods. Zero error is displayed as $10^{-6}$ in the log-scale plot.}
   \label{fig:run_time11}
\end{figure}

\begin{figure}[h!] 
   \centering
      \subfigure{\includegraphics[trim={1.5cm 7cm 1.5cm 7cm},clip,scale=0.45]{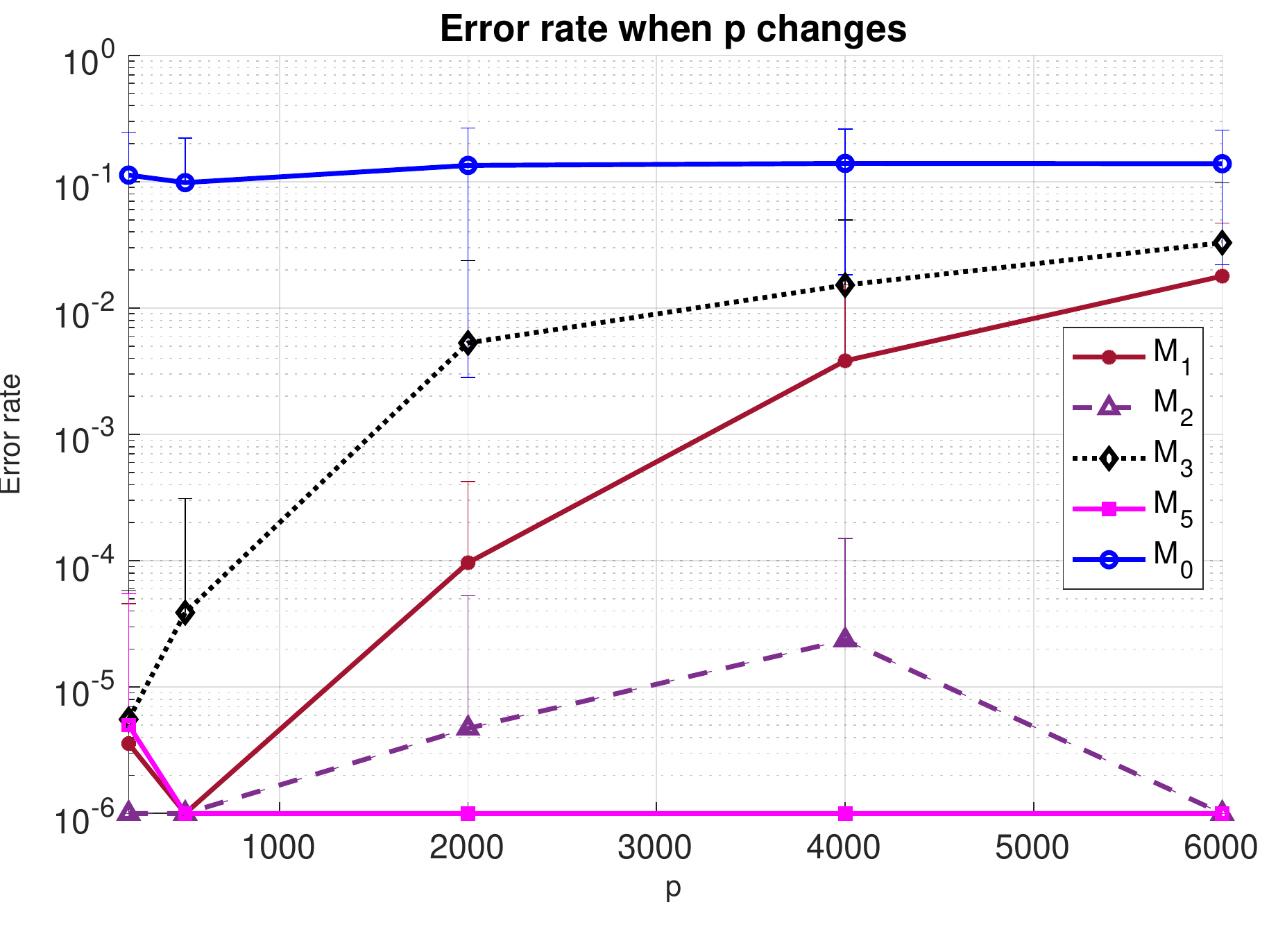}} 
      \subfigure{\includegraphics[trim={1.5cm 7cm 1.5cm 7cm},clip,scale=0.45]{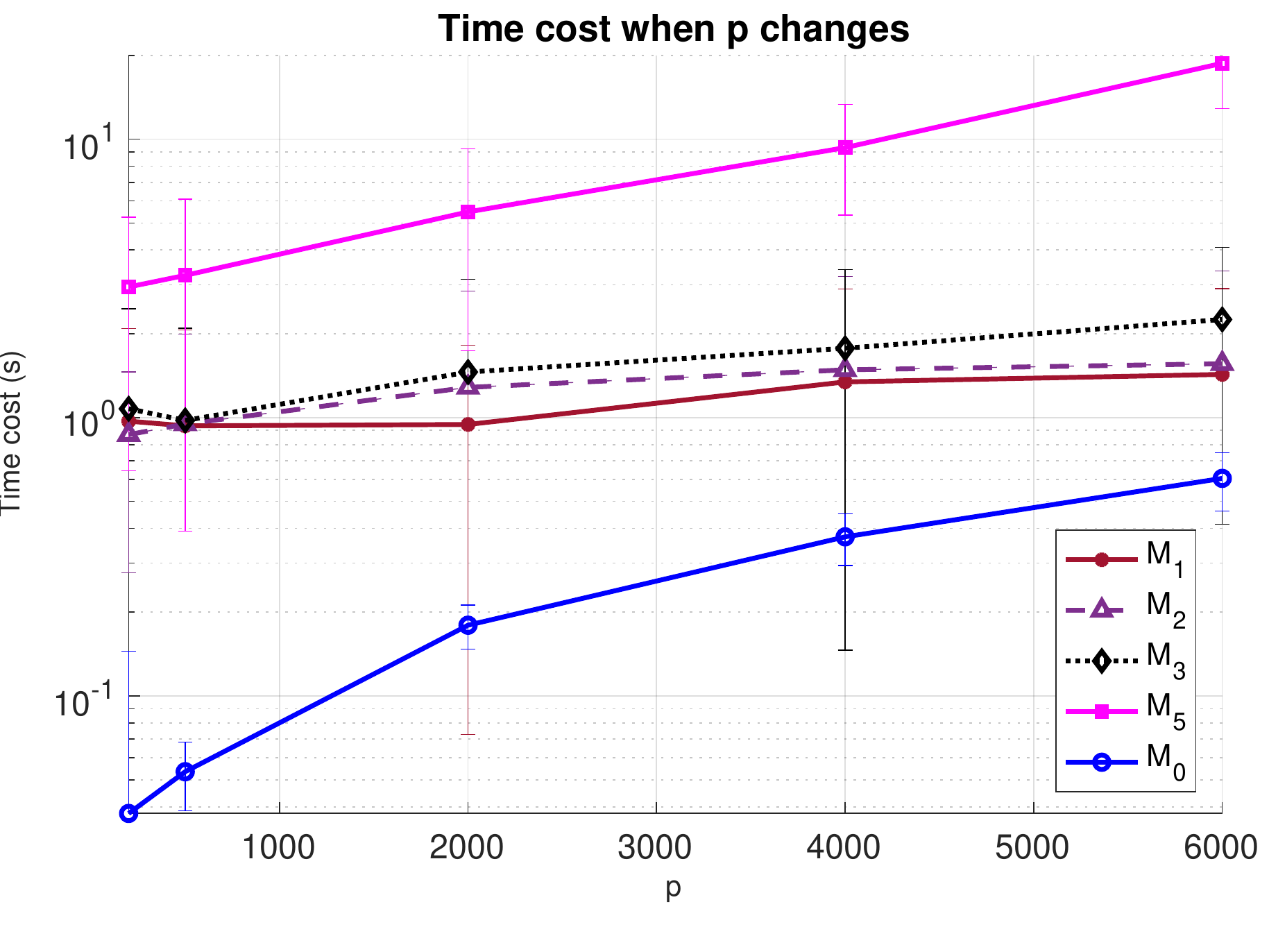}} \\[-3ex]
      
   \caption{Log-scale error rates and runtime (with error bars) v.s. $p$ under the setting of unequal cluster size ($n_1=n_2=n/8, n_3=n_4=3n/8$) and $\Delta^2= (\lambda^{*}\bar\Delta'_\gamma)^{2}$. Zero error is displayed as $10^{-6}$ in the log-scale plot.}
   \label{fig:run_time12}
\end{figure}

\begin{figure}[h!] 
   \centering
      \subfigure{\includegraphics[trim={1.5cm 7cm 1.5cm 7cm},clip,scale=0.45]{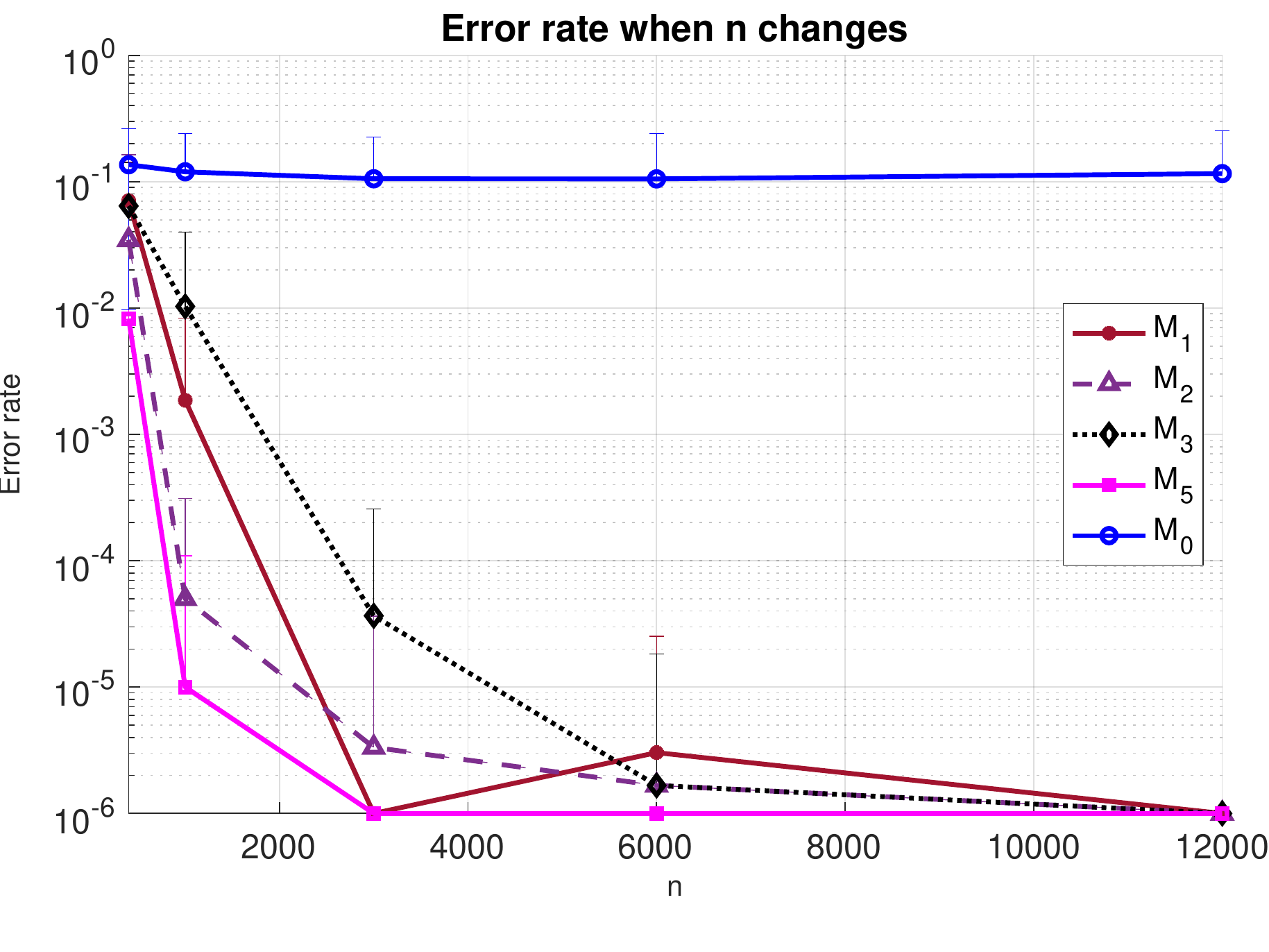}} 
      \subfigure{\includegraphics[trim={1.5cm 7cm 1.5cm 7cm},clip,scale=0.45]{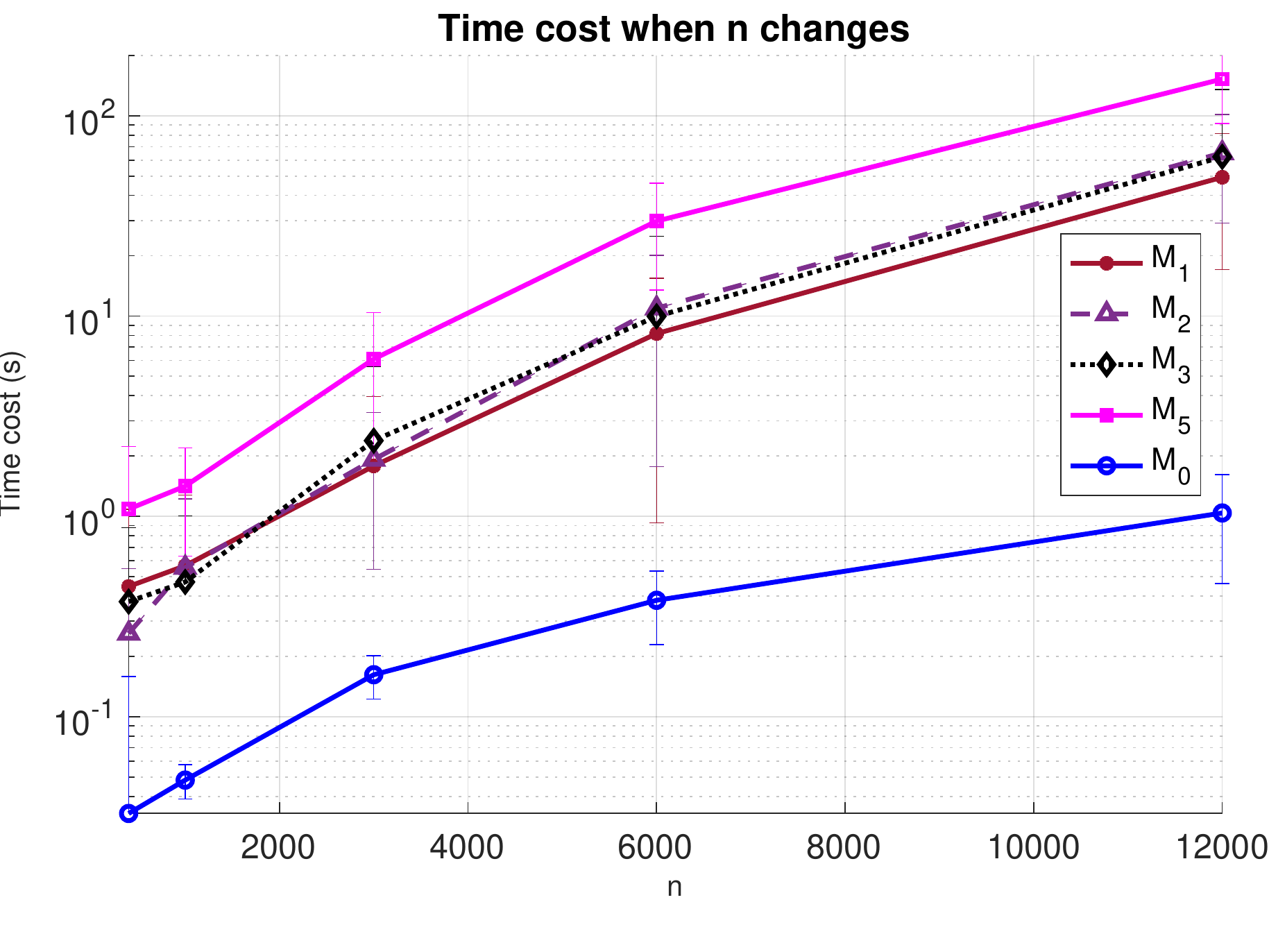}} \\[-3ex]
      
   \caption{Log-scale error rates and runtime (with error bars) v.s. $n$ under the setting of unequal cluster size ($n_1=n_2=n/8, n_3=n_4=3n/8$) and $\Delta^2= (\lambda^{*}\bar\Delta'_\gamma)^{2}$. Zero error is displayed as $10^{-6}$ in the log-scale plot.}
   \label{fig:run_time13}
\end{figure}

\begin{figure}[h!] 
   \centering
      \subfigure{\includegraphics[trim={1.5cm 7cm 1.5cm 7cm},clip,scale=0.45]{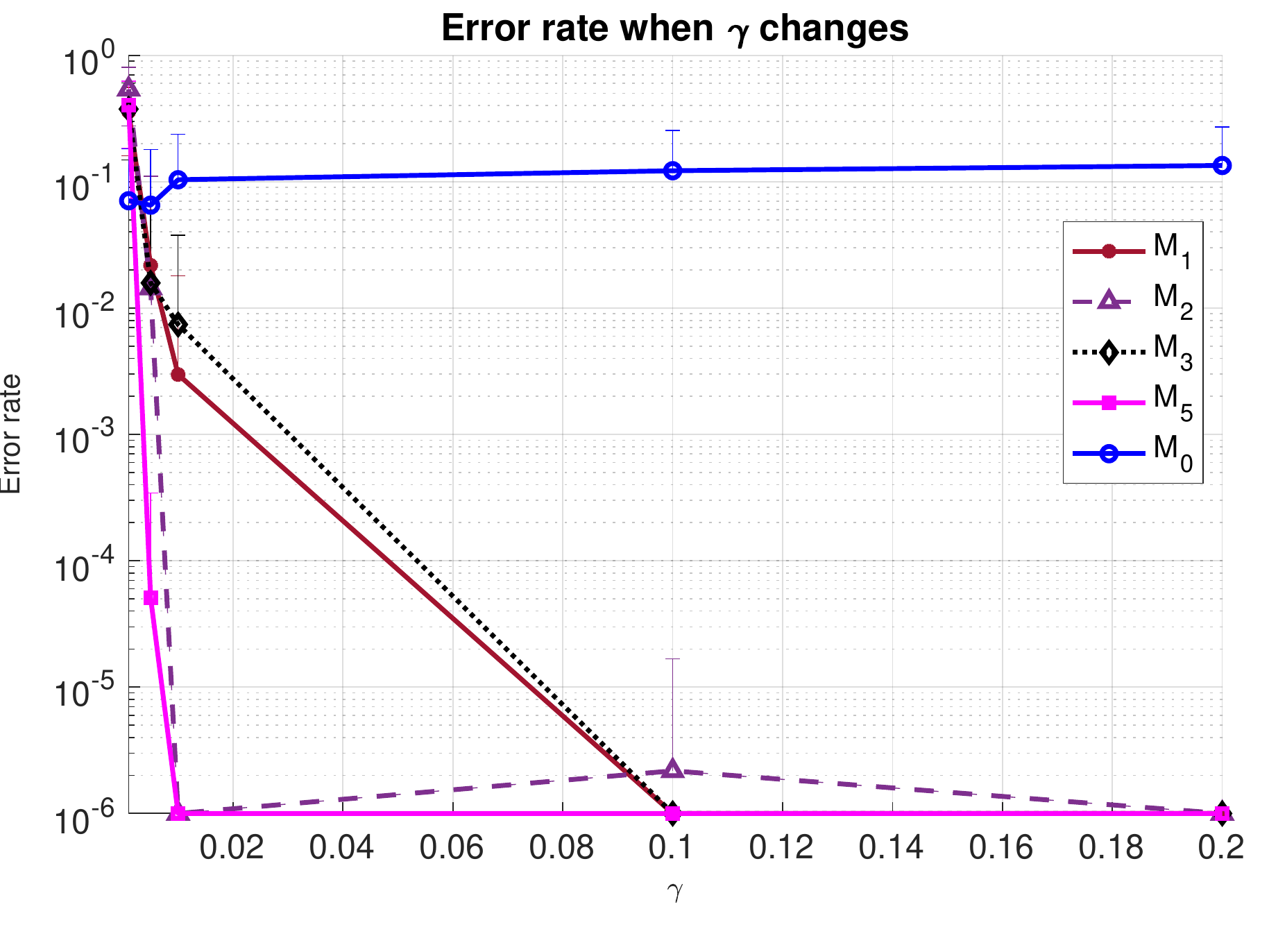}} 
      \subfigure{\includegraphics[trim={1.5cm 7cm 1.5cm 7cm},clip,scale=0.45]{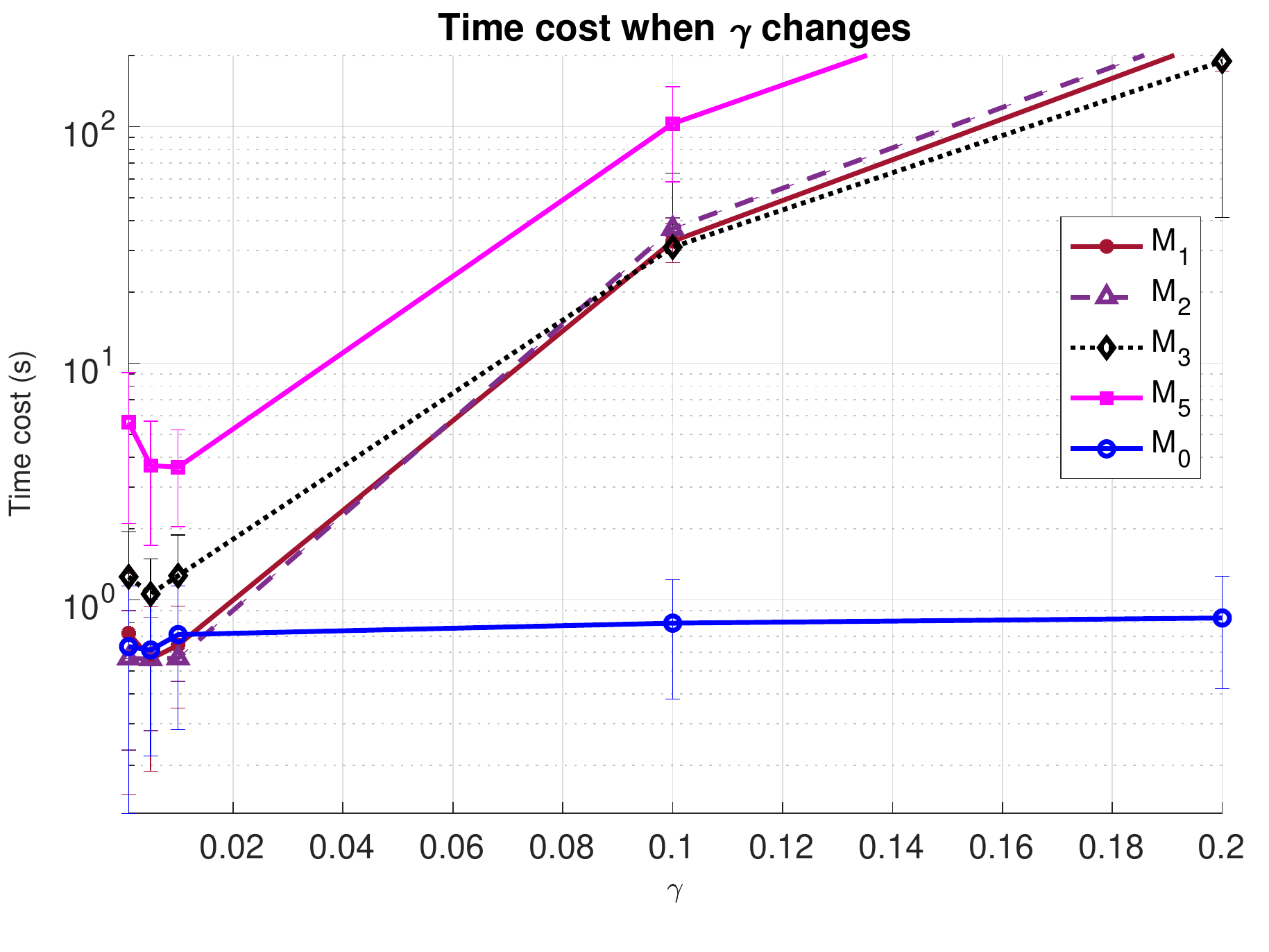}} \\[-3ex]
      
   \caption{Log-scale error rates and runtime (with error bars) v.s. $\gamma$ under the setting of unequal cluster size ($n_1=n_2=n/8, n_3=n_4=3n/8$) and $\Delta^2= (\lambda^{*}\bar\Delta'_\gamma)^{2}$. Zero error is displayed as $10^{-6}$ in the log-scale plot.}
   \label{fig:run_time14}
\end{figure}

\begin{figure}[h!] 
   \centering
      \subfigure{\includegraphics[trim={1.5cm 7cm 1.5cm 7cm},clip,scale=0.47]{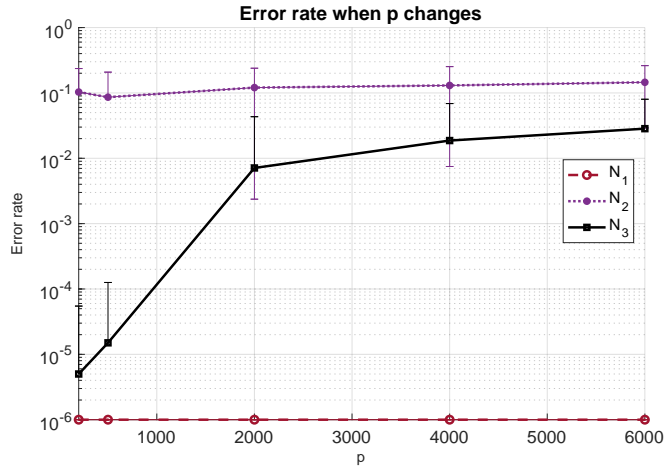} } \\[-3ex]
   \caption{Log-scale error rates (with error bars) v.s. $p$ under the setting of unequal cluster size ($n_1=n_2=n/8, n_3=n_4=3n/8$) and $\Delta^2= (\lambda^{*}\bar\Delta'_\gamma)^{2}$. $N_1$ is WSL when we plug in the true weights. $N_2$ is $K$-means++ method. $N_3$ is WSL when we plug in the weights based on $K$-means++ method. Zero error is displayed as $10^{-6}$ in the log-scale plot. }
   \label{fig:run_time15}
\end{figure}

\begin{figure}[h!] 
   \centering
   \subfigure{\includegraphics[trim={1.5cm 7cm 1.5cm 7cm},clip,scale=0.45]{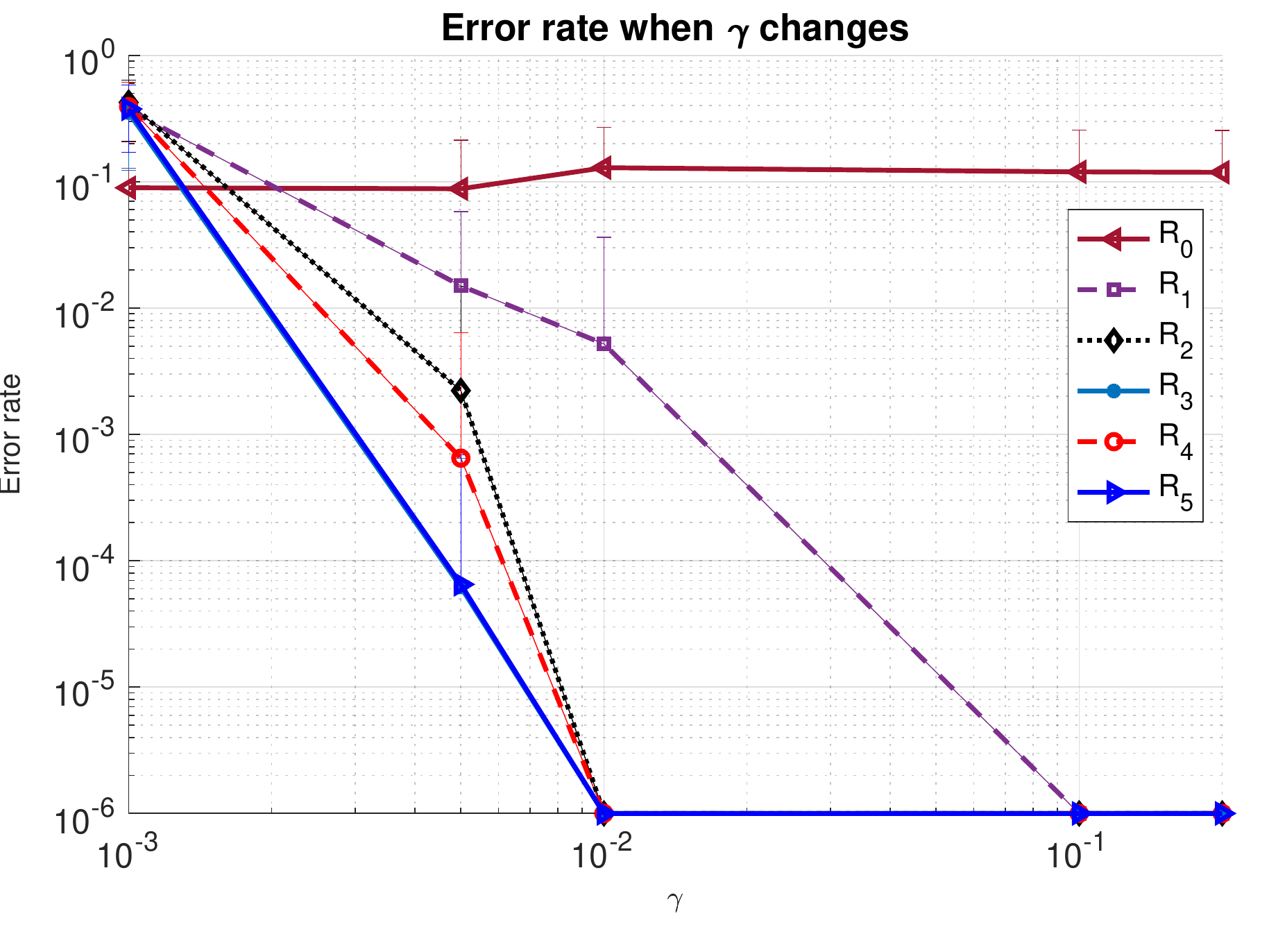} } 
   \subfigure{\includegraphics[trim={1.5cm 7cm 1.5cm 7cm},clip,scale=0.45]{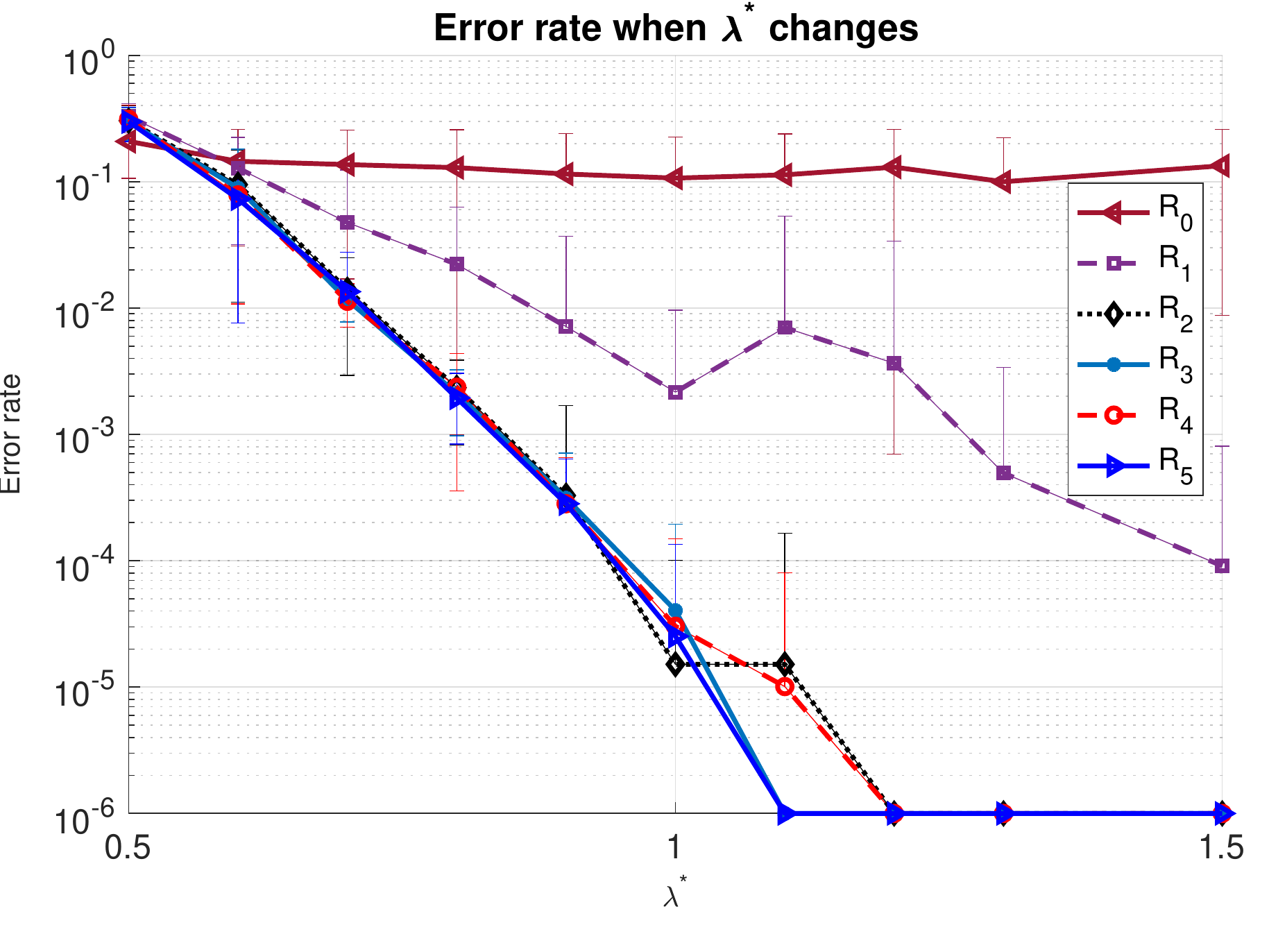} } \\[-3ex]
   \subfigure{\includegraphics[trim={1.5cm 7cm 1.5cm 7cm},clip,scale=0.45]{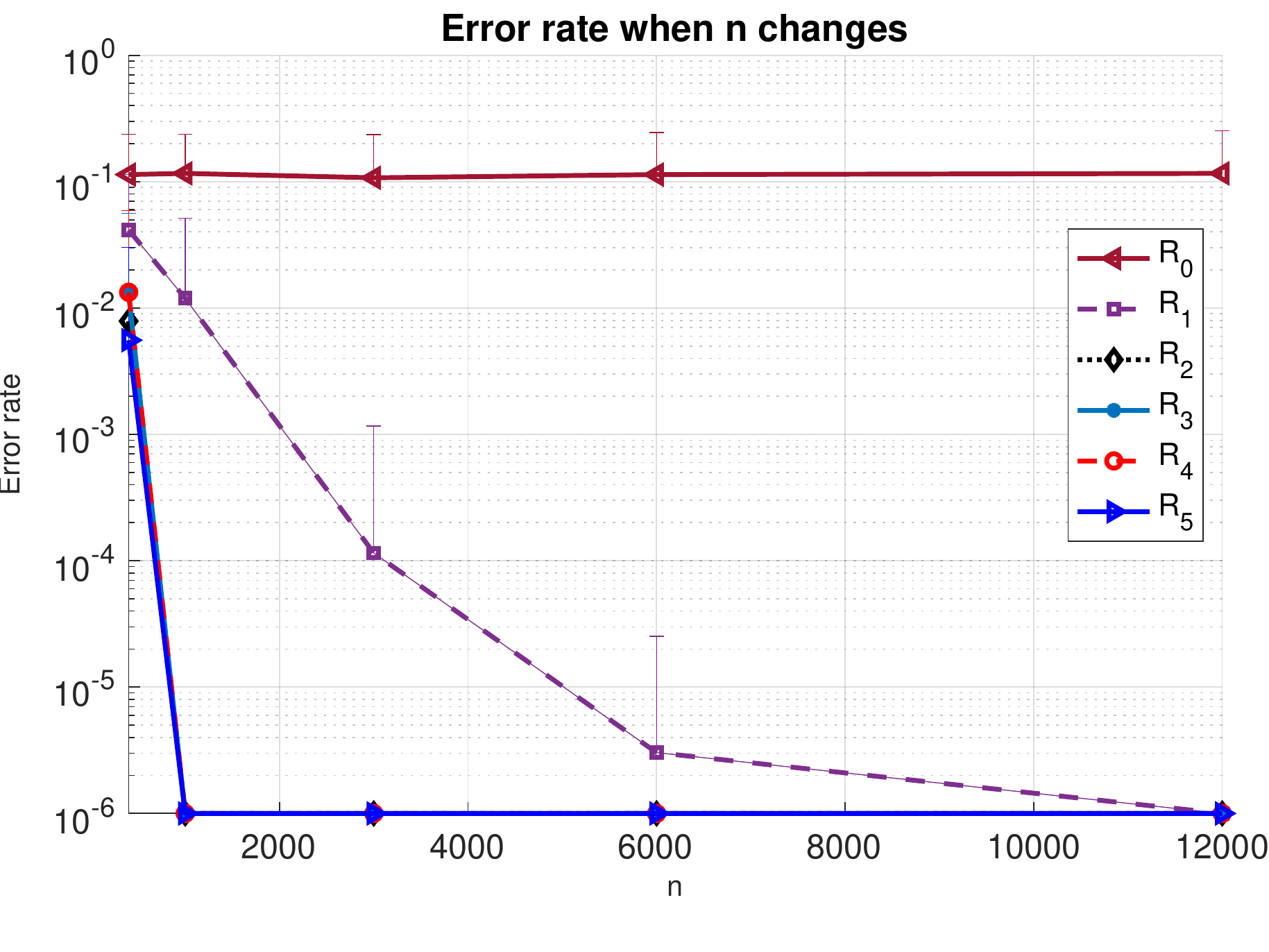} } 
   \subfigure{\includegraphics[trim={1.5cm 7cm 1.5cm 7cm},clip,scale=0.45]{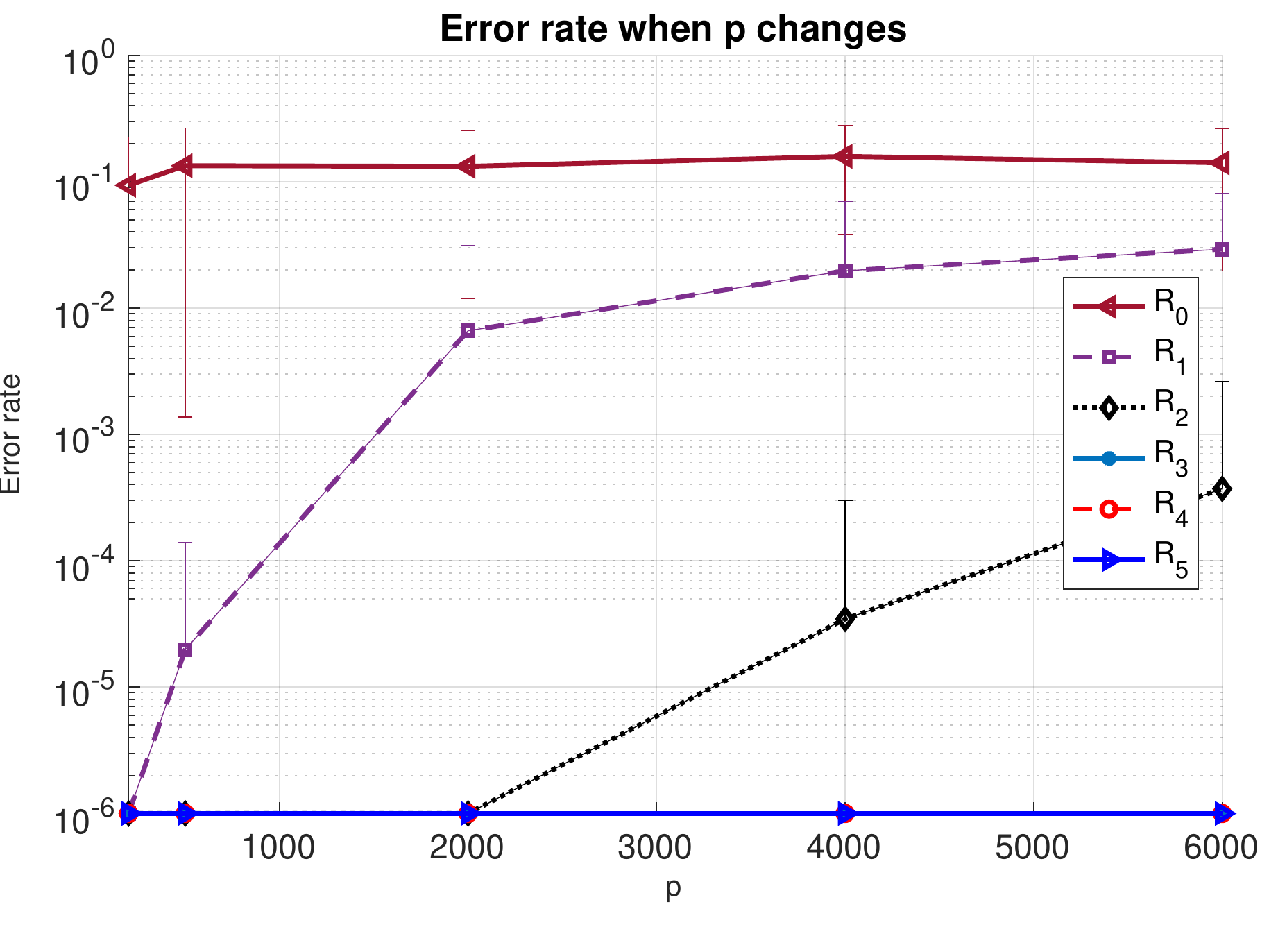} } \\[-3ex]
   \caption{Error rates of MR-WSL. $R_0$ stands for the initial $K$-means++ as the warm start for our MR-WSL and $R_i$ stands for the $i$-th round of MR-WSL. Zero error is displayed as $10^{-6}$ in the log-scale plot. In particular, we take the log-scale for $x$ axis when $\gamma$ changes. The settings for Figure~\ref{fig:multi-round_WSL16} are the same as Figure~\ref{fig:run_time15}.  From the plots we can see that after 3-4 rounds, the performance gets stable and achieved optimal.}
   \label{fig:multi-round_WSL16}
\end{figure}

In this section, we show the full simulation results. 
The error rates and running times on log-scale are shown in Figures~\ref{fig:run_time1} to~\ref{fig:multi-round_WSL16} (with captions describing the simulation setup). The baseline setup is $\lambda^{*}=1.2, p=1000, K=4, \gamma=0.1, n=2000$ by default, expect when $\gamma$ is changing, we use $n=10000$. For simplicity, we set all the distance between centers to be equal. i.e., $ \|\mu_k-\mu_l\|^2=\Delta^2,$ $\forall 1 \leq k \neq l \leq K.$ In particular, we arrange all the centers on vertices of a regular simplex. i.e., $\mu_l=\Delta/\sqrt{2} \cdot e_l, \forall l\in[K]$, where $e_l$ is a vector that the $l$-th component is $1$ and $0$ otherwise. The variance of Gaussian distributions is chosen to be 1.  The detailed explanation for methods $M_0-M_5$ and $O$ can be found in Section~\ref{sec:experiment}.

Figures~\ref{fig:run_time1} to~\ref{fig:run_time7}  are under the setting of equal cluster size. Figures~\ref{fig:run_time8} to~\ref{fig:run_time14}  are for the unequal cluster size case. In particular, Figures~\ref{fig:run_time1} to~\ref{fig:run_time4},  Figures~\ref{fig:run_time8} to~\ref{fig:run_time11} are under the separation condition from the theoretical cutoff $\bar\Delta_\ast^2$ in~\eqref{eqn:full_data_threshold}; the separation for Figures~\ref{fig:run_time5} to~\ref{fig:run_time7} is the theoretical cutoff $\bar\Delta_\gamma^2$ for SL methods; and the separation for Figures~\ref{fig:run_time12} to~\ref{fig:run_time14} is the theoretical cutoff $\bar\Delta_\gamma^{'2}$ for BCSL methods.  Figures~\ref{fig:run_time1} to~\ref{fig:run_time4} correspond to Figure~\ref{fig:error_rates} and Figure~\ref{fig:run_time} in Section~\ref{sec:experiment}. By comparing Figures~\ref{fig:run_time1} to~\ref{fig:run_time4} (equal cluster size) to Figures~\ref{fig:run_time8} to~\ref{fig:run_time11} (unequal cluster size), we can see that when $p$ changes, $M_2$ and $M_3$ which are the improved methods aiming at handling large $p$ unequal cluster size settings have better performance than $M_1,$ which are consistent with our theory. Moreover, $M_5$, the multi-round WSL (at $4$-th round), is optimal among those methods in Figure~\ref{fig:run_time8}. For the time cost, we can still see that SL methods have the same linear $O(n)$ complexity as the fast $K$-means++ (with difference only occurring in the leading constant) for unequal cluster size from Figure~\ref{fig:run_time9} corresponding to Figure~\ref{fig:run_time} in Section~\ref{sec:experiment}.
We consider two threshold settings (Figures~\ref{fig:run_time5} to~\ref{fig:run_time7} and Figures~\ref{fig:run_time12} to~\ref{fig:run_time14}) in order to show that the thresholds $\bar\Delta_\gamma^2$ for SL methods and $\bar\Delta_\gamma^{'2}$ for BCSL methods are nearly optimal in the sense that they are neither too large that all methods can achieve exact recovery easily nor too small that we can observe the exact recoveries in most cases, which can be found in those figures.

\begin{figure}[h!] 
   \centering
   \subfigure{\includegraphics[trim={1.5cm 7cm 1.5cm 7cm},clip,scale=0.45]{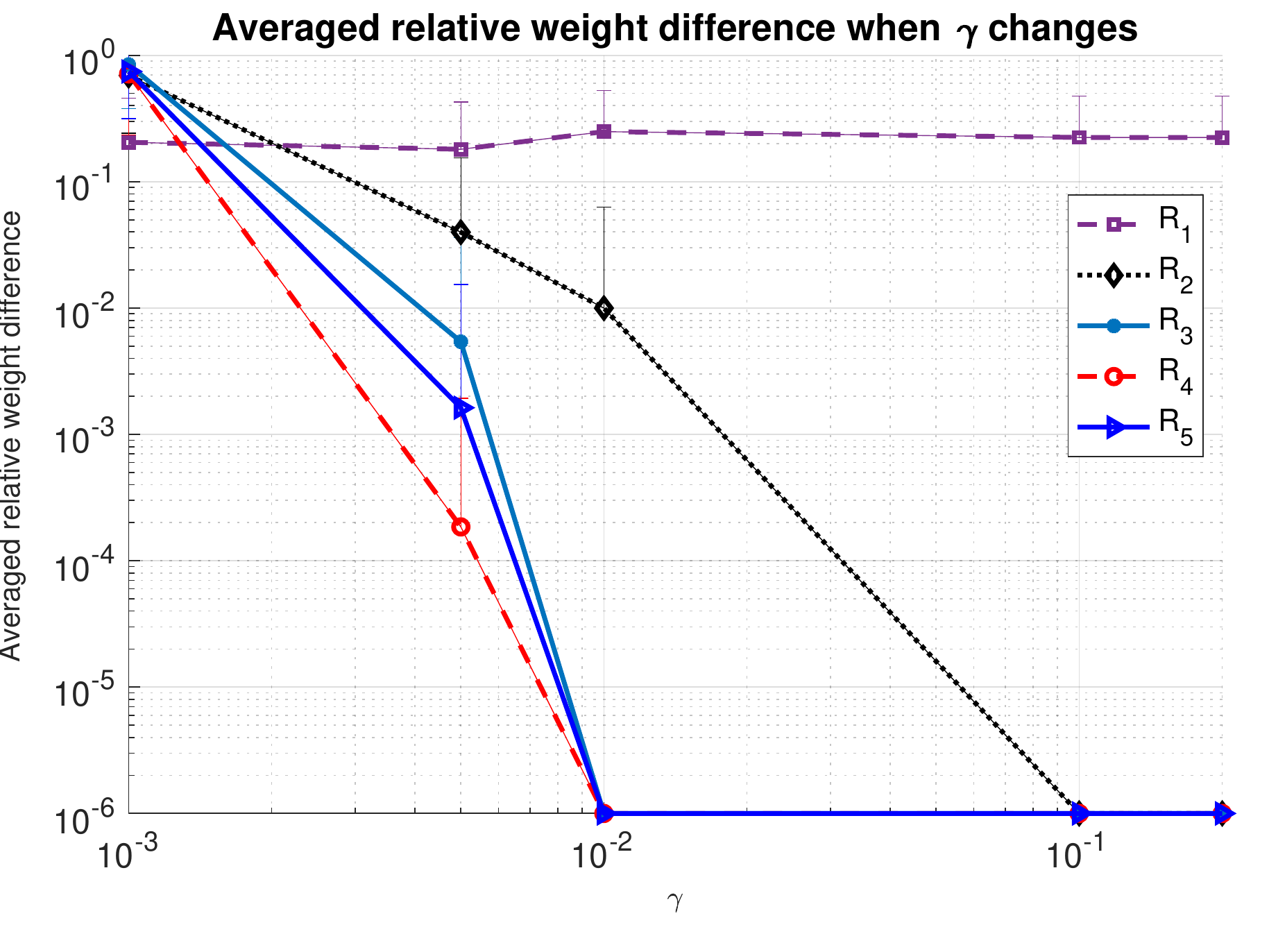} } 
   \subfigure{\includegraphics[trim={1.5cm 7cm 1.5cm 7cm},clip,scale=0.45]{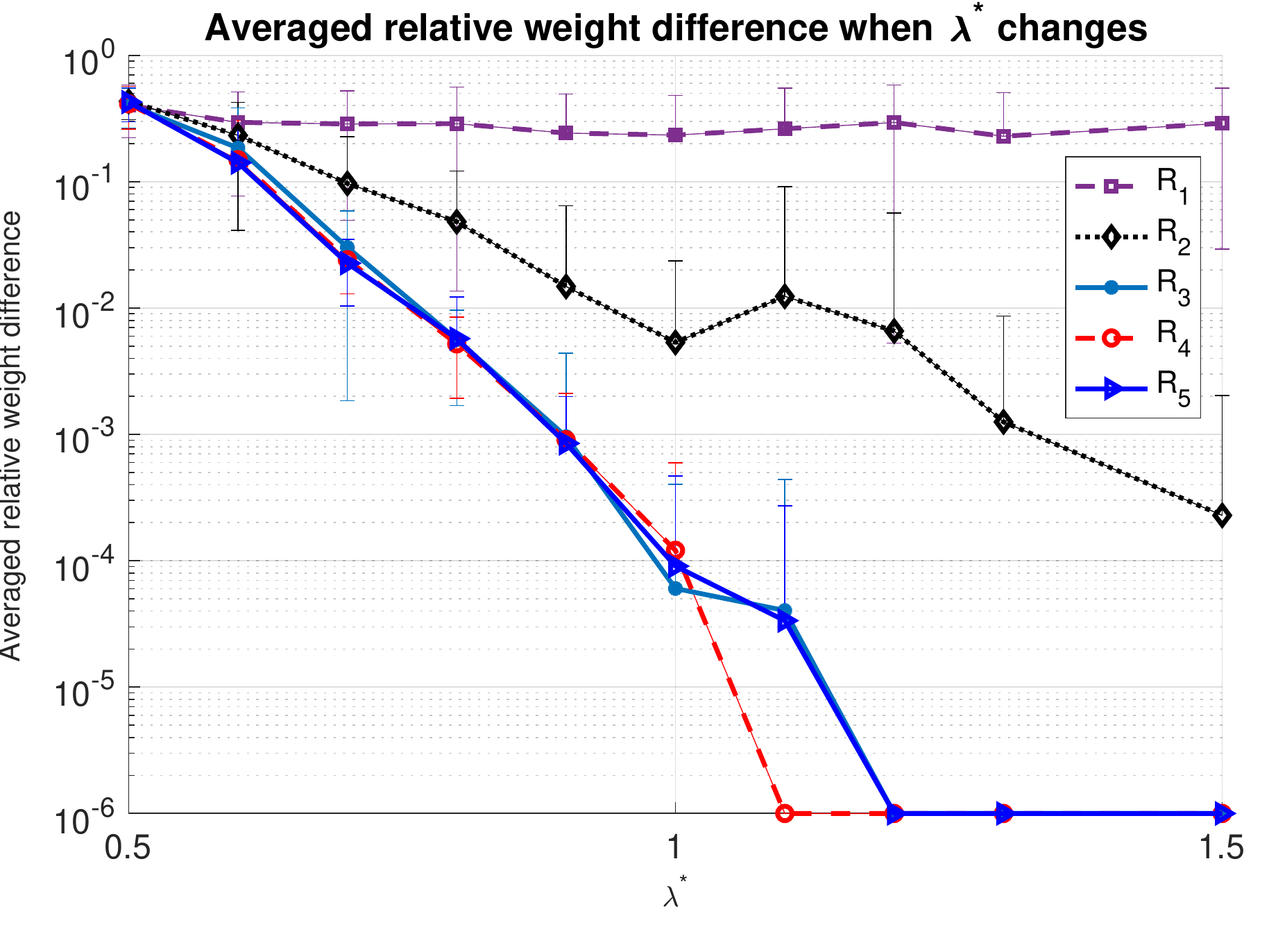} } \\[-3ex]
   \subfigure{\includegraphics[trim={1.5cm 7cm 1.5cm 7cm},clip,scale=0.45]{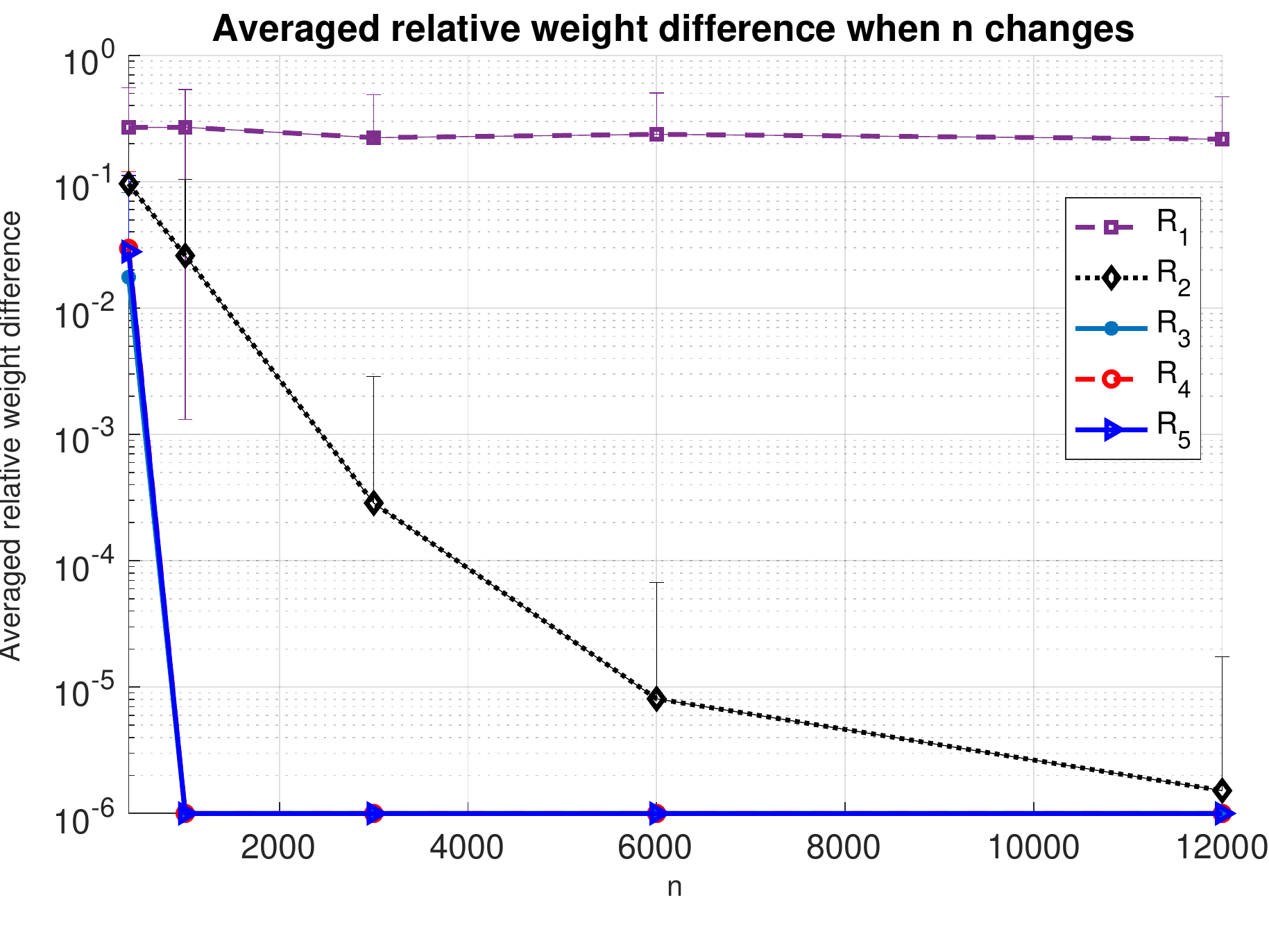} } 
   \subfigure{\includegraphics[trim={1.5cm 7cm 1.5cm 7cm},clip,scale=0.45]{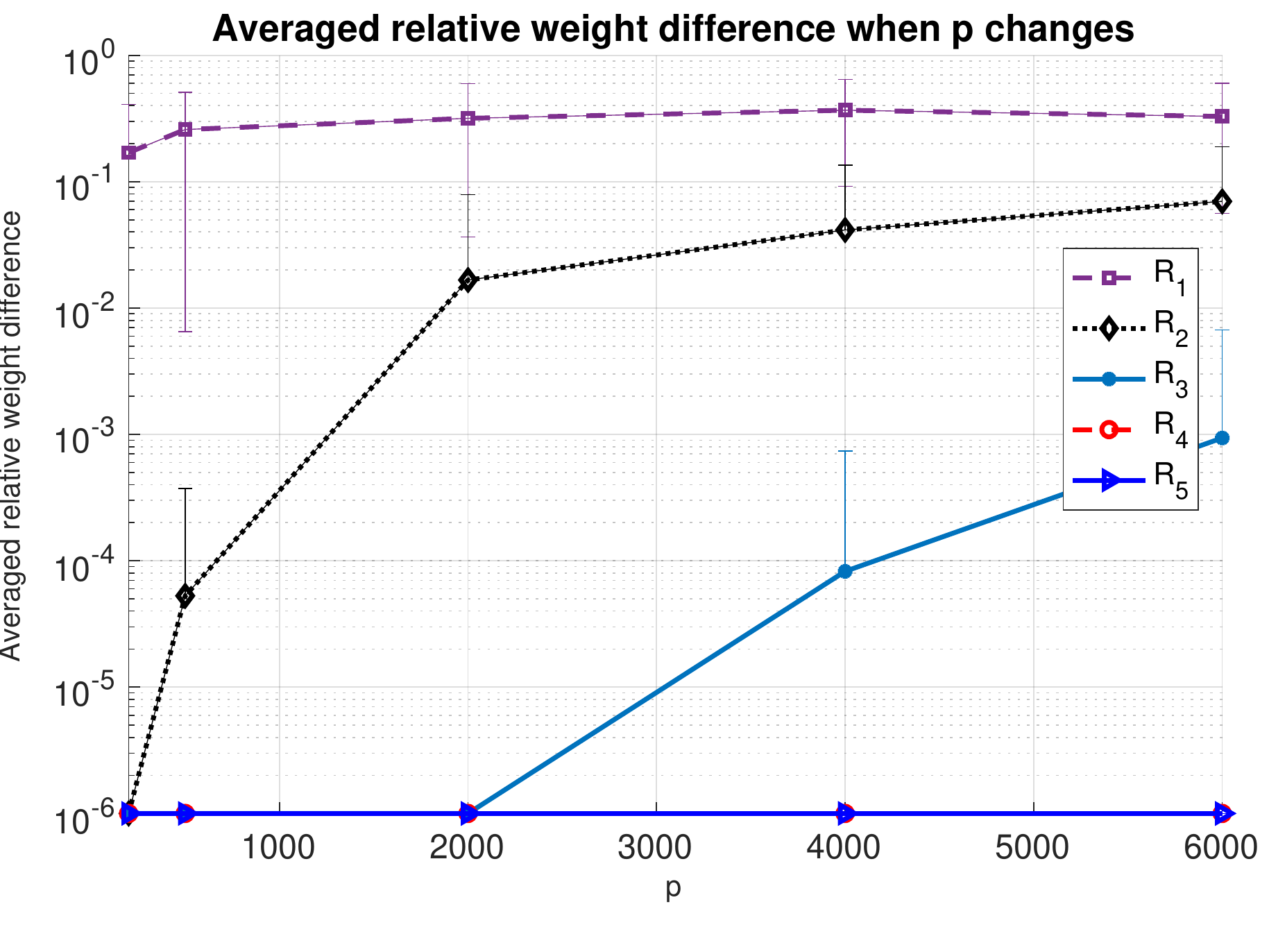} } \\[-3ex]
   \caption{Averaged relative weight difference of MR-WSL. Averaged relative weight difference is defined by $\tfrac{1}{n} \sum_{j\in [n]} |w_j-w_j^*|/w_j^*. $ where $w_j, w_j^*$ is defined in Section~\ref{subsec:WSL}. $R_i$ stands for the $i$-th round of MR-WSL corresponding to Figure~\ref{fig:multi-round_WSL16}. Zero error is displayed as $10^{-6}$ in the log-scale plot. In particular, we take the log-scale for $x$ axis when $\gamma$ changes. Figure~\ref{fig:multi-round_WSL16} and Figure~\ref{fig:multi-round_WSL17} have same settings.}
   \label{fig:multi-round_WSL17}
\end{figure}

We would like to make a remark on the initial weights estimated by the $K$-means++ algorithm, which is the default $K$-means clustering algorithm in Matlab. From Table~\ref{tab:initial_weights}, we see that as dimension $p$ increases, the fraction $(1-\delta)$ of ``good" weights within the $\epsilon$-distortion level $[(1-\epsilon) w^*_i, (1+\epsilon) w^*_i]$ decreases. This is also reflected in Figure~\ref{fig:run_time15}, where the error rate of cluster label recovery is almost constant ($N_2$ purple curve) when the separation signal is slightly above the cutoff value of exact recovery. With such a warm-start given by $K$-means++, our WSL can improve the recovery error rate in one iteration ($N_3$ black curve). Moreover, if we use the ideal weights, then the WSL can achieve the exact recovery ($N_1$ red curve in Figure~\ref{fig:run_time15}). The one-iteration WSL can be viewed as the multi-round WSL with one round, we can see from Figure~\ref{fig:multi-round_WSL16} that after 3-4 rounds, the multi-round WSL refines the clustering errors and can eventually achieve the exact recovery as if we initialize with the ideal weights. This can be explained by comparing Table \ref{tab:initial_weights} with Table \ref{tab:initial_weights2} to \ref{tab:initial_weights4}, where we can see the decreasing trends for $\delta$, which will become $0$ eventually for all $\epsilon$ we set on the grid. This shows that the weights for $R_4$ we get from iteration are fairly close to or identical to $0$. On the other hand it can be well explained by Figure~\ref{fig:multi-round_WSL17}, which is the corresponding averaged relative weight difference of $R_1$ to $R_5$ in Figure~\ref{fig:multi-round_WSL16}. Averaged relative weight difference is defined by $\tfrac{1}{n} \sum_{j\in [n]} |w_j-w_j^*|/w_j^*. $ where $w_j, w_j^*$ is defined in Section~\ref{subsec:WSL}. $R_i$ stands for the $i$-th round of MR-WSL corresponding to Figure~\ref{fig:multi-round_WSL16}. By comparing Figure~\ref{fig:multi-round_WSL16} and Figure~\ref{fig:multi-round_WSL17} we can see similar trends between the relative weight difference for $R_{i+1}$ and the error rate for $R_{i}$ when various parameters change. The relative weight difference has a decreasing trend as we perform more rounds, which accounts for the improvement of MR-WSL shown in Figure~\ref{fig:multi-round_WSL16}.


\begin{figure}[h!] 
   \centering
      \subfigure{\includegraphics[trim={1.5cm 7cm 1.5cm 7cm},clip,scale=0.45]{eno.pdf}} 
      \subfigure{\includegraphics[trim={1.5cm 7cm 1.5cm 7cm},clip,scale=0.45]{tno.pdf}} \\[-3ex]
      
   \caption{Log-scale error rates and runtime (with error bars) v.s. $n$ under the setting of equal cluster size and $\Delta^2= (\lambda^{*}\bar\Delta_\ast)^{2}$. Zero error is displayed as $10^{-6}$ in the log-scale plot.}
   \label{fig:run_time2}
\end{figure}

\vspace{0.5in}

\begin{table}[h]
\caption{Fraction of $(\epsilon,\delta)$-weights initialized by $K$-means++ algorithm ($1$-st round of WSL, $R_1$ for Figure~\ref{fig:multi-round_WSL16}). }
\vspace{.2in}
\label{tab:initial_weights}
\begin{center}
\begin{tabular}{|L{2cm}|L{2cm}|L{2cm}|L{2cm}|L{2cm}|L{2cm}|L{2cm}|L{2cm}|L{2cm}| } 
\hline
$\epsilon$ & 0 & 0.2 & 0.4 & 0.6 & 0.8 & 1.0 \\
\hline
$p=200$ & 0.2602 & 0.25373 & 0.21828 & 0.10665 & 0.071097 & 0.054791 \\
$p=500$ & 0.2725 & 0.25598 & 0.18602 & 0.10836 & 0.06278 & 0.049165 \\
$p=2000$ & 0.44123 & 0.37208 & 0.27103 & 0.12647 & 0.085495 & 0.065855 \\
$p=4000$ & 0.40125 & 0.33282 & 0.24355 & 0.10086 & 0.068535 & 0.05453 \\
$p=6000$ & 0.48375 & 0.38929 & 0.25716 & 0.13754 & 0.07653 & 0.06671 \\
 \hline
\end{tabular}
\end{center}
   \begin{tablenotes}
  \item We fix a grid of $\epsilon$ and report the estimated values of $\delta$ satisfying~\eqref{eqn:sampling_weights_condition} for each $\epsilon$. Each $\delta$ is averaged over $100$ simulations. The setting for Table~\ref{tab:initial_weights} is the same as Figure~\ref{fig:multi-round_WSL16}.
    \end{tablenotes}

\end{table}

\begin{table}[h]
\caption{Fraction of $(\epsilon,\delta)$-weights for the $2$-nd round of WSL ($R_2$ for Figure~\ref{fig:multi-round_WSL16}). }
\vspace{.2in}
\label{tab:initial_weights2}
\begin{center}
\begin{tabular}{|L{2cm}|L{2cm}|L{2cm}|L{2cm}|L{2cm}|L{2cm}|L{2cm}|L{2cm}|L{2cm}| } 
\hline
$\epsilon$ & 0 & 0.2 & 0.4 & 0.6 & 0.8 & 1.0 \\
\hline
$p=200$ & 0.005 & 0.000005 & 0.000005 & 0.000005 & 0.000005 & 0.000005 \\
$p=500$ & 0.02625 & 0.00006 & 0.00006  & 0.00006  & 0.00006  & 0.00006  \\
$p=2000$ & 0.13125 & 0.028845 & 0.01318 & 0.012105 & 0.012105 & 0.00967 \\
$p=4000$ & 0.3725 & 0.088965 & 0.04746 & 0.02566 & 0.02266 & 0.0198 \\
$p=6000$ & 0.54875 & 0.15886 & 0.095185 & 0.057355 & 0.044775 & 0.03841 \\
 \hline
\end{tabular}

\end{center}
   \begin{tablenotes}
      \item We fix a grid of $\epsilon$ and report the estimated values of $\delta$ satisfying~\eqref{eqn:sampling_weights_condition} for each $\epsilon$. Each $\delta$ is averaged over $100$ simulations. The setting for Table~\ref{tab:initial_weights2} is the same as Figure~\ref{fig:multi-round_WSL16}. 
    \end{tablenotes}

\end{table}
\begin{table}[h]
\caption{Fraction of $(\epsilon,\delta)$-weights for the $3$-rd round of WSL ($R_3$ for Figure~\ref{fig:multi-round_WSL16}).  }
\vspace{.2in}
\label{tab:initial_weights3}
\begin{center}
\begin{tabular}{|L{2cm}|L{2cm}|L{2cm}|L{2cm}|L{2cm}|L{2cm}|L{2cm}|L{2cm}|L{2cm}| } 
\hline
$\epsilon$ & 0 & 0.2 & 0.4 & 0.6 & 0.8 & 1.0 \\
\hline
$p=200$ & 0 & 0 & 0 & 0 & 0 & 0 \\
$p=500$ & 0 & 0 & 0 & 0 & 0 & 0 \\
$p=2000$ & 0 & 0 & 0 & 0 & 0 & 0 \\
$p=4000$ & 0.00875 & 0.005 & 0.00125 & 0.00125 & 0.00125 & 0.00125 \\
$p=6000$ & 0.02125 & 0.00005  & 0.00005  & 0.00005 & 0.00005  & 0.00005  \\
 \hline
\end{tabular}
\end{center}
  \begin{tablenotes}
      \item We fix a grid of $\epsilon$ and report the estimated values of $\delta$ satisfying~\eqref{eqn:sampling_weights_condition} for each $\epsilon$. Each $\delta$ is averaged over $100$ simulations. The setting for Table~\ref{tab:initial_weights3} is the same as Figure~\ref{fig:multi-round_WSL16}.
    \end{tablenotes}

\end{table}

\begin{table}[h]
\caption{Fraction of $(\epsilon,\delta)$-weights for the $4$-th round of WSL ($R_4$ for Figure~\ref{fig:multi-round_WSL16}).  }
\vspace{.2in}
\label{tab:initial_weights4}
\begin{center}
\begin{tabular}{|L{2cm}|L{2cm}|L{2cm}|L{2cm}|L{2cm}|L{2cm}|L{2cm}|L{2cm}|L{2cm}| } 
\hline
$\epsilon$ & 0 & 0.2 & 0.4 & 0.6 & 0.8 & 1.0 \\
\hline
$p=200$ & 0 & 0 & 0 & 0 & 0 & 0 \\
$p=500$ & 0 & 0 & 0 & 0 & 0 & 0 \\
$p=2000$ & 0 & 0 & 0 & 0 & 0 & 0 \\
$p=4000$ & 0 & 0 & 0 & 0 & 0 & 0 \\
$p=6000$ & 0 & 0 & 0 & 0 & 0 & 0 \\
 \hline
\end{tabular}
\end{center}
  \begin{tablenotes}
      \item We fix a grid of $\epsilon$ and report the estimated values of $\delta$ satisfying~\eqref{eqn:sampling_weights_condition} for each $\epsilon$. Each $\delta$ is averaged over $100$ simulations. The setting for Table~\ref{tab:initial_weights4} is the same as Figure~\ref{fig:multi-round_WSL16}.
    \end{tablenotes}

\end{table}

\end{document}